\pdfoutput=1
\documentclass{article}
\usepackage{iclr2025_conference,times}
\PassOptionsToPackage{hyphens}{url}
\usepackage{hyperref}

\usepackage[utf8]{inputenc} 
\usepackage[T1]{fontenc}    
\usepackage{url}            
\usepackage{booktabs}       
\usepackage{amsfonts}       
\usepackage{nicefrac}       
\usepackage{microtype}      
\usepackage{xcolor}         
\usepackage{graphicx}       
\usepackage{subcaption}

\usepackage{amsmath}
\usepackage{amsthm}
\usepackage{amssymb}
\usepackage{bbm}
\usepackage{mathtools}
\usepackage{cleveref}
\usepackage{algorithm}
\usepackage{algorithmic}
\usepackage{caption}
\usepackage{wrapfig} 
\usepackage{multirow}
\usepackage{comment}
\usepackage{titletoc}
\usepackage{float}
\usepackage{thm-restate}
\usepackage{hyperref}
\usepackage{url}

\usepackage{listings}

\definecolor{background}{RGB}{0,168,167}
\newtheorem{theorem}{Theorem}

\newtheorem{assum}{Assumption}

\newtheorem{lemma}{Lemma}
\crefname{assumption}{assumption}{assumptions}
\Crefname{assumption}{Assumption}{Assumptions}
\crefname{assum}{assumption}{assumptions}
\Crefname{assum}{Assumption}{Assumptions}
\crefname{lstlisting}{listing}{listings}
\Crefname{lstlisting}{Listing}{Listings}
\Crefname{equation}{Eq.}{Eqs.}

\allowdisplaybreaks[4]

\newcommand{\codebase}[0]{https://github.com/sowmaster/Sample-Level-Loss-Reweighting-ICLR-2025}

\iclrfinalcopy
\begin{document}

\title{Dynamic Loss-Based Sample Reweighting for Improved Large Language Model Pretraining}

\author{%
    Daouda A. Sow\thanks{Daouda Sow and Herbert Woisetschl\"ager were interns at IBM Research when most of this work was done.} \\
    The Ohio State University\\
    \texttt{sow.53@osu.edu}
    \And
    Herbert Woisetschl\"ager$^{*}$ \\
    Technical University of Munich\\
    \texttt{h.woisetschlaeger@tum.de}
    \And
    Saikiran Bulusu\\
    The Ohio State University\\
    \texttt{bulusu.11@osu.edu}
    \And
    Shiqiang Wang\\
    IBM Research\\
    \texttt{wangshiq@us.ibm.com}\!\!\!
    \And
    Hans-Arno Jacobsen\\
    University of Toronto \\
    \texttt{jacobsen@eecg.toronto.edu}\!\!\!
    \And
    Yingbin Liang\\
    The Ohio State University\\
    \texttt{liang889@osu.edu}
}

\maketitle

\begin{abstract}
Pretraining large language models (LLMs) on vast and heterogeneous datasets is crucial for achieving state-of-the-art performance across diverse downstream tasks. However, current training paradigms treat all samples equally, overlooking the importance or relevance of individual samples throughout the training process. Existing reweighting strategies, which primarily focus on group-level data importance, fail to leverage fine-grained instance-level information and do not adapt dynamically to individual sample importance as training progresses. In this paper, we introduce novel algorithms for dynamic, instance-level data reweighting aimed at improving both the efficiency and effectiveness of LLM pretraining. Our methods adjust the weight of each training sample based on its loss value in an online fashion, allowing the model to dynamically focus on more informative or important samples at the current training stage. In particular, our framework allows us to systematically devise reweighting strategies deprioritizing redundant or uninformative data, which we find tend to work best. 
Furthermore, we develop a new theoretical framework for analyzing the impact of loss-based reweighting on the convergence of gradient-based optimization, providing the first formal characterization of how these strategies affect convergence bounds. We empirically validate our approach across a spectrum of tasks, from pretraining 7B and 1.4B parameter LLMs to smaller-scale language models and linear regression problems, demonstrating that our loss-based reweighting approach can lead to faster convergence and significantly improved performance.

\end{abstract}

\section{Introduction}
The rapid advancement of large language models (LLMs) \citep{brown2020language,raffel2020exploring,touvron2023llama,chowdhery2023palm,achiam2023gpt,dubey2024llama} has revolutionized natural language processing and artificial intelligence (AI) capabilities. 
These models, trained on billions or trillions of tokens, exhibit remarkable generalization capabilities across a wide range of downstream tasks. 
As datasets grow to web-scale proportions and models become increasingly large, the need for more efficient training techniques has become paramount. 

Existing approaches to LLM pretraining predominantly involve two phases: heavy data curation \citep{longpre2023pretrainer,dubey2024llama,pmlr-v235-wettig24a,penedo2024finewebdatasetsdecantingweb} and training with uniform sampling on the constructed corpus. Data curation for LLM pretraining typically involves a combination of automated filtering techniques and manual quality checks. For instance, heuristic-based filters are often employed to remove low-quality content and deduplicate data. Some approaches use perplexity-based filtering or auxiliary classifiers \citep{gao2020pile,penedo2023refinedweb} to identify high-quality samples. Manual curation is then applied to refine these filtered datasets, often involving human evaluation of subsets of the data to ensure quality and relevance. 
However, these approaches face significant limitations due to scalability issues and the static nature of data selection. First, as the pretraining corpora grow to hundreds of billions of tokens, manual curation becomes increasingly infeasible. The sheer volume of data makes it impractical for humans to review even a small fraction of the dataset. Second, data curation methods cannot adapt to the changing importance of samples during the training process, e.g., as the model improves over time certain samples that were useful early in training can become irrelevant in later training stages. 

In addition, conventional gradient-based training pipelines often treat all data points as equally informative regardless of their individual importance at current training stage. This uniform sampling strategy, while simple, overlooks the nuanced diversity within large-scale datasets and potentially wastes computational resources on less informative samples. Recent work \citep{xie2023doremi,fan2023doge} has explored group-level reweighting strategies, which adjust the importance of entire domains or groups of data. While these methods have shown promise, they operate at a coarse level, do not fully leverage the fine-grained information available at the instance level, and still lack dynamicity in importance assignment during the training phase. 

Given these limitations, this paper seeks to address the following fundamental question: 

\textit{How can we dynamically leverage instance-level information to accelerate training and improve model performance without incurring significant computational overhead while potentially reducing the need for extensive data curation?}

Answering this question presents several intertwined technical challenges, particularly in the context of LLM pretraining. 
First, unlike in computer vision, where samples are seen repeatedly over the course of training (often hundreds of times), in LLM pretraining, individual samples are typically encountered only once due to the vast size of the datasets. This renders dynamic reweighting approaches primarily devised for multi-epoch training or relying on historical statistics, such as previous loss or gradient norms \citep{loshchilov2015online,jiang2019accelerating}, inadequate within the context of LLM pretraining. 
Second, storing previous statistics for individual samples becomes computationally infeasible when working with web-scale datasets, and methods requiring the persistent storage of sample-level data would significantly inflate resource requirements. 
Furthermore, for a general-purpose LLM, leveraging small hold-out validation sets to compute importance weights through, e.g., bilevel optimization as in \cite{grangier2023adaptive}, is ineffective. 

Addressing these challenges, our work makes the following significant contributions: 

\textbf{Instance-Level Loss-Based Reweighting Strategies.} We introduce and systematically study a variety of instance-level, loss-based reweighting strategies for improving both the efficiency and effectiveness of ML training, especially within the context of LLM pretraining. Each strategy is meticulously designed to achieve specific goals, 
such as focusing the learning dynamics on different parts of the loss distribution. 
This builds on recent work that emphasizes the importance of data diversity and sample importance in pretraining and extends previous works on group-level reweighting for LLM pretraining, such as DoReMi \cite{xie2023doremi}, DoGE \cite{fan2023doge}. It also adds a new dimension by incorporating more fine-grained per-sample dynamics rather than domain-level adjustments. In fact, combining our reweighting schemes with DoGE/DoReMi achieves better or comparable performance on various few-shot reasoning benchmarks when we train on the SlimPajama dataset. Moreover, our study reveals that, in general, strategies down-weighting the importance of low-loss samples tend to consistently yield performance improvements across various scenarios. 

\textbf{New Theoretical Framework.} We develop a new theoretical framework for analyzing the effects of loss-based reweighting on training acceleration. To the best of our knowledge, this represents the first explicit characterization of loss reweighting effects within the convergence bounds of gradient methods under widely adopted loss geometries. Our derived convergence bounds provide theoretical justification for the empirical success of downweighting low-loss samples, demonstrating faster convergence under this strategy. 

\textbf{Empirical Validation.} We conduct extensive experiments that corroborate our claims and support our theoretical findings. Notably, we demonstrate that the advantages of downweighting low-loss samples are observed across a spectrum of problem scales and complexities: 
(i) in complex, large-scale problems such as LLM pretraining with number of parameters ranging from 124M to 7B, our approach leads to notably improved performance and faster convergence; (ii) in simple, small-scale problems like linear regression, the results highlight the fundamental nature of our findings. 

\section{Related work}
\vspace{-3mm}
\textbf{Training Data Re-weighting/Selection for LLMs.}
Several recent studies \citep{xie2023doremi,chen2023skill,fan2023doge,thakkar2023self} have explored various reweighting techniques to enhance the generalization and efficiency of language models pretraining. For instance, \cite{xie2023doremi} and \cite{fan2023doge} optimize the composition of pretraining corpora to achieve better performance across pretraining domains or for out-of-domain generalization. \citet{chen2023skill} introduce a framework for ordered skill learning, optimizing data selection based on how effectively it teaches interdependent skills for continual pretraining and fine-tuning regimes. Although effective, these techniques operate at the group level, whereas our work explores reweighting at the instance level, offering finer control over how individual samples are treated based on their loss values. Furthermore, we demonstrate that combining domain-level methods such as DoReMi \citep{xie2023doremi} or DoGE \citep{fan2023doge} with our instance-level reweighting methods results in improved performance across multiple domains.
Instance-level reweighting has been used in post-training settings of LLMs \citep{chen2024take,jiang2024importance}. \citet{jiang2024importance} boost the self-improvement abilities of LLMs by employing sample reweighting to filter out self-generated data that have correct answers but exhibit high distribution shifts. \cite{chen2024take} reweight individual samples during continual training/instruction-tuning to focus on medium-loss samples. In contrast, our work systematically studies the effects of various sample-level, loss-based reweighting strategies on the efficiency and effectiveness of LLMs \emph{pretraining}.
The approach in \cite{fan2023irreducible} offers a curriculum learning framework that prioritizes samples with a higher learnability score, which is precomputed using another auxiliary model similar to DoReMi and DoGE. While we do not explicitly address curriculum learning in this work, our re-weighting mechanisms naturally allow for implementing a form of loss-based curriculum learning algorithms without the need to train and store additional proxy models as in \cite{fan2023irreducible}. 

\textbf{Sample-level Re-weighting as generic ML solution.} 
Sample-level reweighting has been extensively explored in other machine learning areas. For image classification, \cite{loshchilov2015online}, \cite{jiang2019accelerating}, and \cite{katharopoulos2018not} introduce pioneering approaches that prioritize samples based on loss values or gradient norms with the goal of accelerating training speed. Although these methods emphasize the importance of selecting high-loss samples during training, they typically require additional forward/backward passes on each training sample and are primarily designed for multi-epoch training, limiting their applicability with the vast pretraining corpus used for LLMs. Using bilevel optimization, \citet{grangier2023adaptive} adapt the data distribution during training to focus more on relevant samples for a target data distribution. Similarly, \citet{ren2018learning} employ a meta-learning approach to adjust sample weights based on validation set performance, which excels in noisy and imbalanced datasets. In this work, we introduce and investigate various lightweight, loss-based reweighting techniques that add little to no computational overhead compared to uniform sampling and do not require any nested optimization routines, often arising with bilevel optimization and/or meta-learning.
Sample-level reweighting has also been explored in the context of adversarial machine learning \citep{zhang2020geometry,liu2021probabilistic,zeng2021adversarial,sow2023doubly}, domain adaptation \citep{jiang2007instance,fang2020rethinking}, data augmentation \citep{yi2021reweighting}, and imbalanced classification \citep{qi2021online,ren2018learning}. Our reweighting mechanisms also have the potential to be studied under these contexts, which we leave as future work.

\section{Preleminaries: Autoregressive language modeling \& Goal}
\vspace{-3mm}
Large language models (LLMs) are typically trained using autoregressive language modeling, where the objective is to predict the next token in a sequence given the previous tokens. Formally, given a sample sequence of tokens  $\mathbf{x} = (x_1, x_2, \dots, x_T)$, where $x_t$ represents the $t$-th token in the sequence, the LLM model parameterized by \( \theta \in \mathbb{R}^d\) is trained to maximize the likelihood of the sequence:
\begin{equation}
    P_{\textrm{LLM}}(\mathbf{x}; \theta) = \prod_{t=1}^T P(x_t | x_1, \dots, x_{t-1}; \theta). \nonumber
\end{equation}
where $P(x_t | x_1, \dots, x_{t-1}; \theta)$ denotes the conditional probability of generating token $x_t$ given $\theta$ and all previously seen tokens $x_1, \dots, x_{t-1}$. In practice, autoregressive LLMs typically compute the loss $f(\mathbf{x}; \theta)$ for sample $\mathbf{x}$ using the negative log-likelihood (NLL) of the predicted tokens. 
\begin{equation}
    f(\mathbf{x}; \theta) = -\log P_{\textrm{LLM}}(\mathbf{x}; \theta) = - \sum_{t=1}^T \log P(x_t | x_1, \dots, x_{t-1}; \theta). \nonumber 
\end{equation}
Based on these sample losses, conventional SGD-like algorithms will then update the model parameters with the average gradient over all samples in a batch $\mathcal{B}$ at each training step $t$, i.e., 
\begin{equation}
    \theta^{t+1} = \theta^{t} - \frac{\eta}{|\mathcal{B}|}\sum_{i \in \mathcal{B}} \nabla f(\mathbf{x}_i; \theta^{t}),
\end{equation}
which results in each sample contributing equally to the overall training update. 
While this approach enjoys simplicity, it inherently treats all data points as equally informative, regardless of their individual relevance or difficulty. This uniform treatment overlooks the nuanced diversity within the data, which is particularly critical for current LLMs trained on vast and heterogeneous web-scale datasets. In such scenarios, the ability to dynamically differentiate between high-value and lesser-value data samples at different training stages is paramount for optimizing training efficiency and model performance. 

In contrast, our work introduces and benchmarks a variety of simple, lightweight reweighting strategies that can be seamlessly integrated into existing training pipelines with little to no computational overhead. These strategies aim to address the inherent limitations of the standard uniform averaging method by assigning dynamic importance to individual data samples based on their loss values, thus providing a more refined control over the training process. 
Ultimately, our objective is twofold: (i) to improve the efficiency and effectiveness of LLM training by allowing the model to dynamically focus on more relevant or informative data at different training stages and (ii) to potentially provide a scalable, automated approach to online data selection during pretraining. This ambitiously seeks to reduce the need for extensive manual data curation and selection efforts, which are increasingly becoming infeasible given the growing size and diversity of pretraining corpora. We envision these reweighting strategies as a foundation for more efficient and adaptive data utilization in LLM pretraining, ultimately reducing the time, cost, and complexity associated with model pretraining. 

\section{Approach}
\subsection{Technical Challenge: Data Reweighting for LLMs Must Be Dynamic and Fully Online}
\vspace{-2mm}
The core of our framework is to \emph{dynamically} reweight individual training samples based on their importance at different training stages. Although sample importance seems static---intuitively, “garbage data is garbage data” regardless of
model or training stage---data reweighting should be dynamic. For instance, similar or duplicated samples in training corpora may be useful early in training, but should be deprioritized once the model captures their patterns. Similarly, a hard but useful sample could be assigned a larger weight, however, as the model learns it could be beneficial to reduce its weight. 
Based on these observations, we update the model at each step $t$ using 
\begin{equation} \label{eq:ours}
    \theta^{t+1} = \theta^{t} - \eta_t \sum_{i \in \mathcal{B}} w(\mathbf{x}_i; \theta^t) \nabla f(\mathbf{x}_i; \theta^{t}),
\end{equation}
where $\sum_{i \in \mathcal{B}} w(\mathbf{x}_i; \theta^t) = 1$ and the weight $w(\mathbf{x}_i; \theta^t)$ is dynamically assigned based on sample $\mathbf{x}_i$ and current model $\theta^{t}$. 
However, designing effective dynamic reweighting methods for LLM pretraining involves several inherent complexities. 
In LLM pretraining, each sample is typically seen only once due to the vast size of the datasets, which contrasts with problems where repeated exposure to data is common, such as in computer vision problems. This makes methods that depend on accumulating historical data, such as previous loss or gradient norms \citep{loshchilov2015online,jiang2019accelerating}, unsuitable. Furthermore, the sheer scale of these large corpus makes it computationally prohibitive to track and store sample-specific statistics, as it would significantly increase resource requirements. 
Additionally, approaches that rely on hold-out validation sets to compute sample weights, as in \cite{grangier2023adaptive}, are also irrelevant for pretraining a general-purpose LLM. 
These unique technical challenges necessitate the development of a fully online, lightweight reweighting approach that scales efficiently with the size of a modern LLM training corpus, which we discuss next.

\subsection{Loss-based Reweighting strategies}
\vspace{-2mm}
To address the aforementioned challenges, we propose a novel framework of loss-based sample reweighting strategies motivated by the following theorem. \Cref{thm1} characterizes the effects of
loss-based reweighting on the convergence bound of reweighted full gradient descent when the sample losses $f(\mathbf{x}_i; \theta^{t})$ are convex. The extension to the stochastic setting and analysis for other types of loss geometries is deferred to \Cref{sec:analysis}. Our full statement of \Cref{thm1} and detailed proofs for all theoretical statements can be found in the appendix. 

\begin{theorem} \label{thm1}
Consider $M$ data points and let each loss $f(\mathbf{x}_i; \cdot)$ be convex. Further, assume the interpolation regime holds, i.e., $\exists \theta^* \in \mathbb{R}^d$ such that $\theta^* \in \arg\min_{\theta \in \mathbb{R}^d} f(\mathbf{x}_i; \theta) ~ \forall i$. 
Then, for a reweighting scheme that satisfies $\max_{i \in \{1,..., M\}} w(\mathbf{x}_i; \theta^t) \leq 2/M$, we have
\begin{align}
   \frac{1}{M} \sum_{i=1}^M f(\mathbf{x}_i; \bar \theta^{T}) - \frac{1}{M} \sum_{i=1}^M f(\mathbf{x}_i; \theta^{*}) \leq & ~ \mathcal{O}\left(\frac{\big\| \theta^{0} - \theta^*\big\|^2}{T}\right) + \frac{1}{T}\sum_{t=0}^{T-1} \delta^t, \label{eq:thm1}
\end{align}
where $\delta^t = \sum_{i=1}^{M} \left(\frac{1}{M} - w(\mathbf{x}_i; \theta^t)\right)\left(f(\mathbf{x}_i; \theta^{t}) - f(\mathbf{x}_i; \theta^{*})\right)$ and $\bar \theta^T = \frac{1}{T}\sum_{t=0}^{T-1}\theta^t$. 
\end{theorem}

\Cref{thm1} illustrates how setting the importance weight $w(\mathbf{x}_i; \theta^t)$ based on the loss gap $f(\mathbf{x}_i; \theta^{t}) - f(\mathbf{x}_i; \theta^{*})$ influences the convergence bound. 
Specifically, we make the following key observations: 
(i) fixing  $w(\mathbf{x}_i; \theta^t) = \frac{1}{M} ~ \forall i$, leads to $\delta_t = 0$ and we recover the traditional convergence bound of gradient descent for convex functions; (ii) assigning larger weights to the smaller loss gaps leads to $\delta_t \geq 0$, which yields a looser convergence bound; (iii) to achieve a better bound ($\delta_t \leq 0$), one should down-weight the importance of samples with low loss gaps. Further, note that \Cref{eq:thm1} critically holds only for reweighting schemes satisfying $w(\mathbf{x}_i; \theta^t) \leq 2/M$, which puts an upper bound on the maximum weight that can be assigned to a single data point, and hence, eliminates worst-case importance assignment strategies---such as those based on distributionally robust optimization \citep{qian2019robust,qi2021online,kumar2023stochastic}---which are prone to overfitting to outliers, a particularly problematic issue when dealing with the large-scale, noisy corpora used in LLM pretraining. \textcolor{black}{Finally, we also note that $\max_{i \in \{1,..., M\}} w(\mathbf{x}_i; \theta^t) \leq 2/M$ essentially means that the weight of any sample (after reweighting) should not be more than twice the uniform weight.} 
This upper bound on the sample weights is directly captured by our convergence analysis. Please see \Cref{sec:discussion_upperbound} for discussions on how it can be satisfied by our proposed reweighting methods.  

\begin{figure}
    \centering
    \begin{tabular}{cc}
    \includegraphics[width=6.5cm,height=3.2cm]{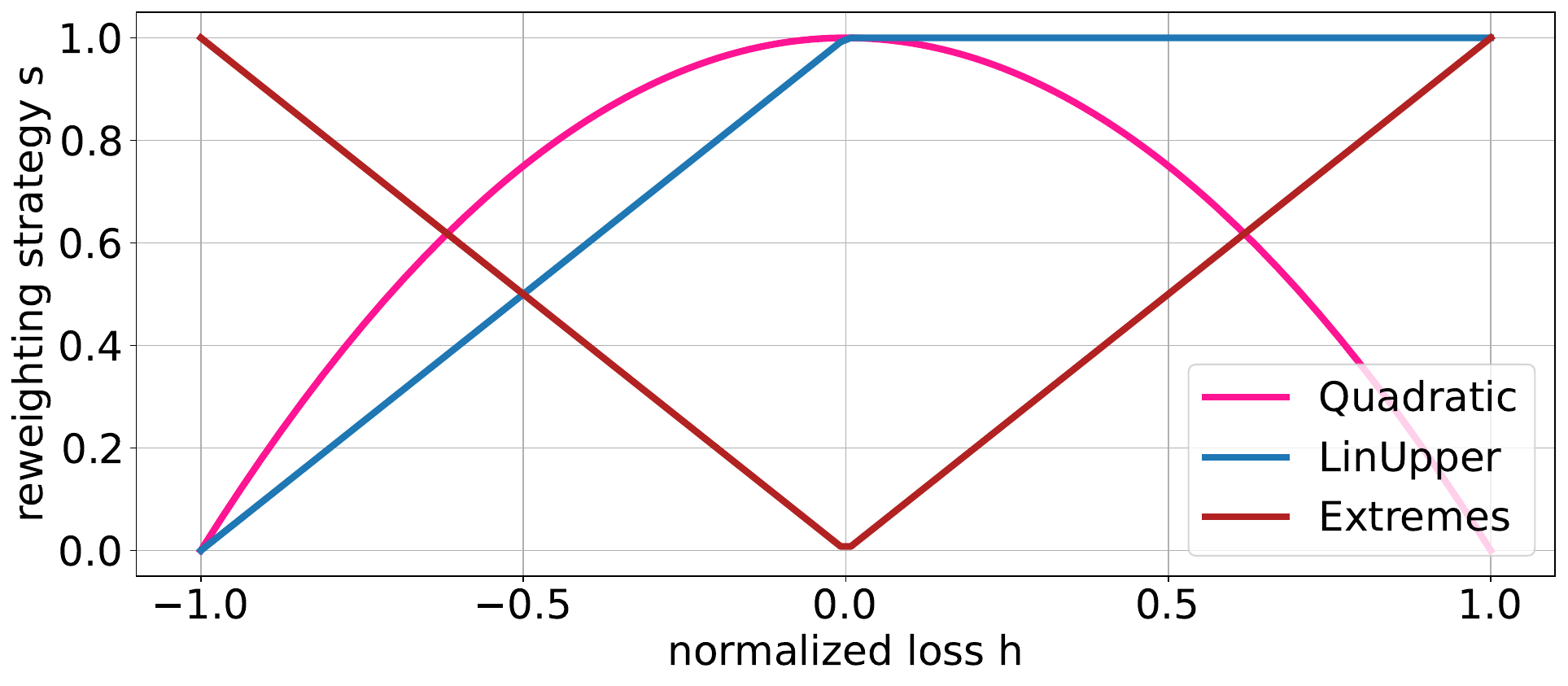}
    &\includegraphics[width=6.5cm,height=3.5cm]{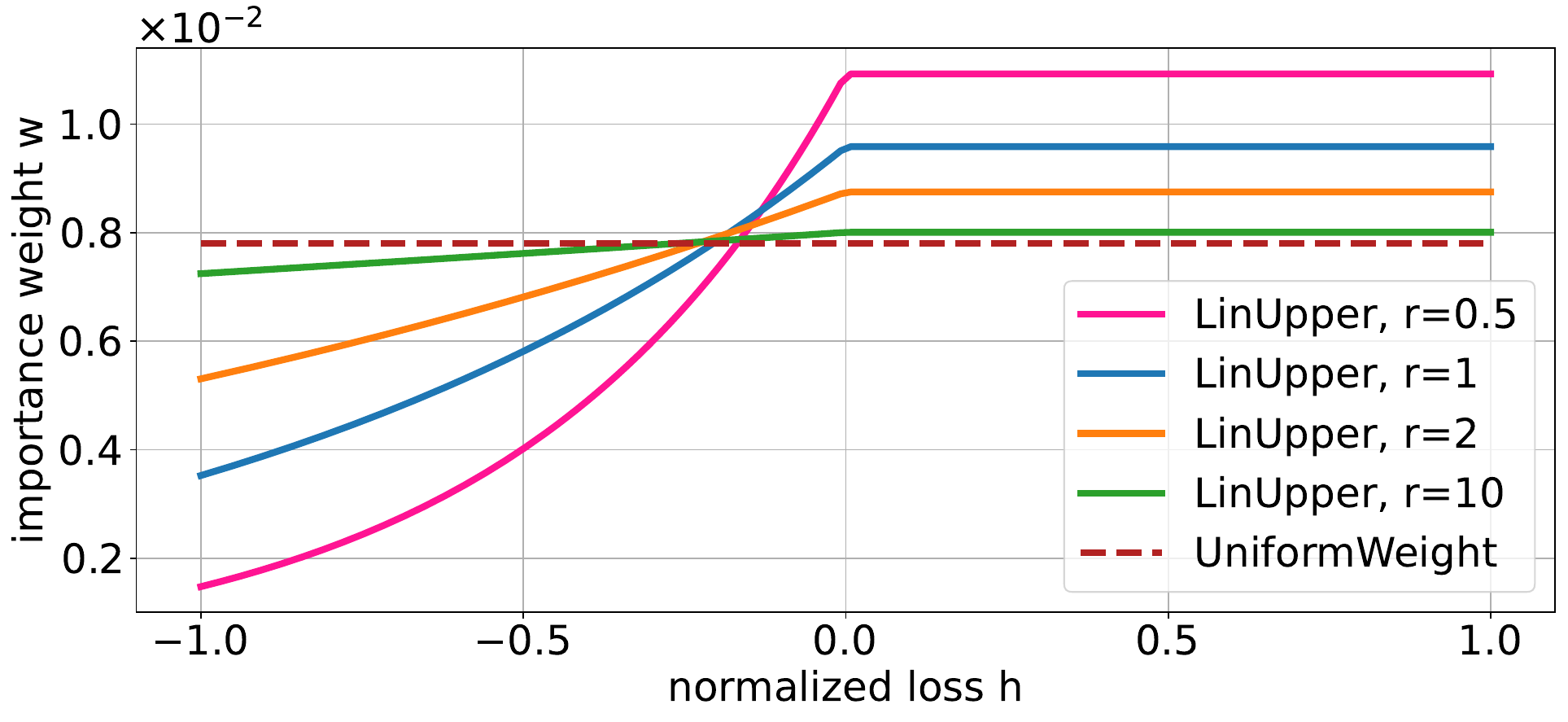}
    \end{tabular}
    \caption{\textbf{Left:} Geometric curves of the different reweighting functions. \textbf{Right:} Shape of the \texttt{LinUpper} strategy after applying \Cref{eq:softm} on top of it for different values of $r$. These plots are obtained for a batch of 128 uniformly drawn losses. As $r$ increases, \texttt{LinUpper} converges to the uniform averaging method.}
    \label{fig:rw_func}
\end{figure}

Based on the insights from \Cref{thm1} on the relationship between loss and importance weight, we study a series of reweighting strategies, 
each carefully designed to focus the learning dynamics on different parts of the loss distribution during training. 
Let $w_i$ be the importance weight for sample $\mathbf{x}_i$ where we drop the dependency on $\theta^t$ for clarity. We first normalize the individual sample losses into a bounded range $[-\alpha, \alpha]$. This step crucially ensures the loss values are scaled consistently and amenable to the subsequent weighting functions. We then apply the following analytical functions on the normalized loss $h_i$. \Cref{fig:rw_func} shows the geometric curves of these different functions. 
\begin{itemize}
    \item \textbf{Linear Upper-Bound Strategy (LinUpper).} In this strategy, the weight is proportional to the normalized loss but is capped at a predefined $\alpha$ value, ensuring that outliers do not dominate the training process. The functional form is $s_i := \min\{h_i + \alpha, \alpha\}$. 
    \item \textbf{Quadratic Strategy (Quadratic).} This strategy prioritizes samples with moderate loss values while down-weighting both low-loss (easy or repetitive) and high-loss (potential outliers) samples by applying a quadratic function $s_i := \alpha\left(1 - \frac{h_i^2}{\alpha^2}\right)$. 
    \item \textbf{Extremes-Based Strategy (Extremes).} This scheme emphasizes both the hardest (high-loss) and the easiest (low-loss) samples by applying $s_i := |h_i|$, ensuring that samples at the extremes of the loss distribution receive higher importance weights. 
\end{itemize}

To further enhance training stability and effectiveness, we add a curriculum-based adjustment mechanism on top of the strategic sample weight $s_i$ by controlling the sharpness of the importance weighting using a temperature parameter $r$. The final weight $w_i$ is computed as: 
\begin{equation}\label{eq:softm}
    w_i = \frac{\textrm{e}^{s_i / r}}{\sum_j \textrm{e}^{s_j / r}},
\end{equation}
where $r$ is annealed during training. Early in training, when the model is still learning fundamental patterns and losses are less informative, we employ a more uniform weighting scheme to ensure diverse feature learning by adopting a large $r$ (see \Cref{fig:rw_func}). As training progresses, we decrease the value of $r$ to enact full use of our reweighting strategies. 
\Cref{alg:foir} depicts our approach. 
It is important to note that our reweighting approach adds nearly zero computational overhead. During the forward pass, we already compute the sample losses, which are then used to derive the importance weights through simple analytical transformations. Since no extra loss or gradient computations are required, our reweighting strategies can be seamlessly integrated into standard training pipelines without incurring significant additional costs in terms of time or resources.

\begin{algorithm}[h]
	\caption{Fully Online Instance Reweighting}
	\small
	\label{alg:foir}
	\begin{algorithmic}[1]
		\STATE {\bfseries Input:} stepsize $\eta$, initial model parameters $\theta_0 \in \mathbb{R}^d$, temperature $\{r_t\}_{t=0}^T$, $f_{\min}$ and $f_{\max}$. 
		\FOR{$t=0,1,2,...,T-1$}
		\vspace{0.05cm}
		\STATE{Draw a minibatch of data samples $\mathcal{B} = \{\mathbf{x}_i\}_{i=1}^b$}
        \vspace{0.05cm}
        \STATE{Run forward pass to compute losses $\{f_{i,t}\}_{i=1}^b$ for all samples in $\mathcal{B}$} 
        \vspace{0.05cm}
        \STATE{Normalize losses into interval $[-\alpha, \alpha]$ using linear transform: $h_{i,t} = \frac{2\left(f_{i,t} - f_{\min}\right)}{f_{\max} - f_{\min}} - 1$}
        \vspace{0.05cm}
        \STATE{Apply reweighting strategy to normalized losses to obtain $\{s_{i,t}\}_{i=1}^b$} 
        // \{LinUpper, Quadratic, Extremes\}
        \vspace{0.05cm}
        \STATE{Apply curriculum adjustment using \Cref{eq:softm} to obtain sample weights $\{w_{i,t}\}_{i=1}^b$}
        \vspace{0.05cm}
        \STATE{Update $\theta^{t+1} = \theta^{t} - \eta \sum_{i \in \mathcal{B}} w_{i,t} \nabla f_{i,t}$}
        \vspace{0.05cm}
        \ENDFOR
	\end{algorithmic}
\end{algorithm}

\textbf{Optimal strategy minimizing $\delta^t$.} We next demonstrate that applying \Cref{eq:softm} on top of our strategy \texttt{LinUpper}, in fact, corresponds to the optimal strategy that minimizes a KL-divergence regularized version of $\delta^t$ in \Cref{thm1}. 

\begin{restatable}{proposition}{weightbound}
\label{prop:prop1}
The optimal strategy that minimizes the KL-divergence regularized $\delta^t$, 
i.e., $\delta^t + r \sum_{i=1}^M w_i \log (Mw_i)$, is given by
    \begin{align}
    \label{eq:weight_bound_eq}
   w_i = C \min\left\{\exp\left(\frac{h_i}{r}\right),\frac{2}{M}\right\},
\end{align}
where $C$ is a normalizing constant that ensures $\sum_i w_i = 1$. 
\end{restatable}
\begin{proof}
    Proof relegated to \Cref{sec:optimal_weights}.
\end{proof}
\vspace{-2mm}

As shown in \Cref{thm1} and \Cref{prop:prop1}, our experiments confirm that the \texttt{LinUpper} reweighting scheme indeed achieves faster convergence and improved performance across a spectrum of LLMs pretraining scales. The incorporation of KL-divergence regularization in the optimization of $\delta^t$ prevents extreme solutions that focus only on the highest-loss samples. Without this regularization term, the optimal solution becomes trivial: it assigns a weight of $\frac{2}{M}$ to the $\frac{M}{2}$ samples with the highest losses and zero weight to the remaining samples. 
{\color{black} It is important to note that such a strategy discards half of the dataset and performs poorly in practice. 
However,  our \texttt{LinUpper} method offers a balance between focusing on high-loss samples and maintaining the diversity of data utilized during training, leading to faster convergence and performance improvements.} 

\section{Analysis of Loss-based Reweighting on Convergence Bound}\label{sec:analysis}
\vspace{-2mm}

In this section, we analyze the effects of loss-based reweighting on the convergence bound in the more generic setting of stochastic gradient methods possibly with momentum. Consider the problem of minimizing a finite sum of $M$ objective functions:
\begin{equation}
\min_{\theta \in \mathbb{R}^d} \Bigg\{f(\theta) := \frac{1}{M} \sum_{i=1}^M f_i(\theta)\Bigg\}
\end{equation}
where each $f_i: \mathbb{R}^d \rightarrow \mathbb{R}$ corresponds to a sample loss function $f(\mathbf{x}_i; \cdot)$. We make the following widely adopted assumptions:

\begin{assum}[Convexity]\label{ass:conv} 
    Each function $f_i$ is convex, i.e., for all $\theta, \theta' \in \mathbb{R}^d$ and $i \in {1, \ldots, M}$:
    \begin{equation}
    f_i(\theta') \geq f_i(\theta) + \langle \nabla f_i(\theta), \theta' - \theta \rangle. 
    \end{equation}
\end{assum}

\begin{assum}[$L$-smoothness]\label{ass:lip} 
    Each function $f_i$ is $L_i$-smooth. Formally, for all $\theta, \theta' \in \mathbb{R}^d$ and $i \in {1, \ldots, M}$:
    \begin{equation}
    \big\|\nabla f_i(\theta') - \nabla f_i(\theta)\big\| \leq L_i \big\|\theta' - \theta\big\|, 
    \end{equation}
    and $L=\max_{i\in [M]}L_i$.
\end{assum}

\begin{assum}[Interpolation condition] \label{ass:interp}
There exists a global optimum $\theta^* \in \mathbb{R}^d$ such that: 
\begin{equation}
\theta^* \in \arg\min_{\theta \in \mathbb{R}^d} f_i(\theta), \quad \forall i \in \{1, \ldots, M\}.
\end{equation}
\end{assum}

\Cref{ass:interp} is widely adopted in modern ML settings \citep{dar2021farewell,radhakrishnan2020overparameterized,jhunjhunwala2023fedexp}, particularly when analyzing overparameterized deep neural networks. This ensures that all individual sample objectives can be minimized simultaneously, effectively eliminating adversarial trade-offs where minimizing one objective could worsen another. 

Next, we characterize the effects of
loss-based reweighting on the convergence bound of reweighted minibatch SGD when the sample losses $f_i$ are convex. {{For ease of notation, we denote $w(\mathbf{x}_i; \theta^t)$ by $w_{i,t}$ in this section and the corresponding proofs in the Appendix.}} 

\begin{restatable}[Minibatch SGD with momentum]{theorem}{minibatchSGDconvex}
\label{thm:minibatch_SGD_convex}
Let Assumptions 
\ref{ass:conv}, \ref{ass:lip}, and \ref{ass:interp} hold. Consider a minibatch of size $|\mathcal{B}|=b$ with reweighting scheme satisfying $\max_{i\in \mathcal{B}}w_{i,t}\leq 2/b$. Then, 
for $\eta = \frac{1}{8L\sqrt{T+1}}, \lambda_t>0$, we have 
    \begin{align}
    \label{eq:minibatch_momentum_convex_bound}
  \mathbb{E}[f( \theta^T) - f(\theta^*)] &\leq\frac{8L\big\| \theta^{0} - \theta^*\big\|^2}{\sqrt{T+1}} +\frac{2}{{T+1}}\sum_{t=0}^{T-1}{\delta_t} + \frac{2}{{T+1}}\sum_{t=0}^{T-1}{\lambda_t\mu_{t}},
\end{align} 
where $\delta_t=\mathbb{E}\left[ \sum_{i\in \mathcal{B}} \left(\frac{1}{b}-w_{i,t}\right) ( f_i(\theta^{t})-f_i(\theta^*))\right]$, $\mu_t=\mathbb{E}\big[ \sum_{i\in \mathcal{B}} \left(\frac{1}{b}-w_{i,t}\right) ( f_i(\theta^{t})-f_i(\theta^{t-1}))\big]$, and $\bar \theta^T = \frac{1}{T}\sum_{t=0}^{T-1}\theta^t$.
\end{restatable}
\begin{proof}
    Proof provided in \Cref{sec:convex_minibatch}.
\end{proof}

We note similar observations as in \Cref{thm1} and again down-weighting the importance of samples with low loss gaps leads to a better bound. 
Typically, the minibatch gradient with uniform weights is unbiased, $\mathbb{E}\left[\sum_{i\in \mathcal{B}}\frac{1}{b}\nabla f_i(\theta^t)\right]=\nabla f(\theta^t)$. However, in our setting, the importance weights are functions of the sample functions, hence, leading to biased gradients and the analysis of \Cref{thm:minibatch_SGD_convex} being more involved. 
Further, we provide the analysis for minibatch SGD when the sample functions are non-convex in \Cref{sec:nonconvex_minibatch}. 

\section{Experimental Evaluation}
\vspace{-2mm}

\subsection{Training and Evaluation Setup}
\vspace{-2mm}
We conduct our first set of experiments using decoder-only transformer models \citep{vaswani2017attention,radford2019language} with parameter sizes of 120M, 210M, and 300M, which we refer to as, respectively, GPT2-mini, GPT2-small, and GPT2-medium. We train on the SlimPajama \citep{cerebras2023slimpajama} corpus that includes seven diverse domains: Common Crawl (CC), C4, GitHub, StackExchange, Book, Arxiv, and Wikipedia. 
In the first step, we pretrain the three GPT2 models on all seven domains where we compare our sample-level reweighting methods (\texttt{LinUpper}, \texttt{Quadratic}, \texttt{Extremes}) against the uniform averaging baseline in which each sample contributes equally. 
In a second step, we assess the impact of combining our reweighting techniques with existing domain-level reweighting methods, specifically DoGE and DoReMi, to demonstrate the potential for further performance gains when fine-grained sample reweighting is added on top of domain-level adjustments. 
Then, we conduct experiments of pretraining larger models with 1.4B and 7B parameters (\Cref{sub:llamaexp}). We also provide results for a regression problem on synthetic data in \Cref{app:toy}, to validate the fundamental advantage of reweighting. 
Model architecture details can be found in \Cref{app:arch}. 
Our codebase for the GPT2 experiments is publicly available.\footnote{\url{\codebase}.}

We measure performance using two types of metrics: 

\textbf{Perplexity on Training Distribution.} 
We evaluate per-domain perplexity and average perplexity on all training domains using hold-out validation sets. 
By doing so, we aim to generate insights on the sample efficiency during pretraining, i.e., how well a model can learn from a given dataset and adapt to the data distribution.

\textbf{Few-Shot Reasoning Ability.} To assess generalization beyond the pretraining distribution, we evaluate models on a series of diverse reasoning tasks in a 5-shot setting, including 
LogiQA \cite{liu2020logiqa}, 
LogiQA 2 \cite{liu2023logiqa},
SciQ \cite{welbl-etal-2017-crowdsourcing}, 
and PiQA \cite{bisk2020piqa}.

\subsection{Comparisons Under Uniform Domain Sampling}
\begin{figure}
    \centering
    \begin{subfigure}{0.24\textwidth}
        \includegraphics[width=\linewidth]{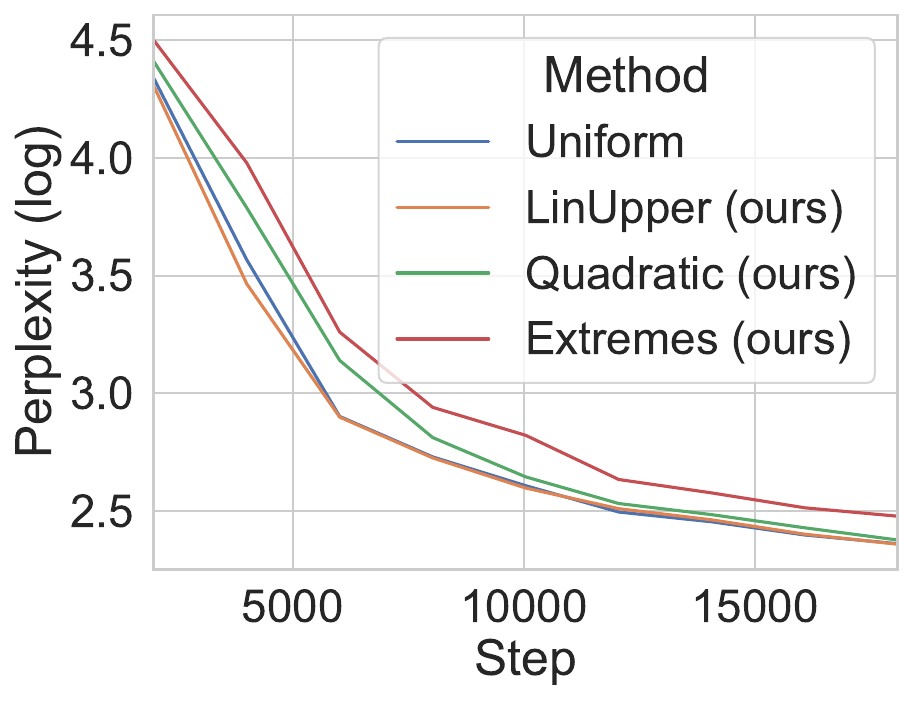}
        \caption{Arxiv Dataset}
        \label{fig:gpt2medium_arxiv_dataset}
    \end{subfigure}
    \begin{subfigure}{0.24\textwidth}
        \includegraphics[width=\linewidth]{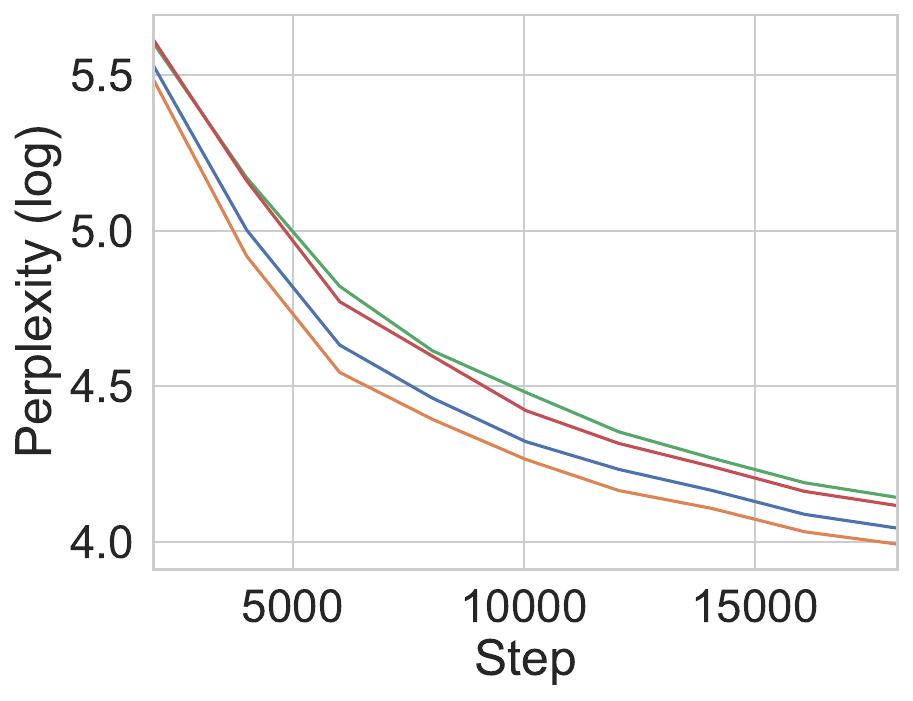}
        \caption{Book Dataset}
        \label{fig:gpt2medium_book_dataset}
    \end{subfigure}
    \begin{subfigure}{0.24\textwidth}
        \includegraphics[width=\linewidth]{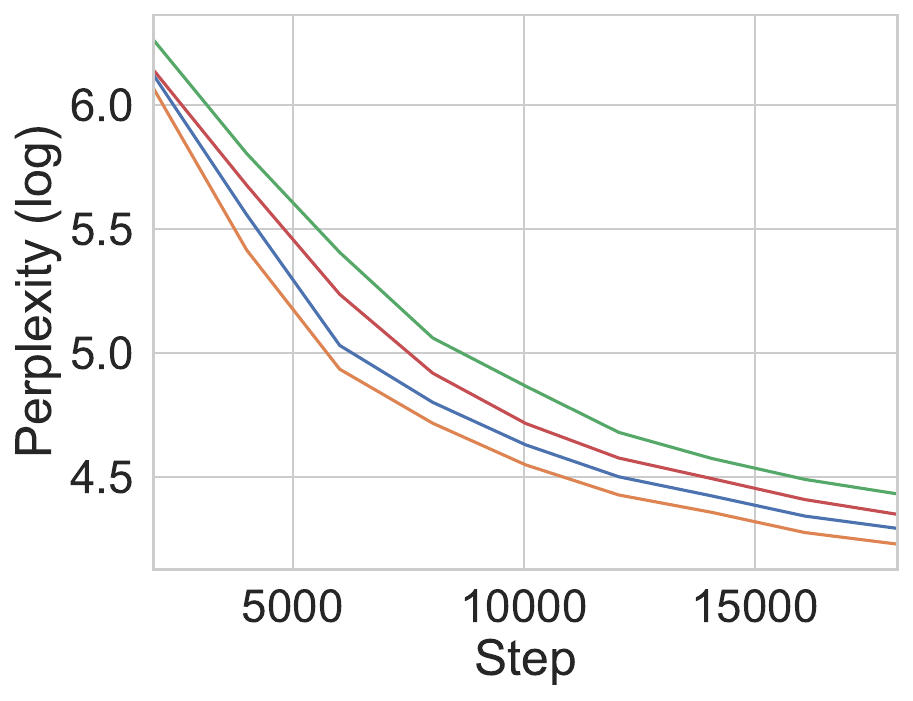}
        \caption{C4 Dataset}
        \label{fig:gpt2medium_c4_dataset}
    \end{subfigure}
    \begin{subfigure}{0.24\textwidth}
        \includegraphics[width=\linewidth]{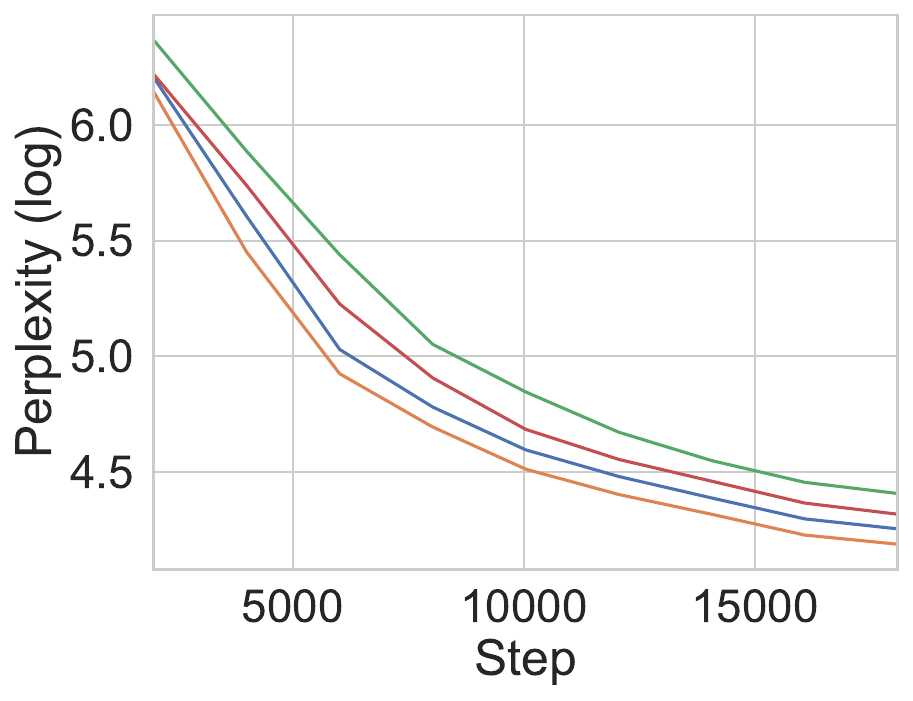}
        \caption{CC Dataset}
        \label{fig:gpt2medium_cc_dataset}
    \end{subfigure}
    \begin{subfigure}{0.24\textwidth}
        \includegraphics[width=\linewidth]{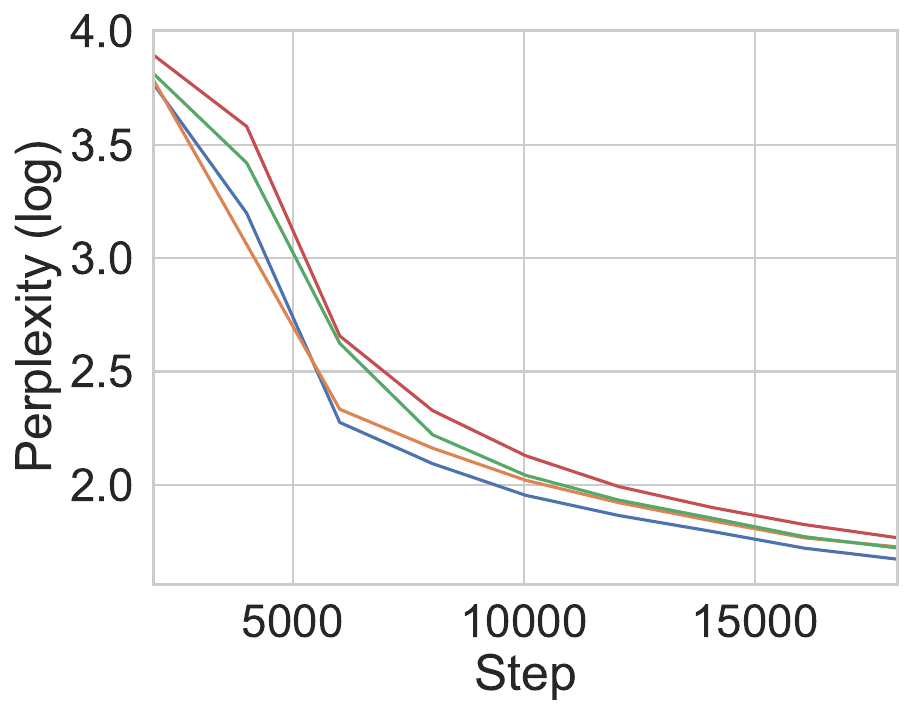}
        \caption{GitHub Dataset}
        \label{fig:gpt2medium_github_dataset}
    \end{subfigure}
    \begin{subfigure}{0.24\textwidth}
        \includegraphics[width=\linewidth]{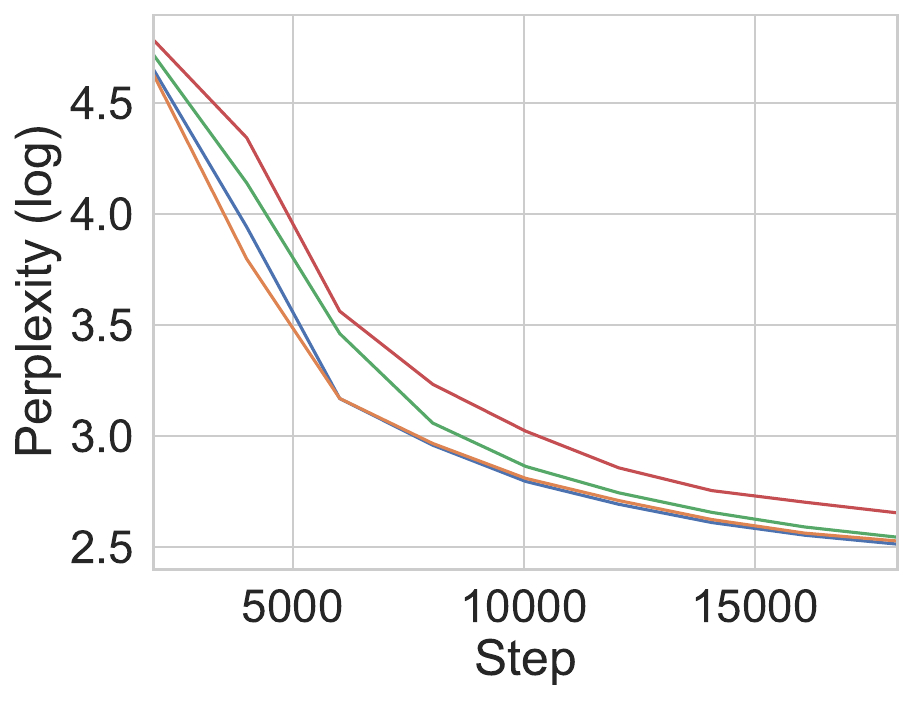}
        \caption{StackExchange Dataset}
        \label{fig:gpt2medium_stackexchange_dataset}
    \end{subfigure}
    \begin{subfigure}{0.24\textwidth}
        \includegraphics[width=\linewidth]{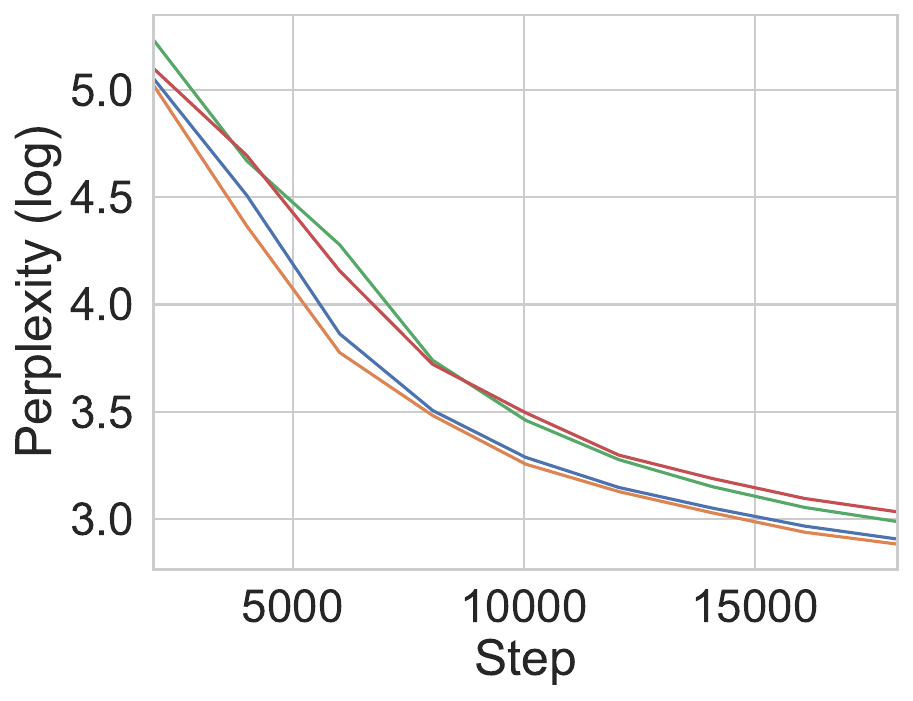}
        \caption{Wikipedia Dataset}
        \label{fig:gpt2medium_wikipedia_dataset}
    \end{subfigure}

    \caption{Per-domain perplexities on hold-out validation sets under the uniform domain sampling setting for the GPT2-medium model. Our reweighting strategy \texttt{LinUpper} strategy achieves better or at least comparable perplexity on 5 out of 7 domains. 
    }
    \label{fig:perplexity_results}
\end{figure}

\vspace{-2mm}
\textbf{Comparisons on Pretrainining Distribution.}
\Cref{fig:perplexity_results} shows the per-domain perplexity for the GPT2-medium model on all seven domains. We provide the final per-domain perplexity results for the GPT2-mini and GPT2-small models in Table \ref{tab:gpt2_perplexity_all_results}. The perplexity plots for GPT2-mini and GPT2-small can be found in \Cref{sec:apdix_experimental_results}. 
\Cref{fig:perplexity_results} shows that our proposed \texttt{LinUpper} strategy outperforms or matches the other baseline in 5 out of 7 domains, with particularly notable improvements in CC, C4, and Book domains. \texttt{LinUpper} achieves significantly lower perplexity in these domains, demonstrating its effectiveness at handling noisier data sources. This suggests that down-weighting low-loss samples --central to \texttt{LinUpper}-- helps reduce the effects of redundant or similar data, and thus accelerates convergence as supported by our theoretical findings. 
We also note competitive results for \texttt{LinUpper} method at the smaller scale models in Table \ref{tab:gpt2_perplexity_all_results}, but the advantage is more significant in the larger GPT2-medium model.

\begin{table}[hbt!]
    \centering
    \caption{Per-domain perplexities on hold-out validation datasets under the \textit{uniform} domain sampling setting for the GPT2 models. Across a wide variety of domains, our \texttt{LinUpper} method outperforms the uniform baseline.}
    \label{tab:gpt2_perplexity_all_results}
    \resizebox{0.90\textwidth}{!}{
        \begin{tabular}{ll|rrrrrrr|r}
            \toprule
                    &       & \multicolumn{7}{c|}{Benchmark} & \\
             Model & Method & Arxiv & Book & C4 & CC & GitHub & StackExchange & Wikipedia & \textit{Mean} \\
            \midrule
            \multirow{4}{*}{GPT2-mini} & Uniform            & \textbf{2.48}     & 4.21          & 4.47          & 4.44 & \textbf{1.84} & \textbf{2.66} & 3.14 & 3.32 \\
                                       & LinUpper (ours)    & 2.49              & \textbf{4.16} & \textbf{4.41} & \textbf{4.38} & 1.89 & 2.67 & \textbf{3.12} & \textbf{3.30} \\
                                       & Quadratic (ours)   & \textbf{2.48}     & 4.26 & 4.55                   & 4.53 & 1.87 & 2.67 & 3.17 & 3.36 \\
                                       & Extremes (ours)    & 2.54              & 4.23 & 4.47                   & 4.44 & 1.90 & 2.74 & 3.19 & 3.36 \\
            \midrule
            \multirow{4}{*}{GPT2-small} & Uniform & \textbf{2.36} & 4.04 & 4.29 & 4.26 & \textbf{1.67} & \textbf{2.51} & 2.90 & 3.15 \\
                                &LinUpper (ours) & \textbf{2.36} & \textbf{3.99} & \textbf{4.23} & \textbf{4.19} & 1.73 & 2.53 & \textbf{2.88} & \textbf{3.13} \\
                                &Quadratic (ours) & 2.38 & 4.14 & 4.43 & 4.41 & 1.72 & 2.54 & 2.99 & 3.23 \\
                                &Extremes (ours) & 2.48 & 4.12 & 4.35 & 4.32 & 1.77 & 2.65 & 3.03 & 3.24 \\
            \midrule
            \multirow{4}{*}{GPT2-medium} & Uniform   & \textbf{2.34} & 4.02          & 4.27              & 4.22 & \textbf{1.66} & \textbf{2.49} & 2.88 & 3.13 \\
                                 &LinUpper (ours)   & \textbf{2.34} & \textbf{3.96} & \textbf{4.20}     & \textbf{4.15} & 1.72 & 2.50 & \textbf{2.85} & \textbf{3.10} \\
                                 &Quadratic (ours)  & 2.35          & 4.11          & 4.40              & 4.36 & 1.70 & 2.51 & 2.93 & 3.19 \\
                                 &Extremes (ours)   & 2.45          & 4.08          & 4.31              & 4.27 & 1.76 & 2.62 & 3.00 & 3.21 \\
            \bottomrule
        \end{tabular}
    }
\end{table}

\textbf{Few-Shot Evaluations.}
\Cref{tab:nodr_benchmar_results} provides the 5-shot evaluations on 4 different tasks for the GPT2-medium model. 
The \texttt{LinUpper} reweighting strategy consistently yields competitive results, outperforming the other compared baselines on 3 out of 4 tasks. 
These results further validate the general effectiveness of \texttt{LinUpper} across different types of reasoning challenges, including logical inference (LogiQA), physical commonsense reasoning (PiQA), and scientific reasoning (SciQ). Notably, \texttt{Quadratic} performs best in LogiQA 2 (28.6\%), indicating that reducing the effects of both high-loss (potential outliers) and low-loss (potentially redundant data) samples can sometimes improve generalization performance even though it generally does not achieve best efficiency for learning the pretraining distribution. This suggests that the \texttt{Quadratic} scheme could be useful when training under noisier and less curated datasets. Also note that the \texttt{Extremes} strategy consistently lags behind the other methods, which further confirms the advantage of reducing the importance of low-loss samples. 

\begin{table}[bt]
    \centering
    \caption{5-shot Performance evaluations under \textit{uniform} domain sampling for the GPT2-medium model. We report the accuracy for every benchmark. The highest accuracy per benchmark is highlighted in \textbf{bold}. 
    The \texttt{LinUpper} reweighting strategy consistently yields competitive results, outperforming the other compared baselines on 3 out of 4 tasks. 
    }
    \label{tab:nodr_benchmar_results}
    \resizebox{0.75\textwidth}{!}{
        \begin{tabular}{l|rrrr}
            \toprule
                      & \multicolumn{4}{c}{Method} \\
            Benchmark Accuracy & Uniform & LinUpper (ours) & Quadratic (ours) & Extremes (ours) \\
            \midrule
            LogiQA & 25.7 & \textbf{27.9} & 25.5 & 26.5 \\
            LogiQA 2 & 27.5 & 27.6 & \textbf{28.6} & 27.7 \\
            SciQ & 49.0 & \textbf{51.8} & 48.7 & 49.6 \\
            PiQA & 55.5 & \textbf{56.2} & 54.6 & 55.2 \\
            \midrule
            \textit{Mean} & 39.4 & \textbf{40.9} & 39.3 & 39.8 \\
            \bottomrule
        \end{tabular}
    }
\end{table}

\begin{table}[t]
    \centering
    \caption{5-shot performance evaluations under \emph{non-uniform} domain sampling for the GPT2-medium model. Our \texttt{LinUpper} instance reweighting approach creates synergies with state-of-the-art domain reweighting techniques.}
    \label{tab:domain_reweighting_comparison}
    \resizebox{0.75\textwidth}{!}{
        \begin{tabular}{l|rr|rr}
            \toprule
                 & \multicolumn{4}{c}{Method} \\
                 Benchmark Accuracy & DoGE & DoGE + LinUpper & DoReMi & DoReMi + LinUpper \\
            \midrule
            LogiQA  & 27.2 & \textbf{28.6} & 27.2 & \textbf{27.6} \\
            LogiQA 2 & 27.5 & \textbf{28.0} & 27.7 & \textbf{27.9} \\
            SciQ  & 52.8 & \textbf{53.2} & 53.3 & \textbf{54.5} \\
            PiQA  & 55.8 & \textbf{56.3} & 55.7 & \textbf{56.1} \\
            \midrule
            \textit{Mean} & 40.8 & \textbf{41.5} & 41.0 & \textbf{41.5} \\
            \bottomrule
        \end{tabular}
    }
\end{table}

\subsection{Comparisons Under Non-Uniform Domain Sampling}
\vspace{-2mm}
The performance of the compared methods under non-uniform domain sampling setting are presented in the \Cref{tab:domain_reweighting_comparison}. The results
clearly demonstrate the value of incorporating our \texttt{LinUpper} strategy on top of existing domain-level reweighting methods such as DoGE and DoReMi. Across all tasks, \texttt{LinUpper} consistently improves the performance of both domain reweighting baselines, reinforcing the effectiveness of dynamically adjusting sample weights at the instance level in conjunction with domain-level reweighting. 
In particular, our reweighting strategy \texttt{LinUpper} provides notable improvements for LogiQA task, boosting accuracy from 27.2\% to 28.6\% with DoGE and from 27.2\% to 27.6\% with DoReMi. The gains in LogiQA 2 are similar, where our reweighting method \texttt{LinUpper} consistently improves the performance of both methods. 
For scientific reasoning (SciQ) and physical commonsense (PiQA) tasks, \texttt{LinUpper} strategy also enhances both baseline methods, with the most substantial improvement observed in SciQ (from 52.8\% to 53.2\% with DoGE and from 53.3\% to 54.5\% with DoReMi). These results indicate that \texttt{LinUpper} is particularly effective in tasks requiring nuanced reasoning and factual knowledge. 

{\color{black}
\subsection{Experiments on 1.4B and 7B Llama Models} \label{sub:llamaexp}
We conduct experiments to train 1.4B (billion) and 7B parameter models with Llama architecture on randomly sampled subsets of the FineWeb 15T dataset. 
For the 1.4B and 7B parameter models, we use approximately 100B and 175B randomly sampled tokens from the FineWeb 15T dataset, respectively. We report the benchmark results on the last checkpoints. 

For these larger models, we use the question-answering (QA) benchmarks and additionally employ language understanding and reasoning (LUR) tasks since their capacity allows for more complex tasks. Please see Appendix A for more details, including hyperparameters, model architectural details, and performance on the full list of benchmarks.

\Cref{tab:llama1.4b_benchmark_avg,tab:llama7b_benchmark_avg} show the average performance among LUR and QA tasks for the 1.4B and 7B parameter models, respectively. For the 1.4B parameter model, we see that our \texttt{LinUpper} strategy improves the average model performance by $0.66\%$ and $1.72\%$ for LUR and QA tasks, respectively. 
In particular, we note that our \texttt{LinUpper} method improves performance for 11 out of 19 tasks in total (see \Cref{tab:llama1.4b_benchmark_results_lur,tab:llama1.4b_benchmark_results_qa} in \Cref{sec:apdix_experimental_results}). For the 7B parameter model, our \texttt{LinUpper} method improves results for 14 out of 19 tasks (see \Cref{tab:llama7b_benchmark_results_lur,tab:llama7b_benchmark_results_qa} in \Cref{sec:apdix_experimental_results})  with an average performance gain of $1.44\%$ among LUR tasks and $1.16\%$ among QA tasks. 
These improvements are particularly noteworthy given the scale of these models, highlighting the scalability and effectiveness of our method. 

\begin{table}
        \parbox{.49\linewidth}{
        \centering
        \color{black}
        \caption{\textcolor{black}{Average performance of the 1.4B parameter model across LUR and QA benchmarks. Overall, our \texttt{LinUpper} method improves results for 11 out of 19 tasks with an average performance gain of $0.66\%$ among LUR tasks and $1.72\%$ among QA tasks.
        }}
        \label{tab:llama1.4b_benchmark_avg}
        \resizebox{0.48\textwidth}{!}{
            \begin{tabular}{lrrr}
                \toprule
                
                Avg. Performance & Uniform & \multicolumn{1}{c}{LinUpper (ours)} & Difference\\
                \midrule
                LUR (13 tasks) & 48.51 & \textbf{49.16} & 0.66 \\ 
                QA (6 tasks) & 42.35 & \textbf{44.07} & 1.72 \\ \midrule
                \textit{Across 19 tasks} & \textit{46.56} & \textit{\textbf{47.55}} & \textit{0.99} \\
                \bottomrule
            \end{tabular}
    }
    \vspace{-1.1\baselineskip}
    \vfill
    }
    \hfill
    \parbox{.49\linewidth}{
        \centering
        \color{black}
        \vspace{1.2\baselineskip}
        \caption{\textcolor{black}{Results of the 7B parameter model. Overall, our \texttt{LinUpper} method improves results for 14 out of 19 tasks with an average performance gain of $1.44\%$ among LUR tasks and $1.16\%$ among QA tasks.}}
        \vspace{1\baselineskip}
    
        \label{tab:llama7b_benchmark_avg}
        \resizebox{0.48\textwidth}{!}{
            \begin{tabular}{lrrr}
            \toprule
            
            Avg. Performance & Uniform & \multicolumn{1}{c}{LinUpper (ours)} & Difference\\
            \midrule
            LUR (13 tasks) & 50.54 & \textbf{51.98} & 1.44 \\
            QA (6 tasks) & 48.15 & \textbf{49.31} & 1.16 \\ \midrule
            \textit{Across 19 tasks} & \textit{49.79}& \textit{\textbf{51.14}} & \textit{1.35} \\
            \bottomrule
            \end{tabular}
        }
    }
\end{table}

}

\section{Conclusion}
\vspace{-2mm}
In this paper, we introduce a novel sample-level reweighting framework aimed at improving the efficiency and effectiveness of large language model (LLM) pretraining. By dynamically adjusting the importance of individual samples based on their loss values, our approach overcomes the limitations of traditional uniform averaging methods and adds a new dimension to existing domain-level reweighting methods by incorporating more fine-grained sample-level dynamics. Through extensive theoretical and empirical validation, we demonstrate that down-weighting low-loss samples accelerates convergence and improves performance. The experiments show that our proposed \texttt{LinUpper} strategy consistently outperforms uniform sampling on common LLM reasoning and commonsense benchmarks. 
Similarly, our approach creates synergies with existing domain reweighting techniques, which underpins the importance of sample-level dynamics.
We observe that the benefits of our method are more pronounced in larger models, while smaller models show less significant improvements on benchmarks. This could be due to the fact that smaller models may have limited capacity to fully exploit the advantages of our approach. 
However, our overall findings highlight the potential of loss-based reweighting strategies to optimize LLM pretraining both in training efficiency and in model evaluation performance. 

\clearpage

\subsubsection*{Acknowledgments}
We thank the IBM Data Engineering Team for their support in running experiments with 1.4B and 7B parameter models. 
The work of D. Sow and Y. Liang was supported in part by the U.S. National Science Foundation with the grants ECCS-2113860, ECCS-2413528, and CNS-2112471. 
The work of H. Woisetschl\"ager was partially supported by the Bavarian State Ministry of Economic Affairs, Regional Development and Energy with grant DIK0446/01.

\bibliographystyle{iclr2025_conference}
\bibliography{main}

\begin{thebibliography}{44}
\providecommand{\natexlab}[1]{#1}
\providecommand{\url}[1]{\texttt{#1}}
\expandafter\ifx\csname urlstyle\endcsname\relax
  \providecommand{\doi}[1]{doi: #1}\else
  \providecommand{\doi}{doi: \begingroup \urlstyle{rm}\Url}\fi

\bibitem[Achiam et~al.(2023)Achiam, Adler, Agarwal, Ahmad, Akkaya, Aleman, Almeida, Altenschmidt, Altman, Anadkat, et~al.]{achiam2023gpt}
Josh Achiam, Steven Adler, Sandhini Agarwal, Lama Ahmad, Ilge Akkaya, Florencia~Leoni Aleman, Diogo Almeida, Janko Altenschmidt, Sam Altman, Shyamal Anadkat, et~al.
\newblock Gpt-4 technical report.
\newblock \emph{arXiv preprint arXiv:2303.08774}, 2023.

\bibitem[Bisk et~al.(2020)Bisk, Zellers, Gao, Choi, et~al.]{bisk2020piqa}
Yonatan Bisk, Rowan Zellers, Jianfeng Gao, Yejin Choi, et~al.
\newblock Piqa: Reasoning about physical commonsense in natural language.
\newblock In \emph{Proceedings of the AAAI conference on artificial intelligence}, volume~34, pp.\  7432--7439, 2020.

\bibitem[Brown et~al.(2020)Brown, Mann, Ryder, Subbiah, Kaplan, Dhariwal, Neelakantan, Shyam, Sastry, Askell, et~al.]{brown2020language}
Tom Brown, Benjamin Mann, Nick Ryder, Melanie Subbiah, Jared~D Kaplan, Prafulla Dhariwal, Arvind Neelakantan, Pranav Shyam, Girish Sastry, Amanda Askell, et~al.
\newblock Language models are few-shot learners.
\newblock \emph{Advances in neural information processing systems}, 33:\penalty0 1877--1901, 2020.

\bibitem[Chen et~al.(2023)Chen, Roberts, Bhatia, WANG, Zhang, Sala, and Re]{chen2023skill}
Mayee~F Chen, Nicholas Roberts, Kush Bhatia, Jue WANG, Ce~Zhang, Frederic Sala, and Christopher Re.
\newblock Skill-it! a data-driven skills framework for understanding and training language models.
\newblock In \emph{Thirty-seventh Conference on Neural Information Processing Systems}, 2023.
\newblock URL \url{https://openreview.net/forum?id=IoizwO1NLf}.

\bibitem[Chen et~al.(2024)Chen, Wang, Sow, Yang, Chen, Liang, Zhou, and Wang]{chen2024take}
Xuxi Chen, Zhendong Wang, Daouda Sow, Junjie Yang, Tianlong Chen, Yingbin Liang, Mingyuan Zhou, and Zhangyang Wang.
\newblock Take the bull by the horns: Hard sample-reweighted continual training improves llm generalization.
\newblock \emph{arXiv preprint arXiv:2402.14270}, 2024.

\bibitem[Chowdhery et~al.(2023)Chowdhery, Narang, Devlin, Bosma, Mishra, Roberts, Barham, Chung, Sutton, Gehrmann, et~al.]{chowdhery2023palm}
Aakanksha Chowdhery, Sharan Narang, Jacob Devlin, Maarten Bosma, Gaurav Mishra, Adam Roberts, Paul Barham, Hyung~Won Chung, Charles Sutton, Sebastian Gehrmann, et~al.
\newblock Palm: Scaling language modeling with pathways.
\newblock \emph{Journal of Machine Learning Research}, 24\penalty0 (240):\penalty0 1--113, 2023.

\bibitem[Dar et~al.(2021)Dar, Muthukumar, and Baraniuk]{dar2021farewell}
Yehuda Dar, Vidya Muthukumar, and Richard~G Baraniuk.
\newblock A farewell to the bias-variance tradeoff? an overview of the theory of overparameterized machine learning.
\newblock \emph{arXiv preprint arXiv:2109.02355}, 2021.

\bibitem[Dubey et~al.(2024)Dubey, Jauhri, Pandey, Kadian, Al-Dahle, Letman, Mathur, Schelten, Yang, Fan, et~al.]{dubey2024llama}
Abhimanyu Dubey, Abhinav Jauhri, Abhinav Pandey, Abhishek Kadian, Ahmad Al-Dahle, Aiesha Letman, Akhil Mathur, Alan Schelten, Amy Yang, Angela Fan, et~al.
\newblock The llama 3 herd of models.
\newblock \emph{arXiv preprint arXiv:2407.21783}, 2024.

\bibitem[Fan \& Jaggi(2023)Fan and Jaggi]{fan2023irreducible}
Simin Fan and Martin Jaggi.
\newblock Irreducible curriculum for language model pretraining.
\newblock \emph{arXiv preprint arXiv:2310.15389}, 2023.

\bibitem[Fan et~al.(2023)Fan, Pagliardini, and Jaggi]{fan2023doge}
Simin Fan, Matteo Pagliardini, and Martin Jaggi.
\newblock {DOGE}: Domain reweighting with generalization estimation.
\newblock In \emph{Second Agent Learning in Open-Endedness Workshop}, 2023.
\newblock URL \url{https://openreview.net/forum?id=qiKqsqwYXm}.

\bibitem[Fang et~al.(2020)Fang, Lu, Niu, and Sugiyama]{fang2020rethinking}
Tongtong Fang, Nan Lu, Gang Niu, and Masashi Sugiyama.
\newblock Rethinking importance weighting for deep learning under distribution shift.
\newblock \emph{Advances in neural information processing systems}, 33:\penalty0 11996--12007, 2020.

\bibitem[Gao et~al.(2020)Gao, Biderman, Black, Golding, Hoppe, Foster, Phang, He, Thite, Nabeshima, et~al.]{gao2020pile}
Leo Gao, Stella Biderman, Sid Black, Laurence Golding, Travis Hoppe, Charles Foster, Jason Phang, Horace He, Anish Thite, Noa Nabeshima, et~al.
\newblock The pile: An 800gb dataset of diverse text for language modeling.
\newblock \emph{arXiv preprint arXiv:2101.00027}, 2020.

\bibitem[Grangier et~al.(2023)Grangier, Ablin, and Hannun]{grangier2023adaptive}
David Grangier, Pierre Ablin, and Awni Hannun.
\newblock Adaptive training distributions with scalable online bilevel optimization.
\newblock \emph{arXiv preprint arXiv:2311.11973}, 2023.

\bibitem[Jhunjhunwala et~al.(2023)Jhunjhunwala, Wang, and Joshi]{jhunjhunwala2023fedexp}
Divyansh Jhunjhunwala, Shiqiang Wang, and Gauri Joshi.
\newblock Fedexp: Speeding up federated averaging via extrapolation.
\newblock In \emph{The Eleventh International Conference on Learning Representations}, 2023.

\bibitem[Jiang et~al.(2019)Jiang, Wong, Zhou, Andersen, Dean, Ganger, Joshi, Kaminksy, Kozuch, Lipton, et~al.]{jiang2019accelerating}
Angela~H Jiang, Daniel L-K Wong, Giulio Zhou, David~G Andersen, Jeffrey Dean, Gregory~R Ganger, Gauri Joshi, Michael Kaminksy, Michael Kozuch, Zachary~C Lipton, et~al.
\newblock Accelerating deep learning by focusing on the biggest losers.
\newblock \emph{arXiv preprint arXiv:1910.00762}, 2019.

\bibitem[Jiang et~al.(2024)Jiang, Chan, Xue, Liu, and Guo]{jiang2024importance}
Chunyang Jiang, Chi-min Chan, Wei Xue, Qifeng Liu, and Yike Guo.
\newblock Importance weighting can help large language models self-improve.
\newblock \emph{arXiv preprint arXiv:2408.09849}, 2024.

\bibitem[Jiang \& Zhai(2007)Jiang and Zhai]{jiang2007instance}
Jing Jiang and ChengXiang Zhai.
\newblock Instance weighting for domain adaptation in {NLP}.
\newblock In Annie Zaenen and Antal van~den Bosch (eds.), \emph{Proceedings of the 45th Annual Meeting of the Association of Computational Linguistics}, pp.\  264--271, June 2007.

\bibitem[Katharopoulos \& Fleuret(2018)Katharopoulos and Fleuret]{katharopoulos2018not}
Angelos Katharopoulos and Fran{\c{c}}ois Fleuret.
\newblock Not all samples are created equal: Deep learning with importance sampling.
\newblock In \emph{International conference on machine learning}, pp.\  2525--2534. PMLR, 2018.

\bibitem[Kumar et~al.(2023)Kumar, Majmundar, Nagaraj, and Suggala]{kumar2023stochastic}
Ramnath Kumar, Kushal Majmundar, Dheeraj Nagaraj, and Arun~Sai Suggala.
\newblock Stochastic re-weighted gradient descent via distributionally robust optimization.
\newblock \emph{arXiv preprint arXiv:2306.09222}, 2023.

\bibitem[Liu et~al.(2021)Liu, Han, Liu, Gong, Niu, Zhou, Sugiyama, et~al.]{liu2021probabilistic}
Feng Liu, Bo~Han, Tongliang Liu, Chen Gong, Gang Niu, Mingyuan Zhou, Masashi Sugiyama, et~al.
\newblock Probabilistic margins for instance reweighting in adversarial training.
\newblock \emph{Advances in Neural Information Processing Systems}, 34:\penalty0 23258--23269, 2021.

\bibitem[Liu et~al.(2023)Liu, Liu, Cui, Teng, Duan, Zhou, and Zhang]{liu2023logiqa}
Hanmeng Liu, Jian Liu, Leyang Cui, Zhiyang Teng, Nan Duan, Ming Zhou, and Yue Zhang.
\newblock Logiqa 2.0—an improved dataset for logical reasoning in natural language understanding.
\newblock \emph{IEEE/ACM Transactions on Audio, Speech, and Language Processing}, 2023.

\bibitem[Liu et~al.(2020)Liu, Cui, Liu, Huang, Wang, and Zhang]{liu2020logiqa}
Jian Liu, Leyang Cui, Hanmeng Liu, Dandan Huang, Yile Wang, and Yue Zhang.
\newblock Logiqa: A challenge dataset for machine reading comprehension with logical reasoning.
\newblock \emph{arXiv preprint arXiv:2007.08124}, 2020.

\bibitem[Longpre et~al.(2023)Longpre, Yauney, Reif, Lee, Roberts, Zoph, Zhou, Wei, Robinson, Mimno, et~al.]{longpre2023pretrainer}
Shayne Longpre, Gregory Yauney, Emily Reif, Katherine Lee, Adam Roberts, Barret Zoph, Denny Zhou, Jason Wei, Kevin Robinson, David Mimno, et~al.
\newblock A pretrainer's guide to training data: Measuring the effects of data age, domain coverage, quality, \& toxicity.
\newblock \emph{arXiv preprint arXiv:2305.13169}, 2023.

\bibitem[Loshchilov \& Hutter(2015)Loshchilov and Hutter]{loshchilov2015online}
Ilya Loshchilov and Frank Hutter.
\newblock Online batch selection for faster training of neural networks.
\newblock \emph{arXiv preprint arXiv:1511.06343}, 2015.

\bibitem[Penedo et~al.(2023)Penedo, Malartic, Hesslow, Cojocaru, Cappelli, Alobeidli, Pannier, Almazrouei, and Launay]{penedo2023refinedweb}
Guilherme Penedo, Quentin Malartic, Daniel Hesslow, Ruxandra Cojocaru, Alessandro Cappelli, Hamza Alobeidli, Baptiste Pannier, Ebtesam Almazrouei, and Julien Launay.
\newblock The refinedweb dataset for falcon llm: outperforming curated corpora with web data, and web data only.
\newblock \emph{arXiv preprint arXiv:2306.01116}, 2023.

\bibitem[Penedo et~al.(2024)Penedo, Kydlíček, allal, Lozhkov, Mitchell, Raffel, Werra, and Wolf]{penedo2024finewebdatasetsdecantingweb}
Guilherme Penedo, Hynek Kydlíček, Loubna~Ben allal, Anton Lozhkov, Margaret Mitchell, Colin Raffel, Leandro~Von Werra, and Thomas Wolf.
\newblock The fineweb datasets: Decanting the web for the finest text data at scale, 2024.
\newblock URL \url{https://arxiv.org/abs/2406.17557}.

\bibitem[Qi et~al.(2021)Qi, Guo, Xu, Jin, and Yang]{qi2021online}
Qi~Qi, Zhishuai Guo, Yi~Xu, Rong Jin, and Tianbao Yang.
\newblock An online method for a class of distributionally robust optimization with non-convex objectives.
\newblock \emph{Advances in Neural Information Processing Systems}, 34:\penalty0 10067--10080, 2021.

\bibitem[Qian et~al.(2019)Qian, Zhu, Tang, Jin, Sun, and Li]{qian2019robust}
Qi~Qian, Shenghuo Zhu, Jiasheng Tang, Rong Jin, Baigui Sun, and Hao Li.
\newblock Robust optimization over multiple domains.
\newblock In \emph{Proceedings of the AAAI Conference on Artificial Intelligence}, volume~33, pp.\  4739--4746, 2019.

\bibitem[Radford et~al.(2019)Radford, Wu, Child, Luan, Amodei, Sutskever, et~al.]{radford2019language}
Alec Radford, Jeffrey Wu, Rewon Child, David Luan, Dario Amodei, Ilya Sutskever, et~al.
\newblock Language models are unsupervised multitask learners.
\newblock \emph{OpenAI blog}, 1\penalty0 (8):\penalty0 9, 2019.

\bibitem[Radhakrishnan et~al.(2020)Radhakrishnan, Belkin, and Uhler]{radhakrishnan2020overparameterized}
Adityanarayanan Radhakrishnan, Mikhail Belkin, and Caroline Uhler.
\newblock Overparameterized neural networks implement associative memory.
\newblock \emph{Proceedings of the National Academy of Sciences}, 117\penalty0 (44):\penalty0 27162--27170, 2020.

\bibitem[Raffel et~al.(2020)Raffel, Shazeer, Roberts, Lee, Narang, Matena, Zhou, Li, and Liu]{raffel2020exploring}
Colin Raffel, Noam Shazeer, Adam Roberts, Katherine Lee, Sharan Narang, Michael Matena, Yanqi Zhou, Wei Li, and Peter~J Liu.
\newblock Exploring the limits of transfer learning with a unified text-to-text transformer.
\newblock \emph{The Journal of Machine Learning Research}, 21\penalty0 (1):\penalty0 5485--5551, 2020.

\bibitem[Ren et~al.(2018)Ren, Zeng, Yang, and Urtasun]{ren2018learning}
Mengye Ren, Wenyuan Zeng, Bin Yang, and Raquel Urtasun.
\newblock Learning to reweight examples for robust deep learning.
\newblock In \emph{International conference on machine learning}, pp.\  4334--4343. PMLR, 2018.

\bibitem[Sebbouh et~al.(2021)Sebbouh, Gower, and Defazio]{sebbouh2021almost}
Othmane Sebbouh, Robert~M Gower, and Aaron Defazio.
\newblock Almost sure convergence rates for stochastic gradient descent and stochastic heavy ball.
\newblock In \emph{Conference on Learning Theory}, pp.\  3935--3971. PMLR, 2021.

\bibitem[Soboleva et~al.(2023)Soboleva, Al-Khateeb, Myers, Steeves, Hestness, and Dey]{cerebras2023slimpajama}
Daria Soboleva, Faisal Al-Khateeb, Robert Myers, Jacob~R Steeves, Joel Hestness, and Nolan Dey.
\newblock {SlimPajama: A 627B token cleaned and deduplicated version of RedPajama}.
\newblock \url{https://cerebras.ai/blog/slimpajama-a-627b-token-cleaned-and-deduplicated-version-of-redpajama}, June 2023.
\newblock URL \url{https://huggingface.co/datasets/cerebras/SlimPajama-627B}.

\bibitem[Sow et~al.(2023)Sow, Lin, Wang, and Liang]{sow2023doubly}
Daouda Sow, Sen Lin, Zhangyang Wang, and Yingbin Liang.
\newblock Doubly robust instance-reweighted adversarial training.
\newblock \emph{arXiv preprint arXiv:2308.00311}, 2023.

\bibitem[Thakkar et~al.(2023)Thakkar, Bolukbasi, Ganapathy, Vashishth, Chandar, and Talukdar]{thakkar2023self}
Megh Thakkar, Tolga Bolukbasi, Sriram Ganapathy, Shikhar Vashishth, Sarath Chandar, and Partha Talukdar.
\newblock Self-influence guided data reweighting for language model pre-training.
\newblock \emph{arXiv preprint arXiv:2311.00913}, 2023.

\bibitem[Touvron et~al.(2023)Touvron, Martin, Stone, Albert, Almahairi, Babaei, Bashlykov, Batra, Bhargava, Bhosale, et~al.]{touvron2023llama}
Hugo Touvron, Louis Martin, Kevin Stone, Peter Albert, Amjad Almahairi, Yasmine Babaei, Nikolay Bashlykov, Soumya Batra, Prajjwal Bhargava, Shruti Bhosale, et~al.
\newblock Llama 2: Open foundation and fine-tuned chat models.
\newblock \emph{arXiv preprint arXiv:2307.09288}, 2023.

\bibitem[Vaswani(2017)]{vaswani2017attention}
A~Vaswani.
\newblock Attention is all you need.
\newblock \emph{Advances in Neural Information Processing Systems}, 2017.

\bibitem[Welbl et~al.(2017)Welbl, Liu, and Gardner]{welbl-etal-2017-crowdsourcing}
Johannes Welbl, Nelson~F. Liu, and Matt Gardner.
\newblock Crowdsourcing multiple choice science questions.
\newblock In \emph{Proceedings of the 3rd Workshop on Noisy User-generated Text}, pp.\  94--106, Copenhagen, Denmark, September 2017. Association for Computational Linguistics.

\bibitem[Wettig et~al.(2024)Wettig, Gupta, Malik, and Chen]{pmlr-v235-wettig24a}
Alexander Wettig, Aatmik Gupta, Saumya Malik, and Danqi Chen.
\newblock {Q}u{R}ating: Selecting high-quality data for training language models.
\newblock In Ruslan Salakhutdinov, Zico Kolter, Katherine Heller, Adrian Weller, Nuria Oliver, Jonathan Scarlett, and Felix Berkenkamp (eds.), \emph{Proceedings of the 41st International Conference on Machine Learning}, volume 235 of \emph{Proceedings of Machine Learning Research}, pp.\  52915--52971. PMLR, 21--27 Jul 2024.
\newblock URL \url{https://proceedings.mlr.press/v235/wettig24a.html}.

\bibitem[Xie et~al.(2023)Xie, Pham, Dong, Du, Liu, Lu, Liang, Le, Ma, and Yu]{xie2023doremi}
Sang~Michael Xie, Hieu Pham, Xuanyi Dong, Nan Du, Hanxiao Liu, Yifeng Lu, Percy Liang, Quoc~V Le, Tengyu Ma, and Adams~Wei Yu.
\newblock Doremi: Optimizing data mixtures speeds up language model pretraining.
\newblock \emph{arXiv preprint arXiv:2305.10429}, 2023.

\bibitem[Yi et~al.(2021)Yi, Hou, Shang, Jiang, Liu, and Ma]{yi2021reweighting}
Mingyang Yi, Lu~Hou, Lifeng Shang, Xin Jiang, Qun Liu, and Zhi-Ming Ma.
\newblock Reweighting augmented samples by minimizing the maximal expected loss.
\newblock \emph{arXiv preprint arXiv:2103.08933}, 2021.

\bibitem[Zeng et~al.(2021)Zeng, Zhu, Goldstein, and Huang]{zeng2021adversarial}
Huimin Zeng, Chen Zhu, Tom Goldstein, and Furong Huang.
\newblock Are adversarial examples created equal? a learnable weighted minimax risk for robustness under non-uniform attacks.
\newblock In \emph{Proceedings of the AAAI Conference on Artificial Intelligence}, volume~35, pp.\  10815--10823, 2021.

\bibitem[Zhang et~al.(2020)Zhang, Zhu, Niu, Han, Sugiyama, and Kankanhalli]{zhang2020geometry}
Jingfeng Zhang, Jianing Zhu, Gang Niu, Bo~Han, Masashi Sugiyama, and Mohan Kankanhalli.
\newblock Geometry-aware instance-reweighted adversarial training.
\newblock \emph{arXiv preprint arXiv:2010.01736}, 2020.

\end{thebibliography}

\newpage
\appendix
\begin{center}
    {\bf\Large Appendix}
\end{center}

\startcontents[sections]
\printcontents[sections]{l}{1}{\setcounter{tocdepth}{2}}
\clearpage

\section{Experimental Details} \label{app:more_exp}

For full reproducibility of our work we provide insights on extending existing training routines with our minimal-invasive loss reweighting scheme.

\subsection{Reference Implementation \& Reproducibility}
To integrate our loss reweighting method with a PyTorch training routine, we have to add two plug-and-play components.
First, we need to define four \textbf{utility functions} to ensure the proper functionality of our approach with reference implementations we provide in \Cref{list:utility_func}: 

\begin{enumerate}
    \item \texttt{apply\_strategy(...)}: Implements different reweighting strategies such as focusing on medium losses (quadratic), capping high losses (linupper), or emphasizing extreme values (extremes). 

    \item \texttt{scale\_losses(...)}: Applies exponential scaling to adjust the importance of losses based on a temperature parameter $r$.

    \item \texttt{normalize\_losses(...)}: Normalizes losses to a bounded interval $[-\delta, \delta]$ for consistency in subsequent reweighting.

    \item \texttt{get\_batch\_loss\_from\_logits(...)}: Computes per-sample losses using cross-entropy, ensuring padding tokens do not affect the results. 
    
\end{enumerate}

These four functions are then employed inside the training step. We need to capture the unprocessed raw losses and apply our reweighting scheme. We provide a reference implementation for PyTorch FSDP training in \Cref{list:fsdp_training}
to enable multi-GPU training.

For a full reference, we provide a ready-to-use implementation on GitHub for the GPT2 experiments: \url{\codebase}.

\begin{lstlisting}[language=Python, caption=Utility functions, label=list:utility_func]
import torch

# Exponential scaling function
def scale_losses(losses, r):
    return torch.exp(losses / r)

# Normalization function for losses
def normalize_losses(losses, delta=1., l_min=0., l_max=1.):
    return 2. * delta * losses / max(l_max - l_min, 1e-6) - delta * (l_max + l_min) / max(l_max - l_min, 1e-6)

# Reweighting strategies
def apply_strategy(losses, delta=1.0, strategy="linupper"):
    if strategy == "linupper": 
        return torch.minimum(losses + delta, delta * torch.ones_like(losses))
    elif strategy == "uniform": 
        # We do not reweight here. This is our baseline.
        return losses
    elif strategy == "quadratic": 
        return 1 - losses**2 / delta**2
    elif strategy == "extremes": 
        return torch.abs(losses)
    else: 
        raise NotImplementedError

# Compute batch losses from logits
def get_batch_loss_from_logits(logits, labels):
    ignore_index = -100
    shift_logits = logits[..., :-1, :].contiguous()
    shift_labels = labels[..., 1:].contiguous()
    num_active = (shift_labels != ignore_index).sum(dim=1)
    loss_fct = torch.nn.CrossEntropyLoss(reduction='none')
    loss = loss_fct(shift_logits.view(-1, logits.size(-1)), shift_labels.view(-1).long())
    return loss.view(logits.size(0), -1).sum(dim=1) / num_active, num_active
\end{lstlisting}

\begin{lstlisting}[language=Python, caption=Integration of reweighting strategies into a multi-GPU training loop., label=list:fsdp_training]
import torch
import torch.distributed as dist

# Define reweighting function mapping
STRATEGY = "linupper"

# This has to be placed inside a training step.
...

# Assume model outputs (logits) and labels are already computed
device_losses, len_norms = get_batch_loss_from_logits(outputs, labels)

# Initialize placeholder to store the losses from all GPUs
gathered_losses = torch.zeros(
    dist.get_world_size(),
    len(device_losses),
    device=device_losses.device,
    dtype=device_losses.dtype
)

# Gather losses across all GPUs into tenor
dist.all_gather_into_tensor(gathered_losses, device_losses.detach())

r = r_scheduler(step, cfg.initial_r)
# Compute sample weights
with torch.no_grad():
    min_loss = gathered_losses.min().item()
    max_loss = gathered_losses.max().item()
    normalized_losses = normalize_losses(gathered_losses.view(-1), delta=1., l_min=min_loss, l_max=max_loss)
    reweighted_losses = apply_strategy(normalized_losses, delta=1., strategy=STRATEGY)
    scaled_losses = scale_losses(reweighted_losses - reweighted_losses.max().item(), l=r)
    weights = scaled_losses / scaled_losses.sum()

    # for instance local_rank can be obtained with dist.get_rank()
    device_weights = weights.view(dist.get_world_size(), -1)[local_rank, :] 

# Reweight losses and scale appropriately
loss = torch.sum(device_weights * device_losses) * dist.get_world_size()
loss.backward()
\end{lstlisting}

\subsection{Training \& Evaluation Details}
\label{app:arch}
In the following, we outline the hyper-parameter configuration and hardware requirements to fully reproduce our experiments.

\textbf{Models}. 
We employ three different \textit{GPT2 models} to study the effectiveness of our reweighting approach on various model sizes. 
We report the architectural details in \Cref{tab:model_architecture}. 
Furthermore, to demonstrate the performance of our reweighting approach on state-of-the-art models, we employ a 1.4B parameter model with \textit{Llama} architecture.

\begin{table}[H]
    \centering
    \caption{Model Architecture}
    \label{tab:model_architecture}
    \resizebox{0.9\textwidth}{!}{
        \begin{tabular}{@{}llrrrrrr}
            \toprule
            Architecture & Name & \multicolumn{1}{l}{Parameters} & \multicolumn{1}{l}{Layers} & \multicolumn{1}{l}{Attention Heads} & \multicolumn{1}{l}{Embedding Dimensions} & \multicolumn{1}{l}{Hidden Dimensions} & Max. Sequence Length \\ \midrule
            \multirow{3}{*}{GPT-2}      & Mini      & 124M  & 12 & 12  & 768  & 3072  & 512  \\
                                        & Small     & 210M  & 24 & 16  & 786  & 3072  & 512  \\
                                        & Medium    & 300M  & 36 & 24  & 786  & 3072  & 512  \\ \midrule
            \multirow{2}{*}{LLAMA}      & 1.4B      & 1400M & 24 & 16  & 2048 & 2048  & 8192 \\ 
                                        & 7B        & 7000M & 32 & 32  & 4096 & 4096  & 8192 \\ \bottomrule
        \end{tabular}
    }
\end{table}

\textbf{Datasets}. As a dataset for training our GPT2 models, we employ the SlimPajama-6B dataset, consisting of 7 domain partitions (ArXiv, Book, CC, C4, GitHub, StackExchange, and Wikipedia). The exact number of documents seen is reported in \Cref{tab:model_hparams}. 
We use the FineWeb 15T dataset to train the larger Llama-1.4B and Llama-7B models. 
For the 1.4B and 7B models, we use approximately 100B and 175B randomly sampled tokens from the FineWeb 15T dataset, respectively. 

\textbf{Pretraining procedure}. 
We pretrain all models with masked language modeling and do not apply any instruction tuning routine for subsequent evaluations. 
We set $r$ to 100 for a specific number of warmup steps and then reduce it to 1 for the remaining steps. 
The GPT2 models are trained for 20,000 steps in total, and the Llama models are trained for 100,000 and 170,000 steps, respectively, for the 1.4B and 7B models. More hyperparameters are provided in \Cref{tab:model_hparams}. For the GPT2 models, we use the Huggingface \texttt{trainer} interface and the \texttt{accelerate} library for distributed training. For the Llama models, we use fully sharded data parallel (FSDP). 

\begin{table}[H]
    \centering
    \caption{Training Hyperparameters for our benchmark evaluations}
    \label{tab:model_hparams}
    \resizebox{1.0\textwidth}{!}{
        \begin{tabular}{@{}llrrrlrrrrrr@{}}
            \toprule
            Architecture & Name & \multicolumn{1}{c}{Minibatch} & \multicolumn{1}{c}{Learning} & \multicolumn{1}{c}{Weight} & \multicolumn{1}{c}{LR} & \multicolumn{1}{c}{LR} & \multicolumn{1}{c}{Warmup} & \multicolumn{1}{c}{Max. Grad} & \multicolumn{1}{c}{r} & \multicolumn{1}{c}{Training} & \multicolumn{1}{c}{Total} \\
             & &\multicolumn{1}{c}{Size} & \multicolumn{1}{c}{Rate (LR)} & \multicolumn{1}{c}{Decay} & \multicolumn{1}{c}{Schedule} & \multicolumn{1}{c}{End} & \multicolumn{1}{c}{Steps} & \multicolumn{1}{c}{Norm} & & \multicolumn{1}{c}{Steps} & \multicolumn{1}{c}{Documents Seen} \\ \midrule
            \multirow{3}{*}{GPT-2} & Mini & 32 & 0.0005 & 0.01 & Linear Warmup Cosine & 0.0001 & 500  & 1.0  & 0.4 & 20,000 & 640,000  \\
             & Small  & 48 & 0.0005 & 0.01 & Linear Warmup Cosine & 0.0001 & 500  & 1.0  & 0.4 & 20,000 & 960,000  \\
             & Medium & 48 & 0.0005 & 0.01 & Linear Warmup Cosine & 0.0001 & 500  & 1.0  & 0.4 & 20,000 & 960,000  \\ \midrule
            \multirow{2}{*}{LLAMA}  & 1.4B & 128  & 0.0003 & 0.01 & Linear Warmup Cosine & 0.00003  & 2,000  & 1.0  & 1.0 & 100,000 & 12,800,000  \\ 
                                    & 7B   & 128  & 0.0003 & 0.01 & Linear Warmup Cosine & 0.00003  & 2,000  & 1.0  & 2.0 & 170,000 & 21,760,000 \\ \bottomrule
                                    
        \end{tabular}
    }
\end{table}

\textbf{Benchmarks}. We use the LM Eval Harness to generate our benchmark results. We evaluate the GPT2 models every 2000 steps and the Llama model every 5000 steps. 
For the GPT2 models, we find that only question-answering (QA) tasks work well at this scale, which is why we report 5-shot results on LogiQA, LogiQA-2, SciQ, and PiQA. 
All other benchmarks yielded random results at this scale, i.e., they were not useful for model evaluation. 
The reason is the limited capacity of the GPT2 models compared to larger models like the Llama-1.4B or 7B models.
For the Llama architecture models, we employ 13 language understanding and reasoning (LUR) and 6 QA tasks to provide a holistic picture.

\subsection{Additional Experimental Results}
\label{sec:apdix_experimental_results}

\textbf{Benchmark results}. 
Using our \texttt{LinUpper} method, the experiments with the 1.4B parameter Llama model show an average performance gain of $0.66\%$, improving on 6 out of 13 benchmarks and tied on one benchmark (\Cref{tab:llama1.4b_benchmark_results_lur}). For QA benchmarks, we improved on 5/6 tasks with an average of $1.72\%$ (\Cref{tab:llama1.4b_benchmark_results_qa}).

We observe that our \textbf{LinUpper} strategy also improves the performance of the 7B model. 
Our strategy outperforms the baseline in 9/13 LUR (\Cref{tab:llama7b_benchmark_results_lur}) and 5/6 QA tasks (\Cref{tab:llama7b_benchmark_results_qa}), respectively. 
On average, across all tasks, our \textbf{LinUpper} strategy improves performance by $1.44\%$ for LUR tasks and $1.16\%$ for QA tasks.

\begin{figure}[H]
    \centering
    \begin{subfigure}{0.24\textwidth}
        \includegraphics[width=\linewidth]{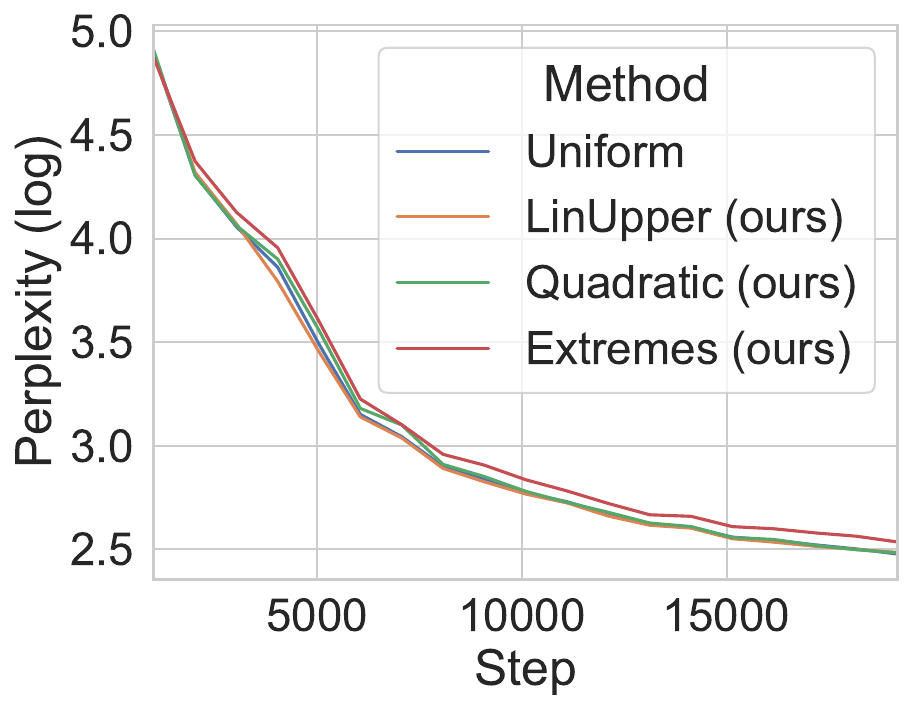}
        \caption{Arxiv Dataset}
        \label{fig:gpt2mini_arxiv_dataset_apdx}
    \end{subfigure}
    \begin{subfigure}{0.24\textwidth}
        \includegraphics[width=\linewidth]{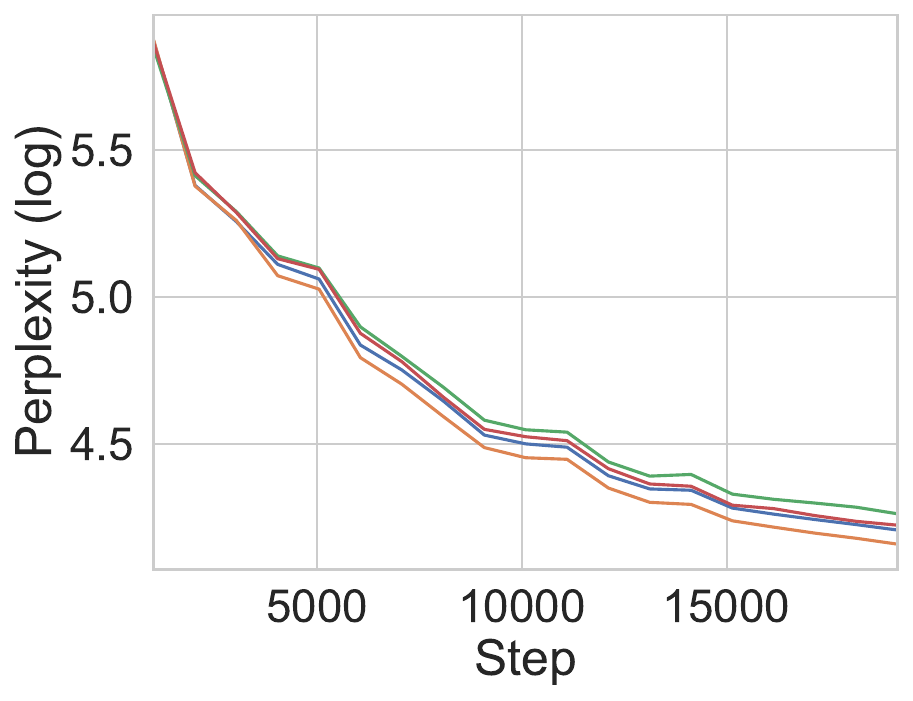}
        \caption{Book Dataset}
        \label{fig:gpt2mini_book_dataset_apdx}
    \end{subfigure}
    \begin{subfigure}{0.24\textwidth}
        \includegraphics[width=\linewidth]{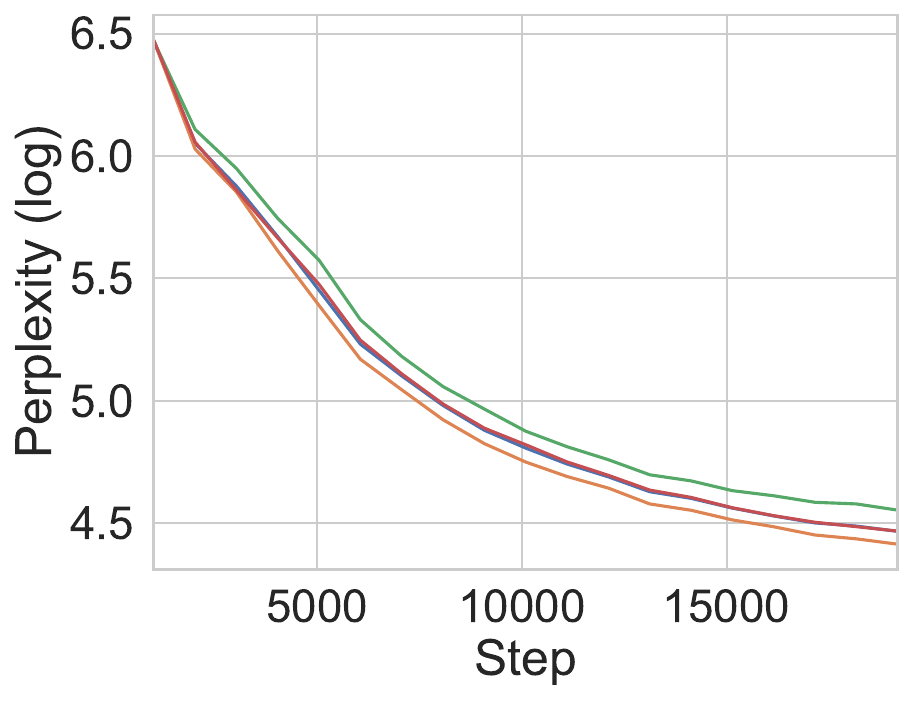}
        \caption{C4 Dataset}
        \label{fig:gpt2mini_c4_dataset_apdx}
    \end{subfigure}
    \begin{subfigure}{0.24\textwidth}
        \includegraphics[width=\linewidth]{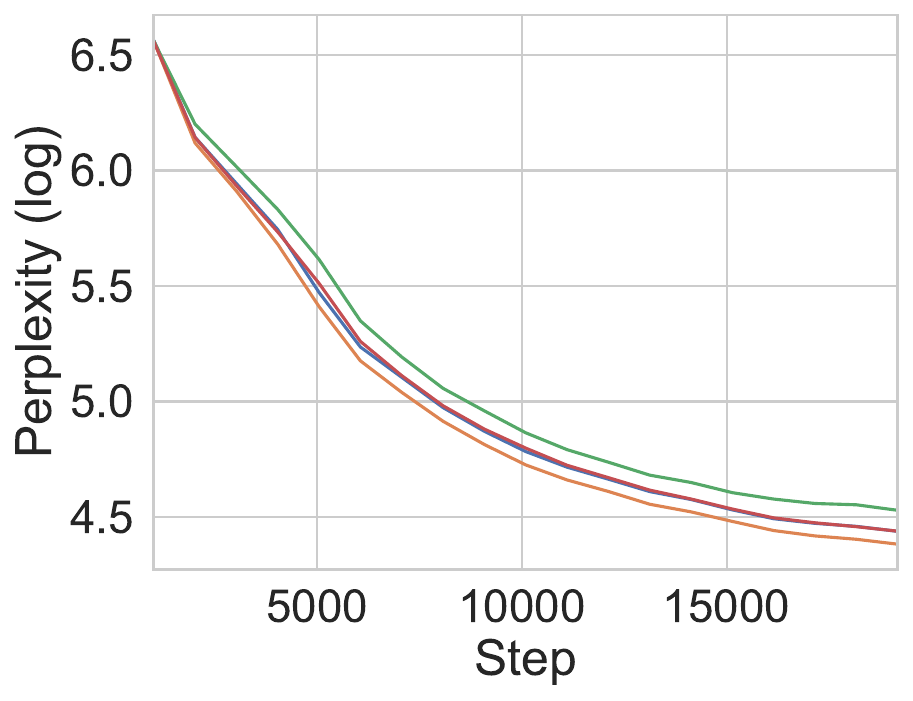}
        \caption{CC Dataset}
        \label{fig:gpt2mini_cc_dataset_apdx}
    \end{subfigure}
    \begin{subfigure}{0.24\textwidth}
        \includegraphics[width=\linewidth]{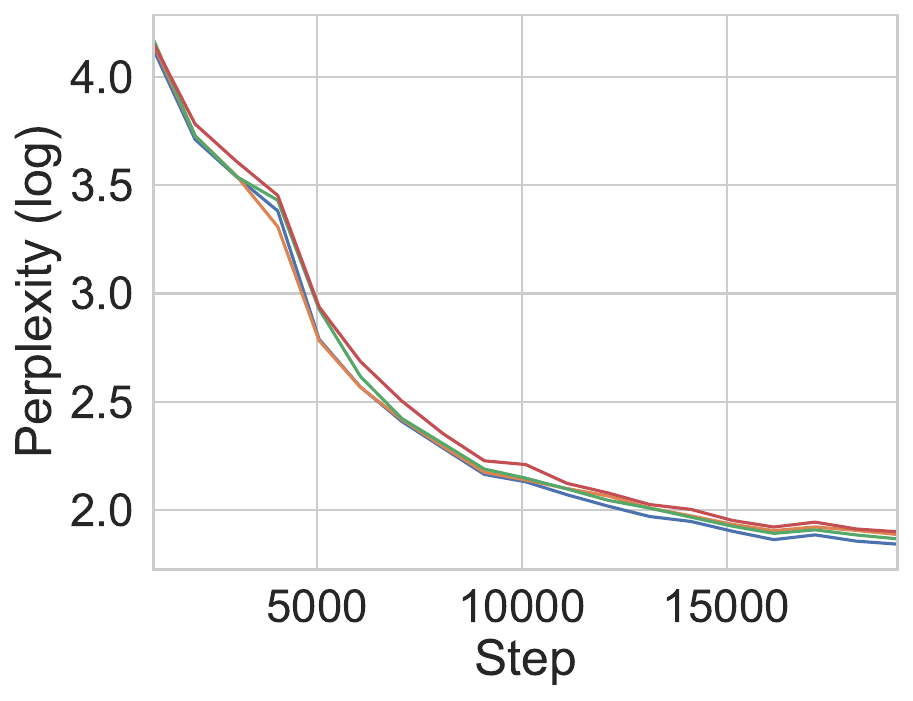}
        \caption{GitHub Dataset}
        \label{fig:gpt2mini_github_dataset_apdx}
    \end{subfigure}
    \begin{subfigure}{0.24\textwidth}
        \includegraphics[width=\linewidth]{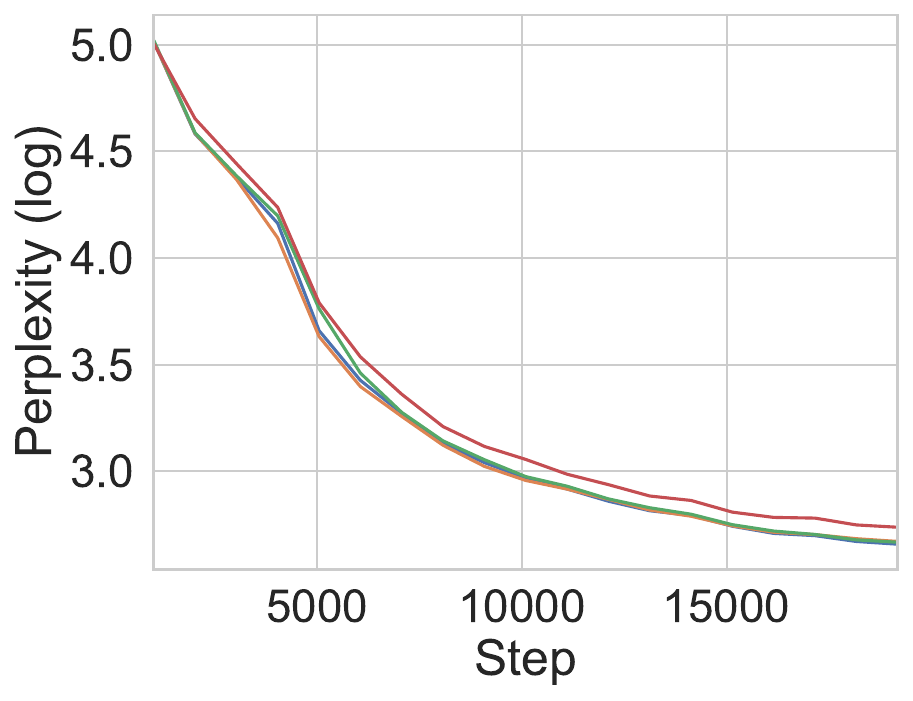}
        \caption{StackExchange Dataset}
        \label{fig:gpt2mini_stackexchange_dataset_apdx}
    \end{subfigure}
    \begin{subfigure}{0.24\textwidth}
        \includegraphics[width=\linewidth]{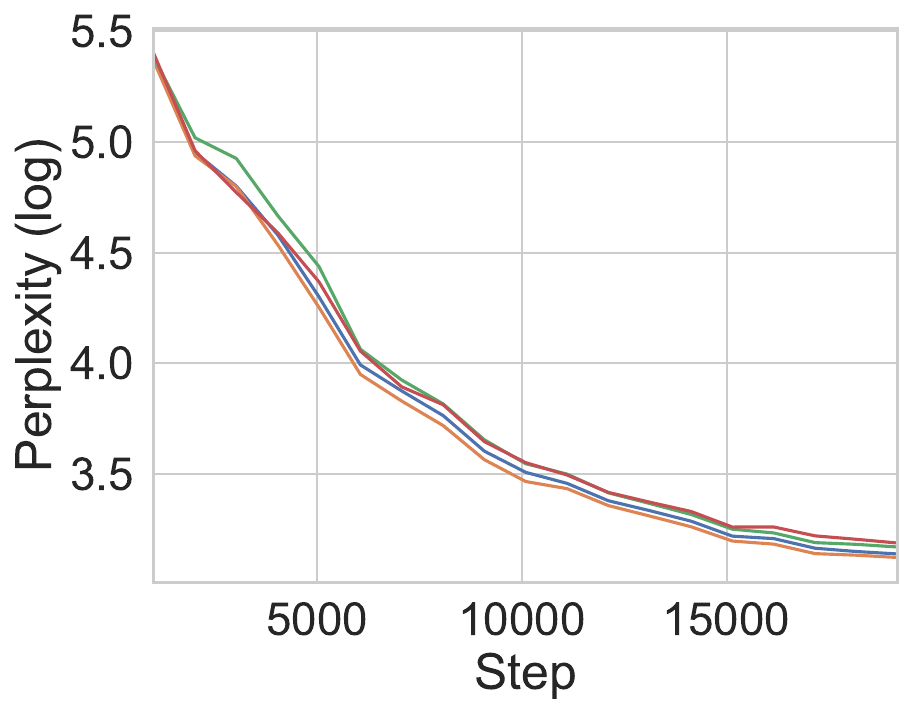}
        \caption{Wikipedia Dataset}
        \label{fig:gpt2mini_wikipedia_dataset_apdx}
    \end{subfigure}

\caption{Per-domain perplexities on hold-out validation sets under the uniform domain sampling setting for the GPT2-mini model. Our reweighting strategy \texttt{LinUpper} strategy achieves better or at least comparable perplexity on 6 out of 7 domains. 
}
\label{fig:perplexity_results_mini_apdx}
\end{figure}

\begin{figure}[H]
    \centering
    \begin{subfigure}{0.24\textwidth}
        \includegraphics[width=\linewidth]{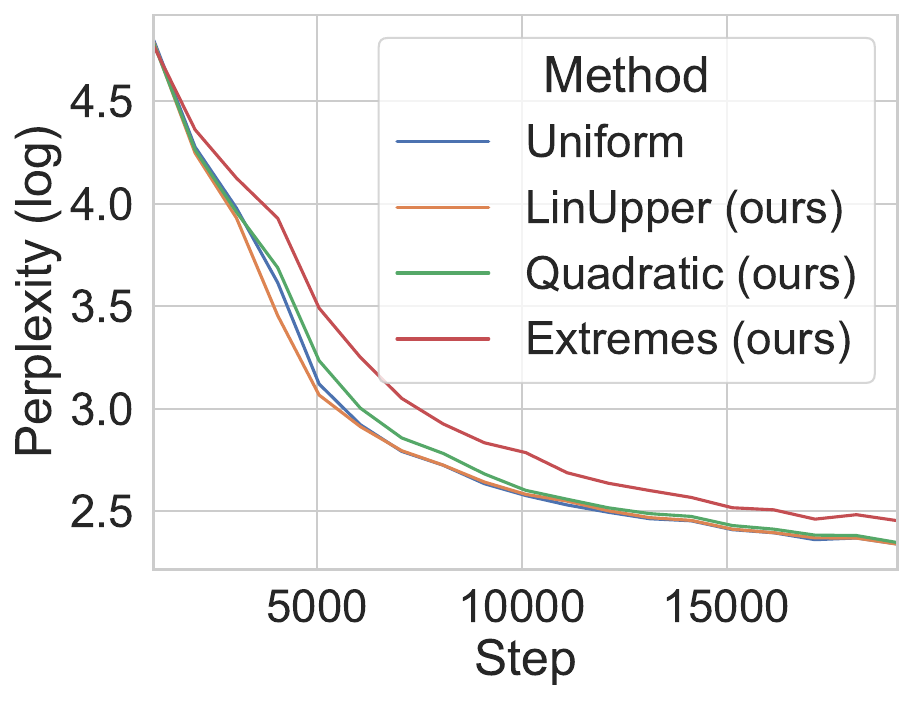}
        \caption{Arxiv Dataset}
        \label{fig:gpt2small_arxiv_dataset_apdx}
    \end{subfigure}
    \begin{subfigure}{0.24\textwidth}
        \includegraphics[width=\linewidth]{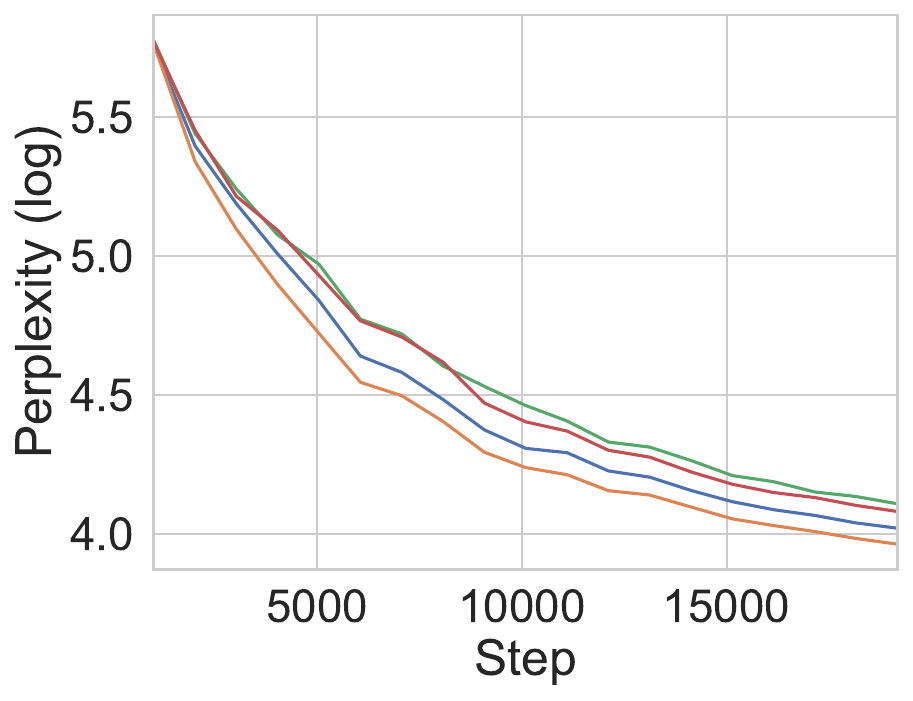}
        \caption{Book Dataset}
        \label{fig:gpt2small_book_dataset_apdx}
    \end{subfigure}
    \begin{subfigure}{0.24\textwidth}
        \includegraphics[width=\linewidth]{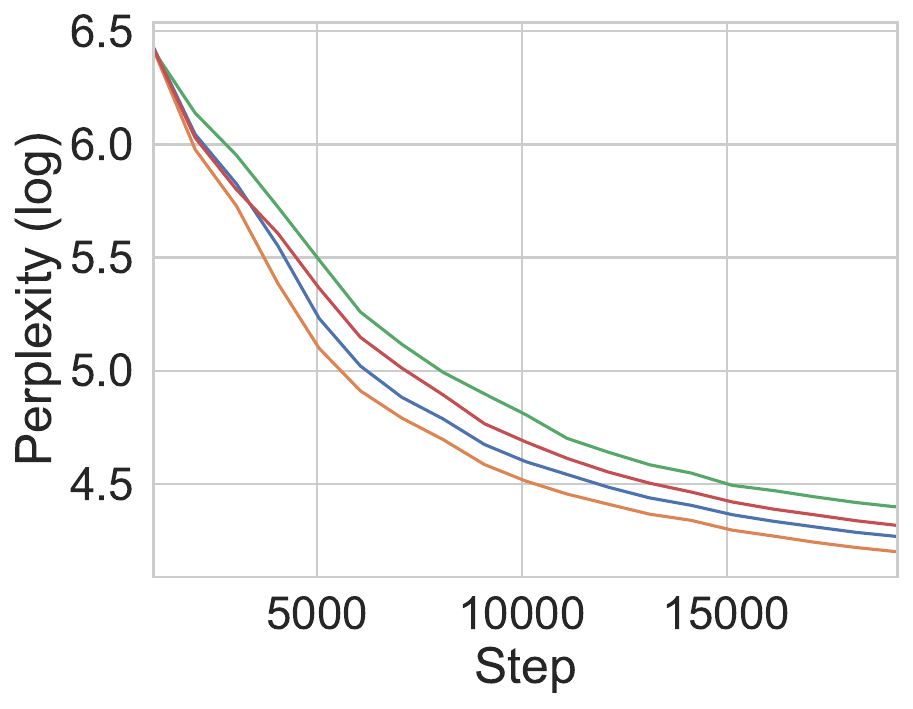}
        \caption{C4 Dataset}
        \label{fig:gpt2small_c4_dataset_apdx}
    \end{subfigure}
    \begin{subfigure}{0.24\textwidth}
        \includegraphics[width=\linewidth]{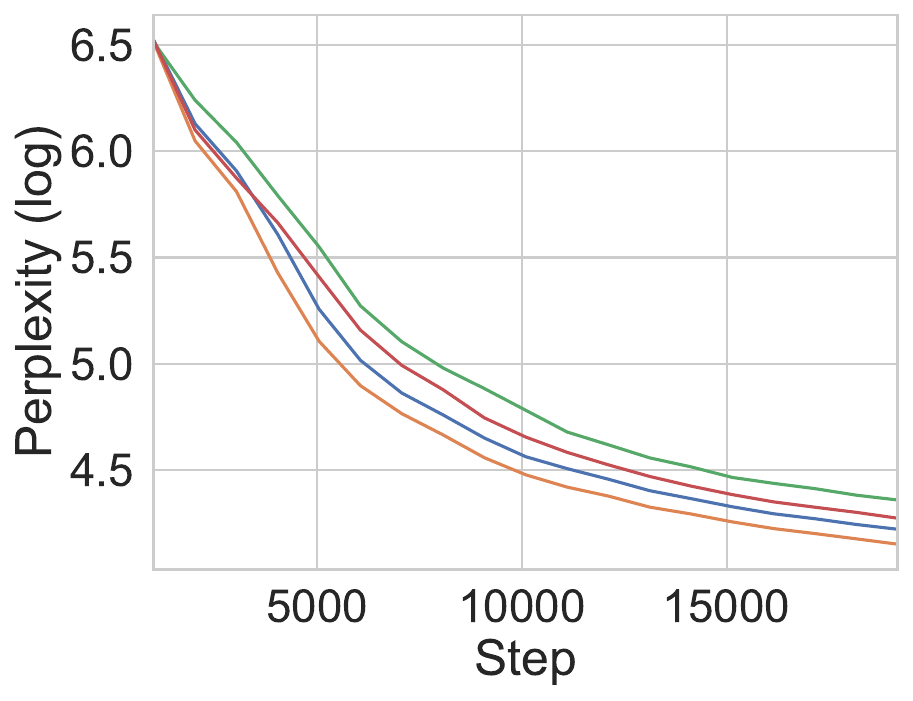}
        \caption{CC Dataset}
        \label{fig:gpt2small_cc_dataset}
    \end{subfigure}
    \begin{subfigure}{0.24\textwidth}
        \includegraphics[width=\linewidth]{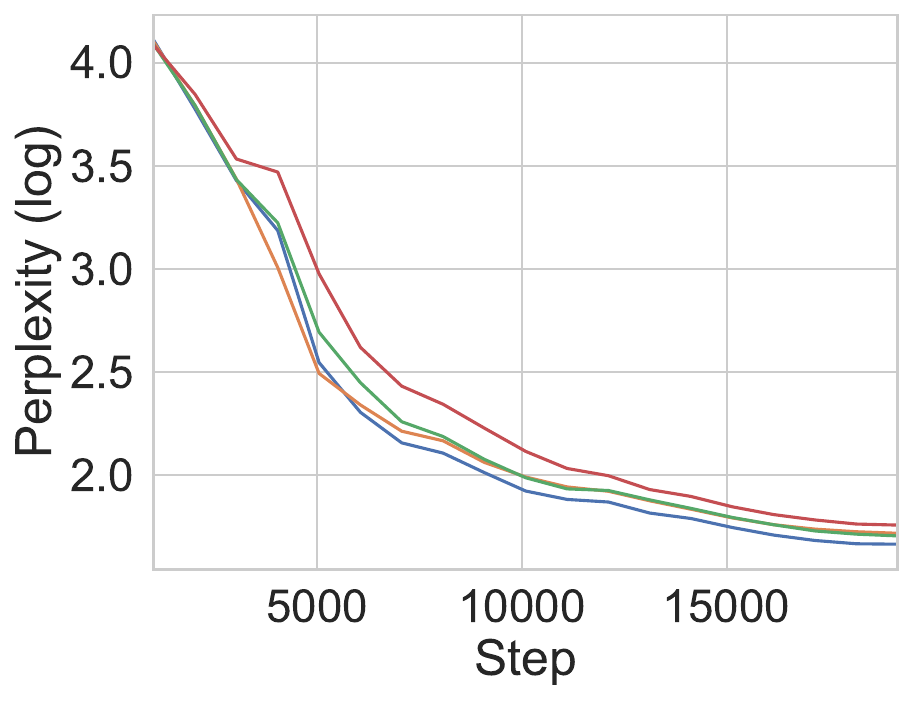}
        \caption{GitHub Dataset}
        \label{fig:gpt2small_github_dataset_apdx}
    \end{subfigure}
    \begin{subfigure}{0.24\textwidth}
        \includegraphics[width=\linewidth]{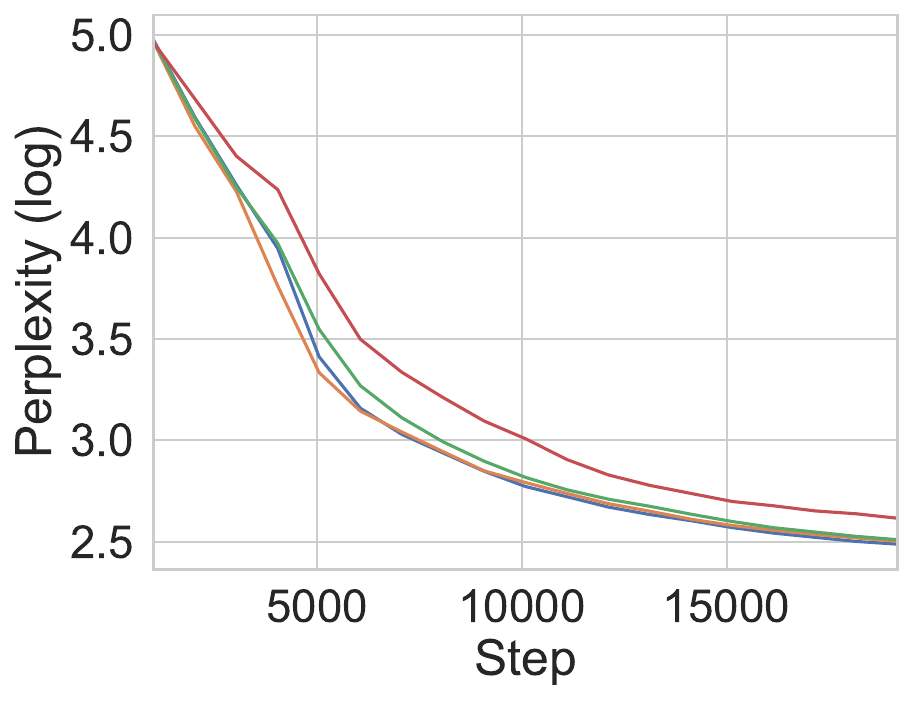}
        \caption{StackExchange Dataset}
        \label{fig:gpt2small_stackexchange_dataset_apdx}
    \end{subfigure}
    \begin{subfigure}{0.24\textwidth}
        \includegraphics[width=\linewidth]{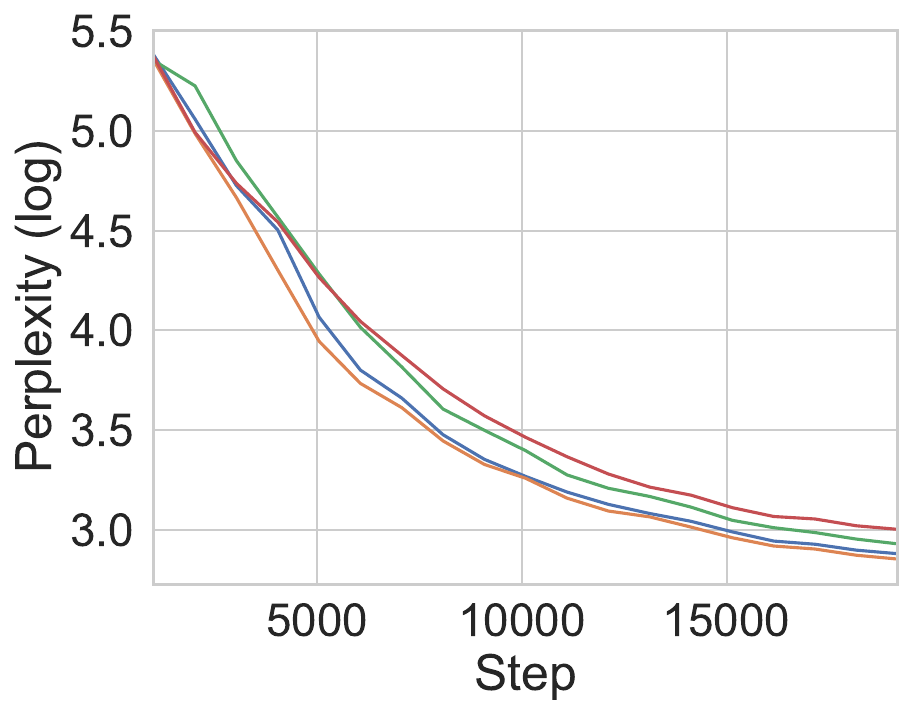}
        \caption{Wikipedia Dataset}
        \label{fig:gpt2small_wikipedia_dataset_apdx}
    \end{subfigure}

    \caption{Per-domain perplexities on hold-out validation sets under the uniform domain sampling setting for the GPT2-small model. Our reweighting strategy \texttt{LinUpper} strategy achieves better or at least comparable perplexity on 6 out of 7 domains. 
    }
    \label{fig:perplexity_results_small_apdx}
\end{figure}

\textbf{Perplexity}. We present additional perplexity visualizations of our GPT2 pretraining runs. \Cref{fig:perplexity_results_mini_apdx} shows the results for GPT2-mini and \Cref{fig:perplexity_results_small_apdx} shows the results for GPT2-small.

\begin{table}[H]
    \parbox{.48\linewidth}{
    \centering
    \caption{\textcolor{black}{Results of 1.4B parameter model on language understanding and reasoning benchmarks. Overall, our \texttt{LinUpper} method improves the results for 6 out of 13 tasks, and there is a tie for one task, with an average performance gain of $0.66\%$.}}
    \label{tab:llama1.4b_benchmark_results_lur}
    \resizebox{0.45\textwidth}{!}{
    \color{black}
        \begin{tabular}{lrrr}
            \toprule
            Benchmark Name & Uniform & LinUpper (ours) & Difference \\
            \midrule
            ARC Challenge & 33.19 & 33.02 & -0.17 \\
            ARC Easy & 63.80 & 63.38 & -0.42 \\
            COPA & 77.0 & 77.0 & 0.00 \\
            Lambada (OpenAI) & 49.87 & 49.60 & -0.27 \\
            Lambada (Standard) & 45.20 & 43.30 & -1.90 \\
            MMLU & 26.04 & 25.49 & -0.54 \\
            MNLI & 31.91 & 36.92 & 5.01 \\
            MNLI (mismatch) & 32.07 & 36.56 & 4.50 \\
            RTE & 51.26 & 52.35 & 1.08 \\
            SST-2 & 66.86 & 59.40 & -7.45 \\
            TinyARC & 40.46 & 40.85 & 0.39 \\
            TinyWinoGrande & 56.17 & 62.11 & 5.93 \\
            WinoGrande & 56.75 & 59.12 & 2.37 \\ \midrule
            \textit{Mean} & \textit{48.51} & \textit{49.16} & \textit{0.66} \\
            \bottomrule
        \end{tabular}

    }
    \vfill
    }
    \hfill
    \parbox{.48\linewidth}{
    \centering
    \caption{\textcolor{black}{Results of 1.4B parameter model on language question answering benchmarks. Overall, our \texttt{LinUpper} method improves results for 5 out of 6 tasks with an average performance gain of $1.72\%$.}}
    \label{tab:llama1.4b_benchmark_results_qa}
    \vspace{\baselineskip}
    \resizebox{0.48\textwidth}{!}{
    \color{black}
        \begin{tabular}{lrrr}
            \toprule
            Benchmark Name & Uniform & LinUpper (ours) & Difference \\
            \midrule
            BoolQ & 61.59 & 65.47 & 3.88 \\
            LogiQA & 20.28 & 23.66 & 3.38 \\
            LogiQA2 & 26.02 & 27.48 & 1.46 \\
            SciQ & 91.30 & 91.90 & 0.60 \\
            SocialIQA & 43.86 & 45.19 & 1.33 \\
            TriviaQA (Exact Match) & 11.05 & 10.73 & -0.32 \\ \midrule
            \textit{Mean} & \textit{42.35} & \textit{44.07} & \textit{1.72} \\
            \bottomrule
        \end{tabular}
    }
    \vspace{4.5\baselineskip}
    }
\end{table}

\begin{table}[H]
    \parbox{.48\linewidth}{
    \centering
    \caption{\textcolor{black}{Results of the 7B parameter model on language understanding and reasoning benchmark tasks. Overall, our \texttt{LinUpper} method improves results for 9 out of 13 tasks with an average performance gain of $1.44\%$.}}
    \label{tab:llama7b_benchmark_results_lur}
    \resizebox{0.45\textwidth}{!}{
        \color{black}
        \begin{tabular}{lrrr}
            \toprule
             Benchmark Name& Uniform & LinUpper (ours) & Difference \\
            \midrule
            ARC Challenge & 41.81 & 39.42 & -2.39 \\
            ARC Easy & 72.77 & 74.24 & 1.47 \\
            COPA & 83.00 & 84.00 & 1.00 \\
            Lambada (OpenAI) & 61.28 & 62.43 & 1.14 \\
            Lambada (Standard) & 58.08 & 58.80 & 0.72 \\
            MMLU & 24.33 & 25.81 & 1.47 \\
            MNLI & 32.28 & 34.24 & 1.97 \\
            MNLI (mismatch) & 32.35 & 34.01 & 1.66 \\
            RTE & 50.54 & 49.10 & -1.44 \\
            SST-2 & 49.89 & 69.38 & 19.50 \\
            TinyARC & 36.17 & 32.80 & -3.37 \\
            TinyWinoGrande & 49.09 & 44.20 & -4.88 \\
            WinoGrande & 65.43 & 67.32 & 1.89 \\ \midrule
            \textit{Mean} & \textit{50.54} & \textit{51.98} & \textit{1.44} \\
            \bottomrule
        \end{tabular}
    }
    \vfill
    }
    \hfill
    \parbox{.48\linewidth}{
    \centering
    \vspace{0\baselineskip}
    \caption{\textcolor{black}{Results of the 7B parameter model on question answering benchmark tasks. Overall, our \texttt{LinUpper} method improves results for 5 out of 6 tasks with an average performance gain of $1.16\%$.}}
    \label{tab:llama7b_benchmark_results_qa}
    \vspace{0\baselineskip}
    \resizebox{0.48\textwidth}{!}{
        \color{black}
        \begin{tabular}{lrrr}
            \toprule
            Benchmark Name & Uniform & LinUpper (ours) & Difference \\
            \midrule
            BoolQ & 73.52 & 74.40 & 0.89 \\
            LogiQA & 23.35 & 27.65 & 4.30 \\
            LogiQA2 & 26.72 & 27.86 & 1.15 \\
            SciQ & 95.20 & 95.00 & -0.20 \\
            SocialIQA & 47.80 & 48.11 & 0.31 \\
            TriviaQA (Exact Match) & 22.30 & 22.79 & 0.49 \\ \midrule
            \textit{Mean} & \textit{48.15} & \textit{49.31} & \textit{1.16}\\
            \bottomrule
        \end{tabular}
    }
    
    \vspace{4.5\baselineskip}
    }
\end{table}

\subsection{Sensitivity to the value of $r$}
\begin{figure}[H]
    \centering
    \begin{subfigure}{0.24\textwidth}
        \includegraphics[width=\linewidth]{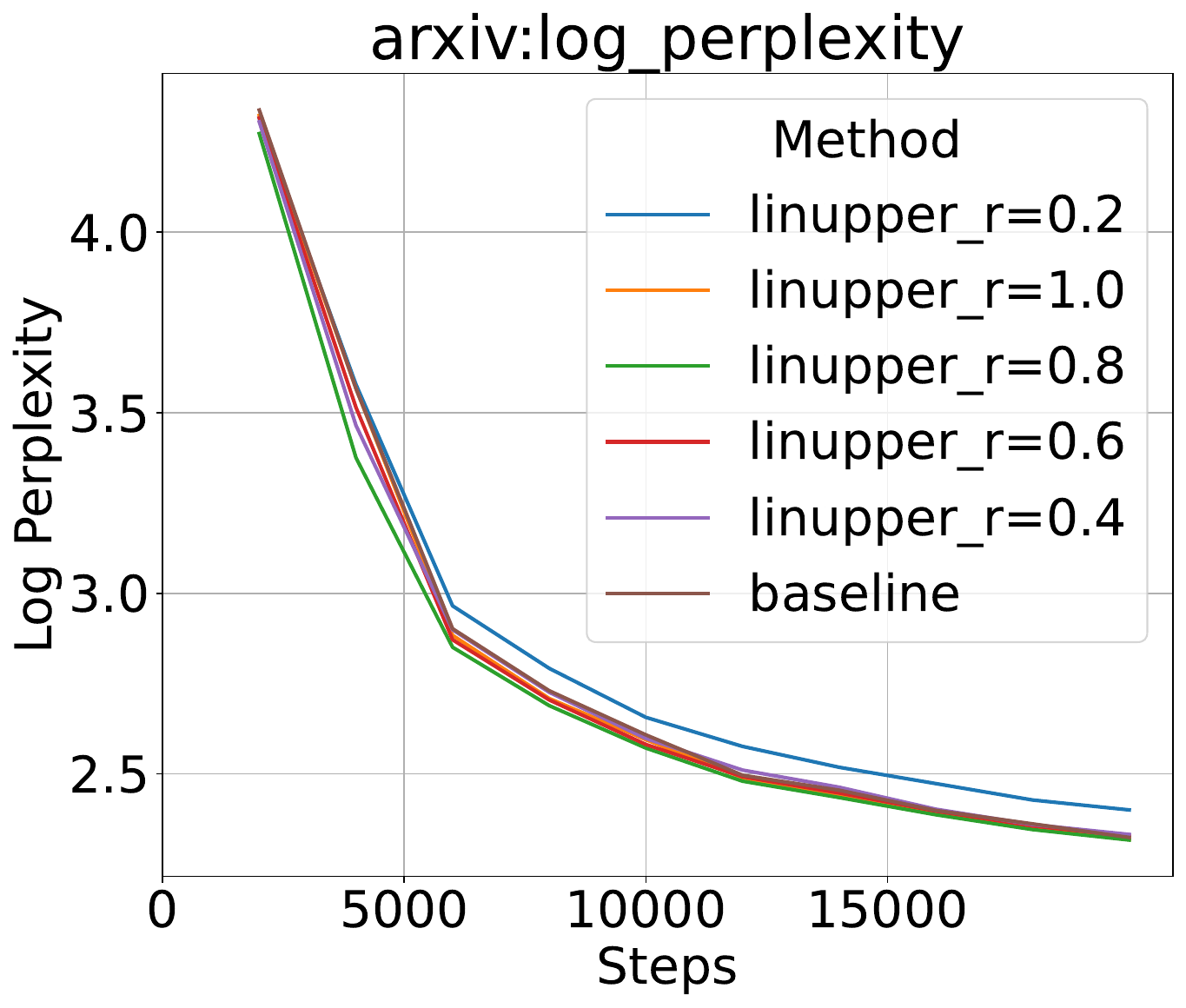}
    \end{subfigure}
    \begin{subfigure}{0.24\textwidth}
        \includegraphics[width=\linewidth]{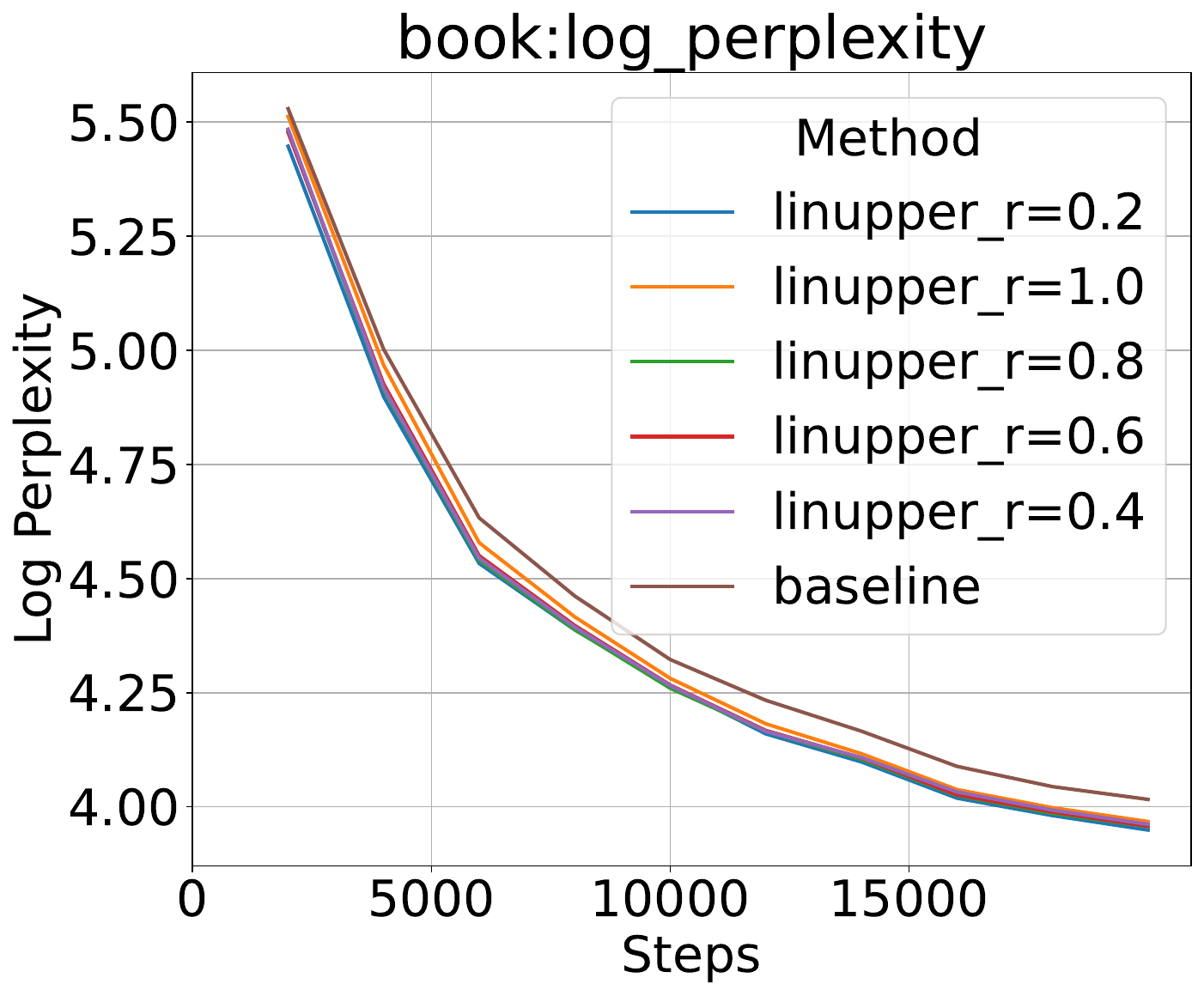}
    \end{subfigure}
    \begin{subfigure}{0.24\textwidth}
        \includegraphics[width=\linewidth]{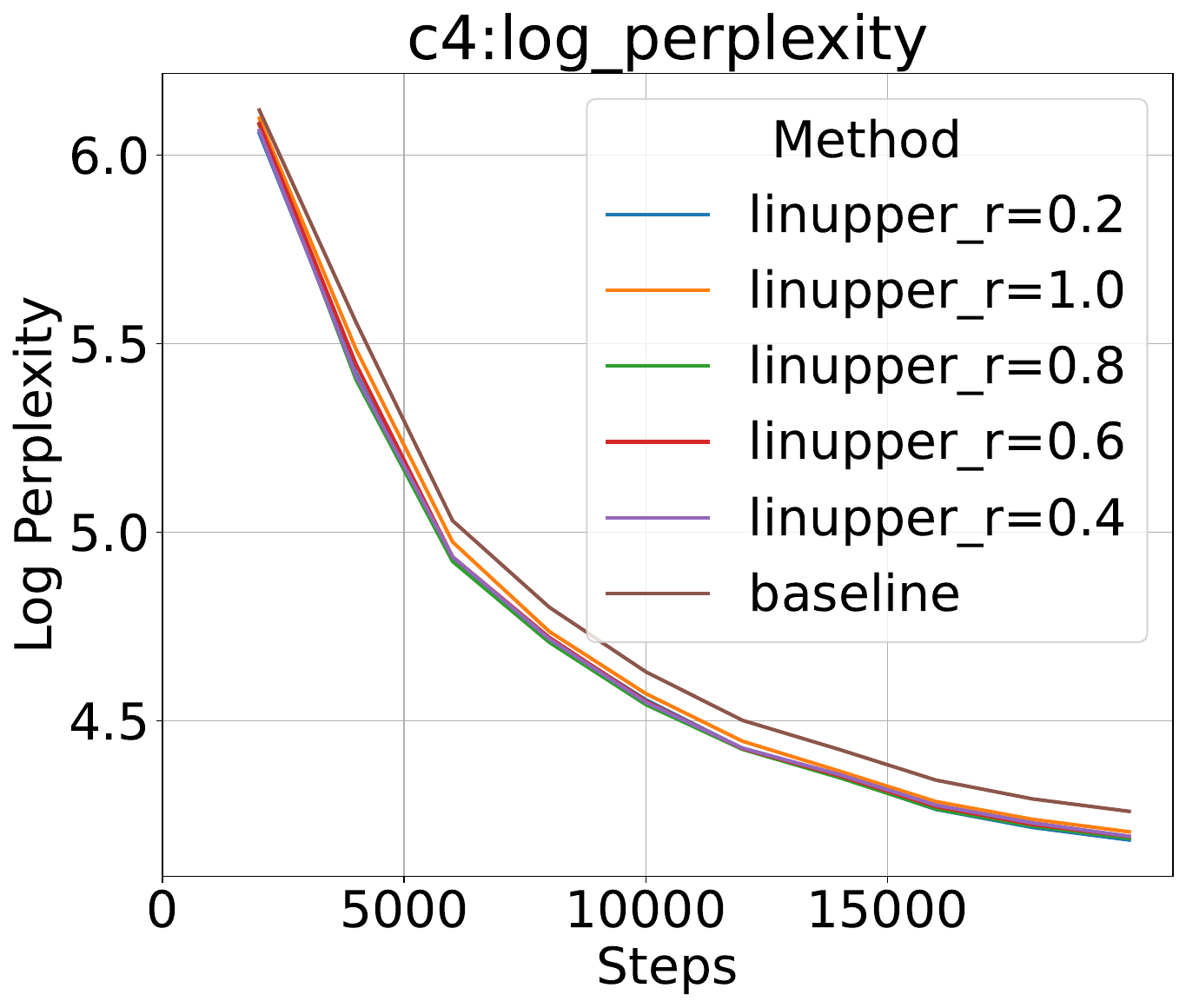}
    \end{subfigure}
    \begin{subfigure}{0.24\textwidth}
        \includegraphics[width=\linewidth]{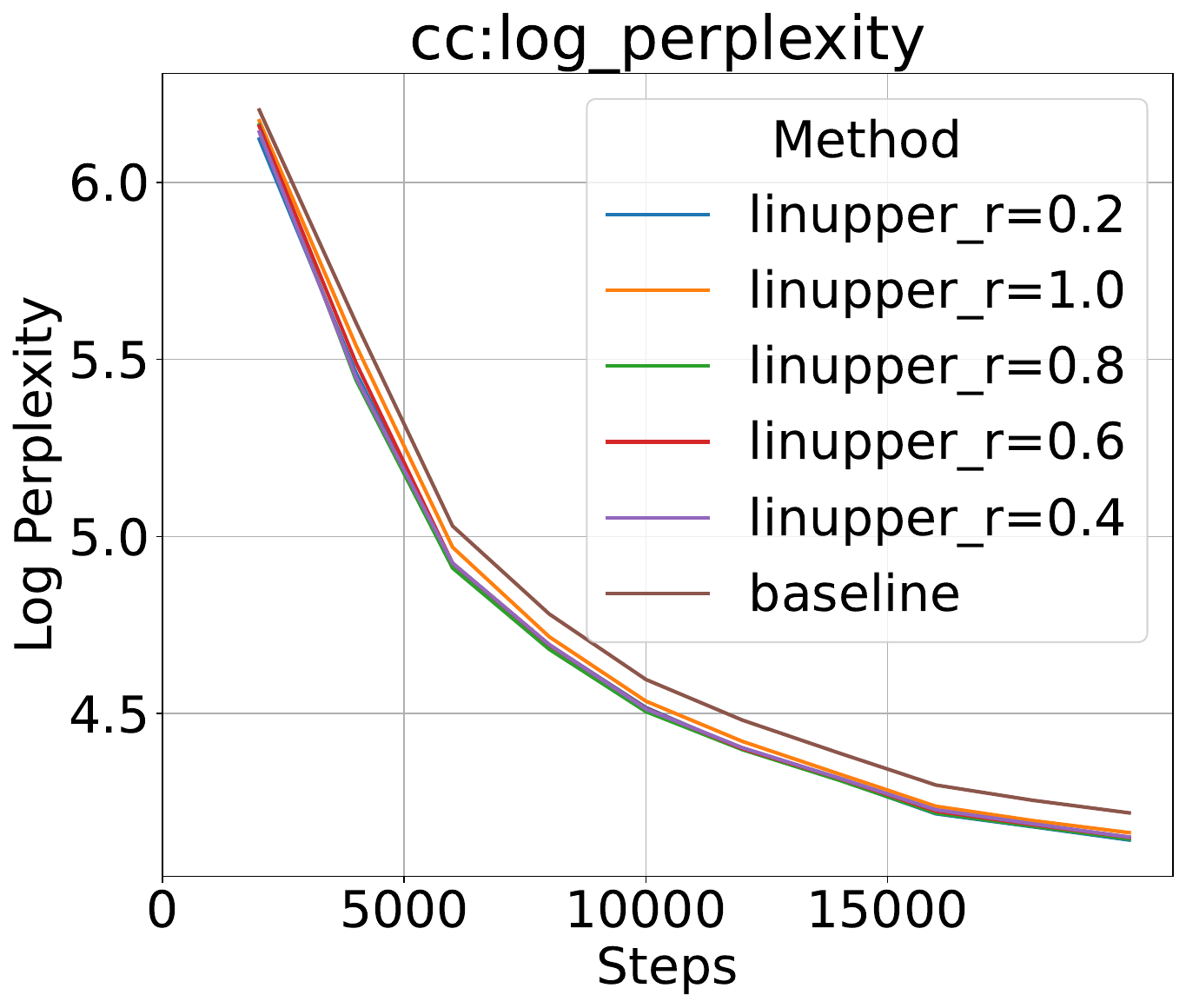}
    \end{subfigure}
    \begin{subfigure}{0.24\textwidth}
        \includegraphics[width=\linewidth]{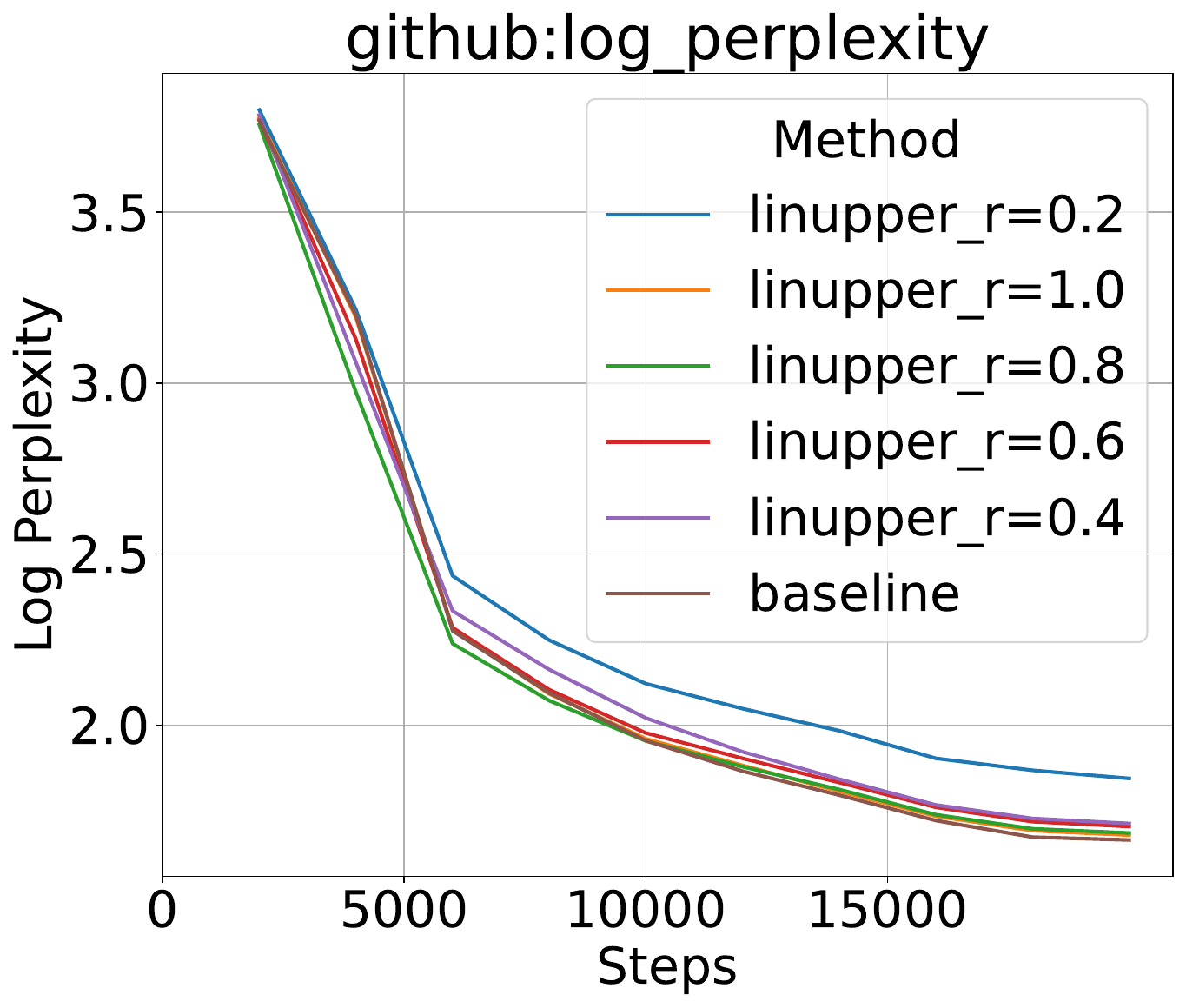}
    \end{subfigure}
    \begin{subfigure}{0.24\textwidth}
        \includegraphics[width=\linewidth]{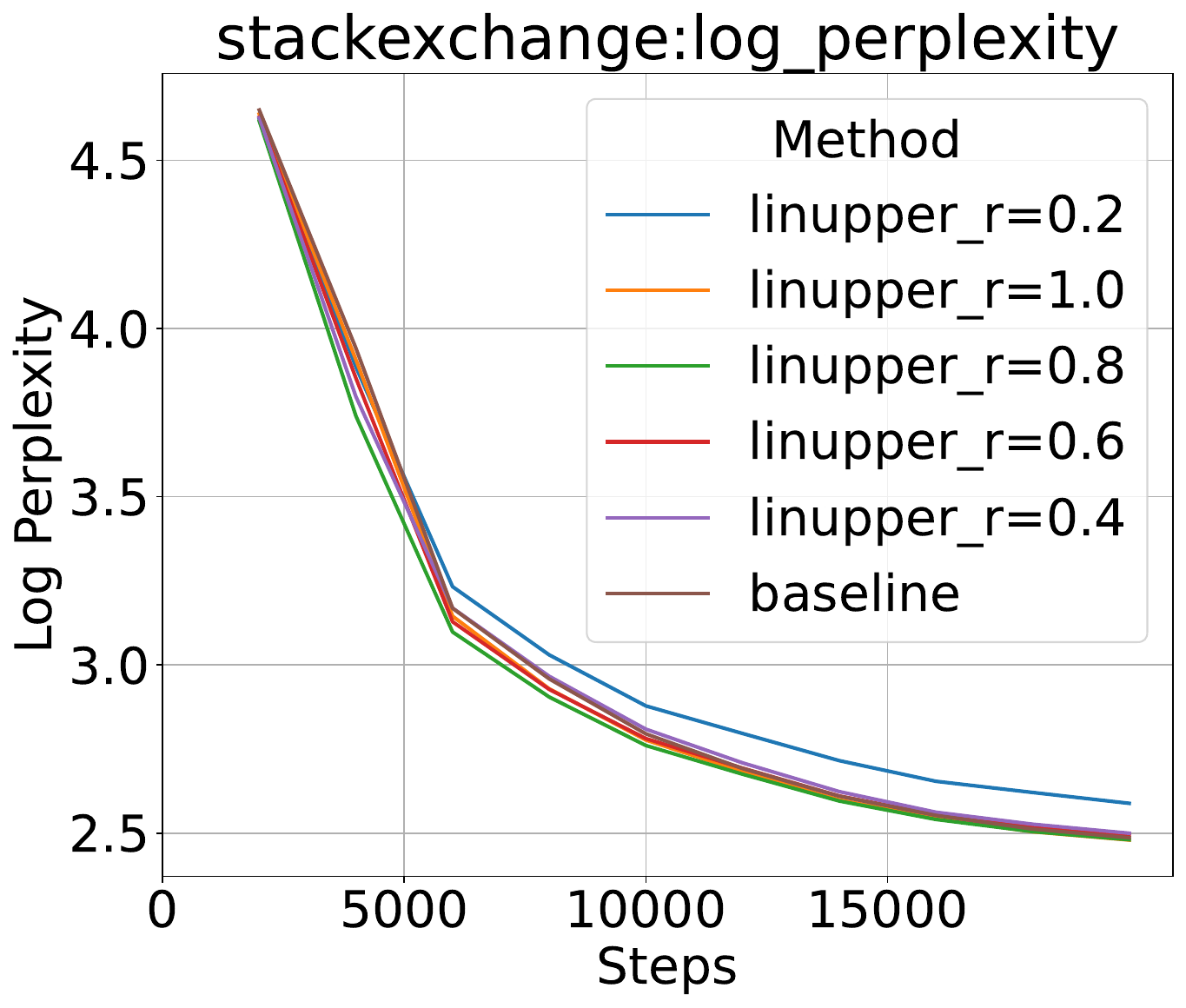}
    \end{subfigure}
    \begin{subfigure}{0.24\textwidth}
        \includegraphics[width=\linewidth]{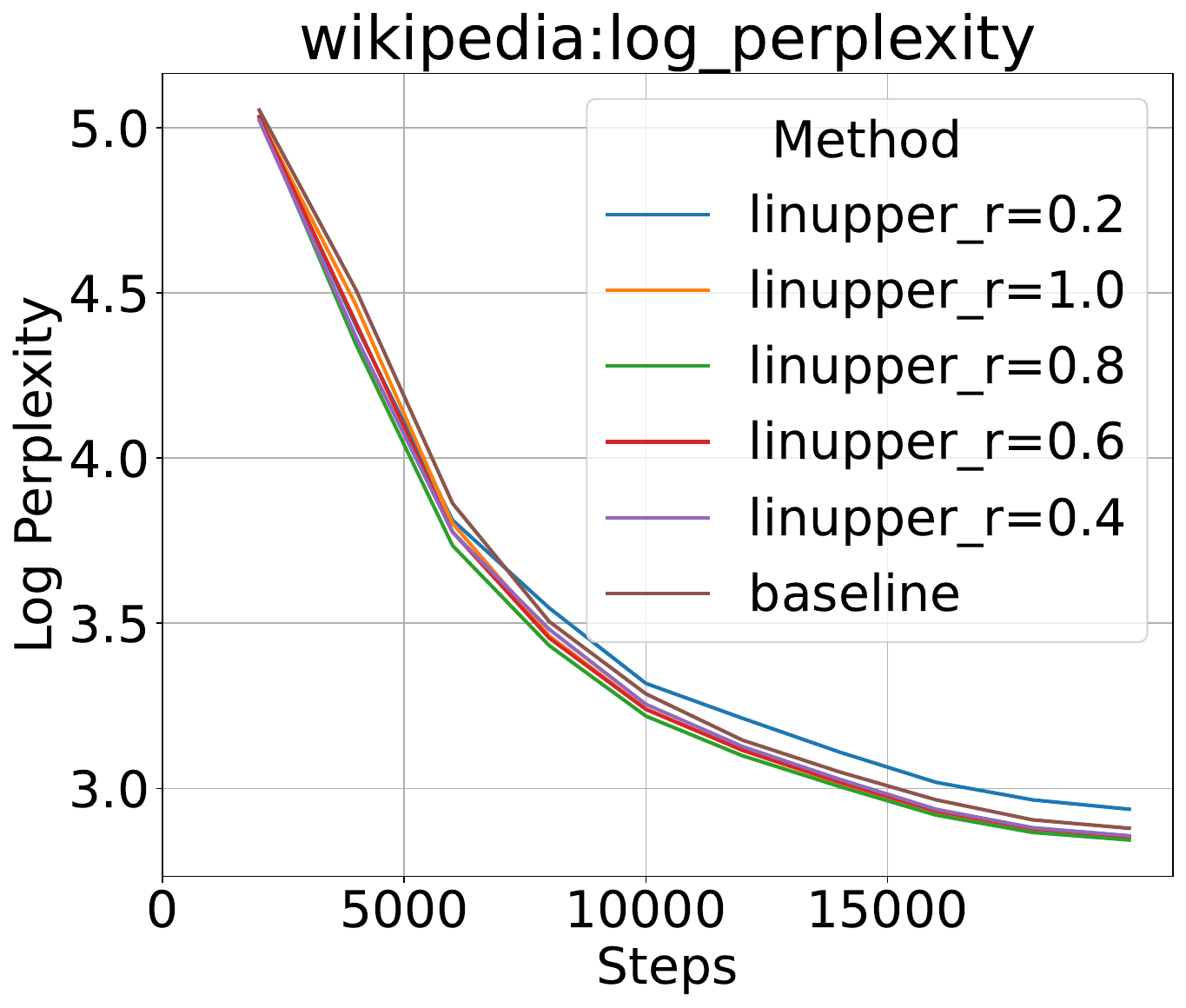}
    \end{subfigure}

    \caption{\textcolor{black}{Sensitivity of our method LinUpper to the value of $r$. When the $r$ is large (e.g. $r=1$) the performance of our method becomes closer to that of the uniform baseline. Decreasing the value of $r$ leads to diminished effect of low-loss samples, but can also have a negative effect on the performance when $r$ is too small (eg. $r=0.2$)}
    }
    \label{fig:perplexity_diff_r}
\end{figure}

We conduct experiments using different values of $r$ to understand the sensitivity of our methods to the value of $r$. The perplexity plots for out method \texttt{LinUpper} with different values of $r$ are provided in \Cref{fig:perplexity_diff_r}, which shows that when the value of $r$ is large (e.g. $r=1$) then, as expected, the performance of our method becomes closer to that of the uniform baseline. Decreasing the value of $r$ leads to the diminished effect of low-loss samples, but this can also have a negative effect on the performance when $r$ is too small (e.g. $r=0.2$), potentially due to data wastage by over filtering the low-loss samples. 

\subsection{Further Discussions About our Reweighting Upper Bound}
\label{sec:discussion_upperbound}
\begin{figure}[H]
\centering
\includegraphics[width=0.9\linewidth]{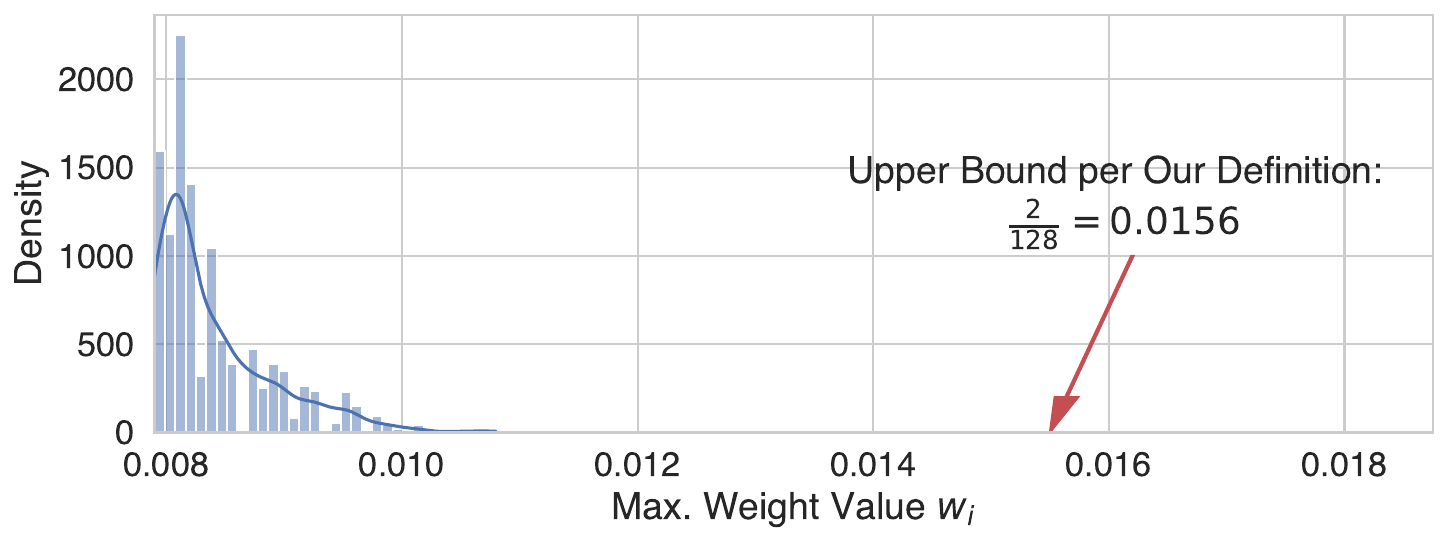}
\caption{\color{black}Distribution of maximum weight values when training our 7B parameter model with Llama architecture. The maximum weight values always stay well below the theoretical upper bound defined in \Cref{thm:minibatch_SGD_convex}. 
}
\label{fig:maxweightdist}
\end{figure}

To demonstrate that in practice the upper bound of $w_i < \frac{2}{b}$ is, in fact, satisfied by our LinUpper method, we provide the distribution of the maximum weights per step (see \Cref{fig:maxweightdist}) during the pretraining of the 7B parameter model with our LinUpper method. In this experiment, the minibatch size is $128$ and hence the upper bound is $\frac{2}{128} = 0.0156$. As depicted in \Cref{fig:maxweightdist}, the maximum weight per step (which is around $0.01$) by our LinUpper method is clearly below the upper bound of $0.0156$.

\clearpage

\section{Additional Discussions} \label{app:toy}
\subsection{Toy Regression Problem Setup}

To further illustrate the effectiveness of our reweighting strategies on simple ML problems, we apply them to a regression problem where the goal is to learn a linear mapping from input features to output labels, while dealing with noisy data and outliers. 
We generate a synthetic dataset as follows. 

The clean input features $\mathbf{X}_\text{clean} \in \mathbb{R}^{n \times p}$ are sampled from the standard normal distribution, where $p$ is the dimensionality of the data, $n$ is the number of clean data points. The target output $\mathbf{y} \in \mathbb{R}^{n}$ for the clean data is generated using the following ground truth linear model with added noise: 
\begin{equation}
\mathbf{y}_\text{clean} = \mathbf{X}_\text{clean} \mathbf{W}^* + b^* \mathbf{1}_n + c \cdot \boldsymbol{\epsilon}, \nonumber
\end{equation}

where $\mathbf{W}^* \in \mathbb{R}^p$ and $b^* \in \mathbb{R}$ are the true model parameters, $\boldsymbol{\epsilon} \sim \mathcal{N}(\mathbf{0}_n, \mathbf{I}_n)$ represents Gaussian noise, and $c$ is a small constant controlling the noise level. For the $m$ outlier data, we generate each input features of $\mathbf{X}_\text{ood} \in \mathbb{R}^{m \times p}$ as $0.1 \cdot \mathcal{N}(0, 1) + 2.0$
with the corresponding target outputs $\mathbf{y}_\text{ood}$ drawn randomly from a standard normal distribution, i.e., $\mathbf{y}_\text{ood} \sim \mathcal{N}(0, 1)$. 

The full training set consists of both the clean data $(\mathbf{X}_\text{clean}, \mathbf{y}_\text{clean})$ and the outlier data $(\mathbf{X}_\text{ood}, \mathbf{y}_\text{ood})$, concatenated together:
\begin{equation}
\mathbf{X}_\text{all} = [\mathbf{X}_\text{clean}; \mathbf{X}_\text{ood}], \quad \mathbf{y}_\text{all} = [\mathbf{y}_\text{clean}; \mathbf{y}_\text{ood}]. \nonumber
\end{equation}

By incorporating outliers into the training set, we test the robustness of our reweighting strategies, which should down-weight the influence of noisy or irrelevant samples, allowing the model to learn the underlying clean data distribution more effectively.

We define the objective function for this regression task as the mean squared error between the model predictions and the target outputs. Given model parameters $\mathbf{W} \in \mathbb{R}^p$ and $b \in \mathbb{R}$, the loss for a single data point $(\mathbf{x}_i, y_i)$ computed as:
\begin{equation}
\ell_i(\mathbf{W}, b) = \frac{1}{2} \left( \mathbf{x}_i^\top \mathbf{W} + b - y_i \right)^2.
\end{equation}

To evaluate model performance, we compute the mean squared error on a held-out test set $(\mathbf{X}_\text{test}, \mathbf{y}_\text{test})$, which was generated using the same process as the clean training data. We use $p=64$, $n=3200$, and $m=800$.

\subsection{Comparison Between Different Reweighting Methods}
\begin{figure}[H]
    \centering
    \begin{tabular}{cc}
    \includegraphics[width=6.5cm,height=4cm]{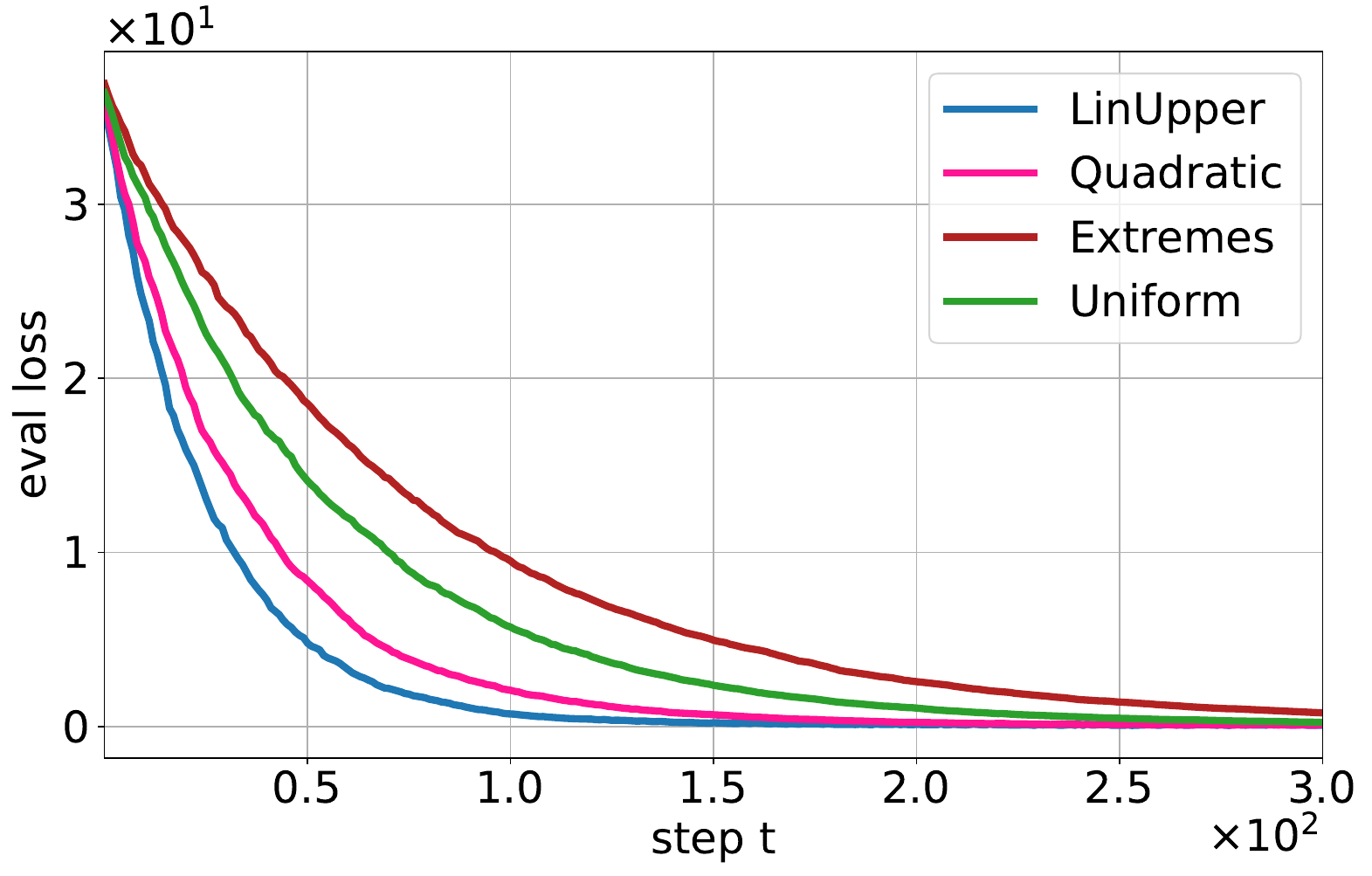}
    &\includegraphics[width=6.5cm,height=4cm]{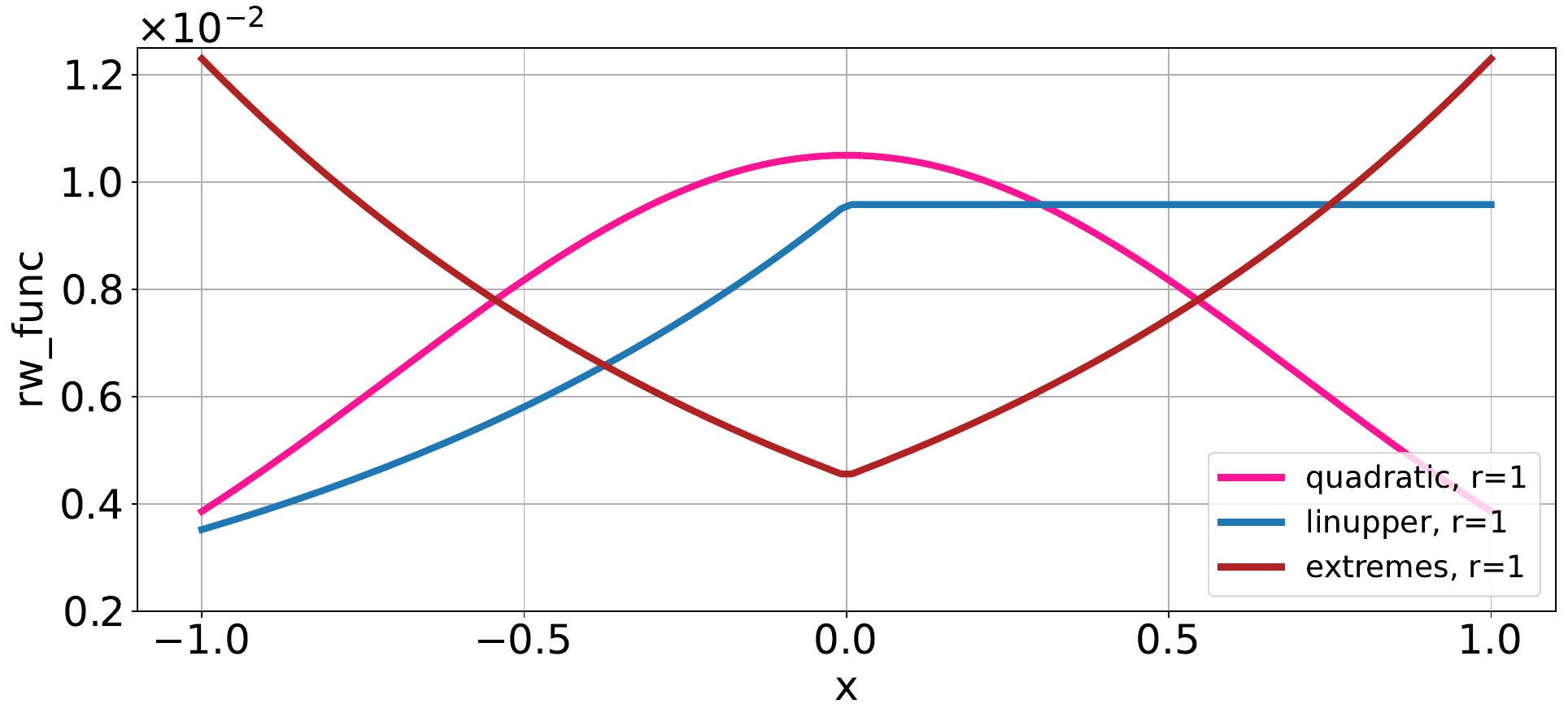}
    \end{tabular}
    \caption{
    \textbf{Left:} Test loss vs training steps for the toy regression problem. Our \texttt{LinUpper} strategy achieves faster convergence compared to all the other methods. Notably, the \texttt{Quadratic} strategy also outperforms the uniform baseline, further highlighting the benefit of selective reweighting. 
    \textbf{Right:} Shape of the different reweighting methods used with $r=1$.}
    \label{fig:toy}
\end{figure}

The results from \Cref{fig:toy} clearly show that \texttt{LinUpper} outperforms all other reweighting strategies and the uniform baseline, achieving the fastest convergence. This supports our theoretical findings that down-weighting low-loss samples allows the model to focus on more informative examples, thus accelerating the optimization. Interestingly, the \texttt{Quadratic} strategy also performs better than the uniform baseline, indicating that selectively up-weighting the moderate loss samples can improve performance. While \texttt{Quadratic} converges more slowly than \texttt{LinUpper}, it still highlights that more balanced reweighting can provide benefits over uniform sampling when dealing with noisy data.

\subsection{Comparison with Robust Optimization Methods}

\textcolor{black}{
Comparing with traditional robust optimization methods, such as those based on distributionally robust optimization (DRO), our methods impose a cap on sample weights ($w_i \leq 2/M$), as justified by our theoretical analysis (\Cref{prop:prop1}). This cap prevents over-reliance on a small subset of high-loss samples, which can lead to overfitting to outliers. In contrast, DRO methods often focus heavily on worst-case scenarios, potentially leading to suboptimal generalization when the training data contains a substantial amount of noisy samples. 
To demonstrate this benefit of our method in practice, we conducted experiments to compare with the traditional KL-divergence regularized distributionally robust optimization (DRO-KL). 
}

\begin{figure}[H]
\centering
\includegraphics[width=0.6\linewidth]{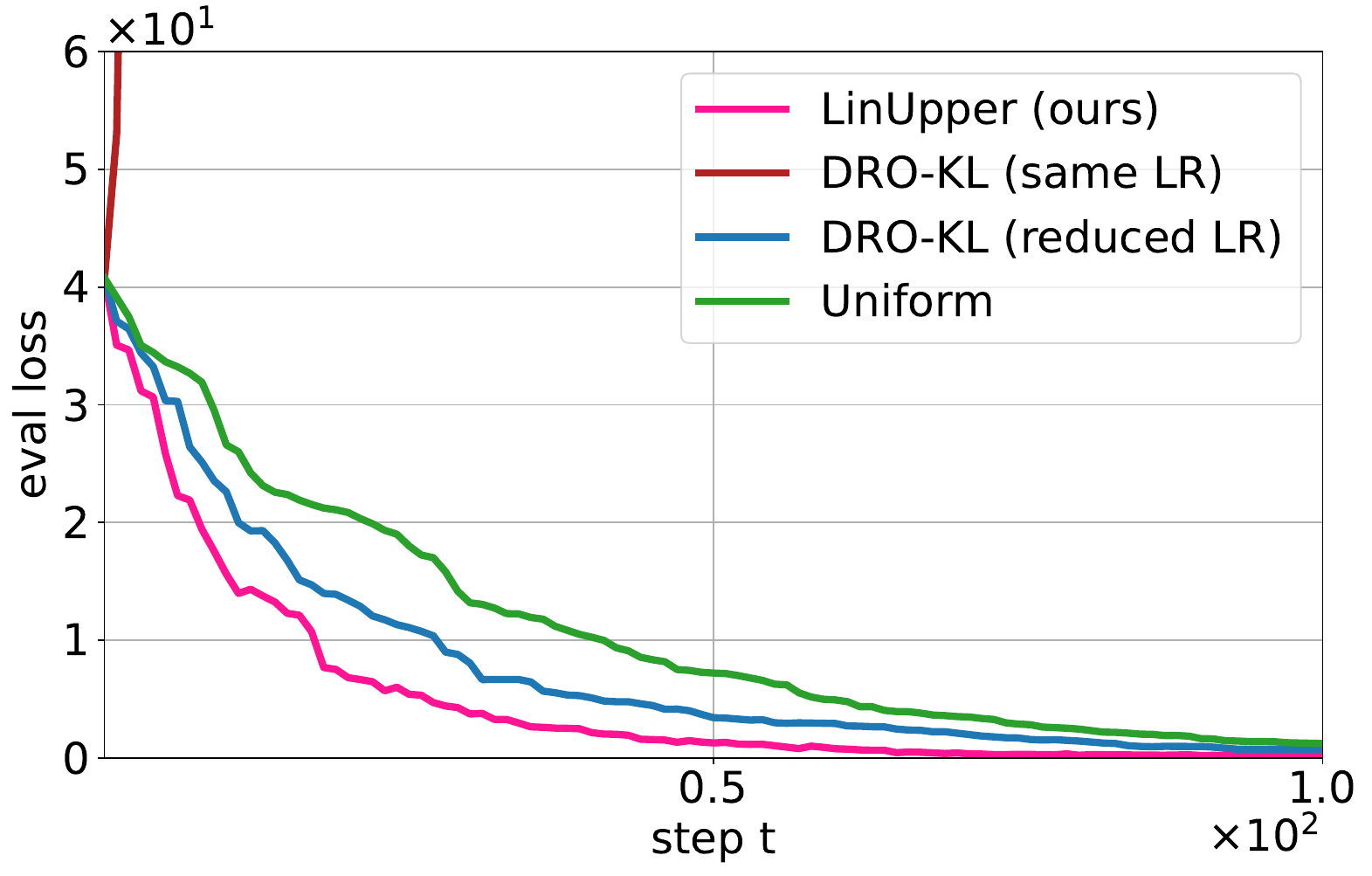}
\caption{\textcolor{black}{
Regression problem with \texttt{DRO-KL}.
}}
\label{fig:compare_dro}
\end{figure}

\textcolor{black}{
First, we created a synthetic regression problem (similar to the regression problem described above) with 25\% of the data being outliers to compare the robustness of the different methods against outliers. 
As shown in \Cref{fig:compare_dro}, we note the following: because it focuses heavily on the hard samples, \texttt{DRO-KL} diverges when we use the same learning rate as the other two methods (our \texttt{LinUpper} method \& the \texttt{Uniform} baseline). The \texttt{DRO-KL} method requires a smaller learning rate to converge, in which case it converges slower than our method. 
Furthermore, we include the \texttt{DRO-KL} method among the compared baselines for the GPT2-medium experiments, as shown in \Cref{tab:dro_kl_comparison}. On average, our \texttt{LinUpper} strategy outperforms the other \texttt{DRO-KL} configurations, which we attribute to its ability to handle outliers. 
}

\begin{table}[H]
    \centering
    \caption{\textcolor{black}{Comparison of our \texttt{LinUpper} loss reweighting strategy with various \texttt{DRO-KL} configurations on the GPT2-medium experimental setup. On average, our \texttt{LinUpper} strategy outperforms the other \texttt{DRO-KL} configurations.}}
    \label{tab:dro_kl_comparison}
    \resizebox{1\textwidth}{!}{
        \color{black}
        \begin{tabular}{lrrrrrrr}
            \toprule
            & Uniform & LinUpper (r = 0.4; ours) & DRO (KL = 0.4) & DRO (KL = 1.0) & DRO (KL = 2.0) & DRO (KL = 5.0) & DRO (KL = 10.0) \\
            \midrule
            LogiQA & 25.7 & \textbf{27.9} & 23.2 & 25.5 & 26.8 & 25.9 & 26.3 \\
            LogiQA2 & 27.5 & 27.6 & 24.4 & \textbf{28.7} & 27.5 & 28.2 & 27.6 \\
            SciQ & 49.0 & \textbf{51.8} & 41.1 & 48.6 & 49.4 & 50.5 & 49.4 \\
            PiQA & 55.5 & 56.2 & 54.0 & 56.2 & 55.7 & \textbf{56.5} & 55.4 \\ \midrule
            \textit{Mean} & \textit{39.4} & \textit{\textbf{40.9}} & \textit{35.7} & \textit{39.8} & \textit{39.9} & \textit{40.3} & \textit{39.7} \\
            \bottomrule
        \end{tabular}
    }
\end{table}

\clearpage

\section{Proofs}

\begin{lemma} \label{lemma:convlip}
Let Assumptions \ref{ass:conv} and \ref{ass:lip} hold. Then we have
\begin{equation*}
    \big\|\nabla f(\theta) - \nabla f(\theta')\big\|^2 \leq 2L \left(f(\theta) - f(\theta') - \langle \nabla f(\theta'), \theta - \theta' \rangle\right). 
\end{equation*}
\end{lemma}
\begin{proof}

Consider the following
\begin{align}
\label{eq:preliminary_step}
    f(\theta_1)-f(\theta_2)&=f(\theta_1)-f(\theta)+f(\theta)-f(\theta_2).
\end{align}
From the $L$-smoothness of $f$, we have for all $\theta_2$, $\theta \in \mathbb{R}^d$, 
\begin{equation}
\label{eq:smooth_condition}
    f(\theta) - f(\theta_2) \leq \langle \nabla f(\theta_2), \theta - \theta_2 \rangle + \frac{L}{2}  \big\| \theta - \theta_2\big\|^2.
\end{equation}
From the the convexity of $f$, we have for all $\theta_1$, $\theta \in \mathbb{R}^d$, 
\begin{equation}
\label{eq:convexity_condition}
    f(\theta_1) - f(\theta) \leq \langle \nabla f(\theta_1), \theta_1 - \theta \rangle. 
\end{equation}
Substituting \Cref{eq:smooth_condition}, \Cref{eq:convexity_condition} in \Cref{eq:preliminary_step}, 
we get
\begin{align}
\label{eq:preliminary_step2}
    f(\theta_1)-f(\theta_2)&\leq \langle \nabla f(\theta_1), \theta_1 - \theta \rangle + \langle \nabla f(\theta_2), \theta - \theta_2 \rangle + \frac{L}{2}  \big\| \theta - \theta_2\big\|^2.
\end{align}

Let $\theta = \theta_2 - \frac{1}{L} \left(\nabla f(\theta_2) - \nabla f(\theta_1)\right)$. Substituting the definition of $\theta$ in \Cref{eq:preliminary_step2}, we obtain
\begin{align}
\label{eq:preliminary_step3}
    f(\theta_1)-f(\theta_2)&\leq \langle \nabla f(\theta_1), \theta_1 - \theta_2+\frac{1}{L}\left(\nabla f(\theta_2)-\nabla f(\theta_1)\right) \rangle - \frac{1}{L}\langle \nabla f(\theta_2), \nabla f(\theta_2) - \nabla f(\theta_1) \rangle \nonumber \\&+ \frac{1}{2L}  \big\| \nabla f(\theta_2) - \nabla f(\theta_1)\big\|^2\nonumber \\
    &= \langle \nabla f(\theta_1), \theta_1 - \theta_2\rangle - \frac{1}{L}\| \nabla f(\theta_2) - \nabla f(\theta_1) \|^2 + \frac{1}{2L}  \big\| \nabla f(\theta_2) - \nabla f(\theta_1)\big\|^2\nonumber \\
    &= \langle \nabla f(\theta_1), \theta_1 - \theta_2\rangle - \frac{1}{2L}\| \nabla f(\theta_2) - \nabla f(\theta_1) \|^2.
\end{align}

\end{proof}
From \cref{lemma:convlip}, we obtain at the optimum $\theta^*$
\begin{equation*}
    \big\|\nabla f(\theta) - \nabla f(\theta^*)\big\|^2 \leq 2L \left(f(\theta) - f(\theta^*)\right). 
\end{equation*}

\subsection{Full Gradient}
\label{sec:full_gradient}
We have 
\begin{align}
    \big\| \theta^{t+1} - \theta^*\big\|^2 &= \big\| \theta^{t} - \theta^*\big\|^2 + \big\| \theta^{t+1} - \theta^{t}\big\|^2 + 2 \left\langle \theta^{t+1} - \theta^{t}, \theta^{t} - \theta^* \right\rangle \nonumber\\
    & = \big\|\theta^{t} - \theta^*\big\|^2 + \big\|\eta_t \sum_{i=1}^{M} w_{i,t} \nabla f_i(\theta^{t})\big\|^2 - 2 \left\langle \eta_t \sum_{i=1}^{M} w_{i,t} \nabla f_i(\theta^{t}), \theta^{t} - \theta^* \right\rangle \nonumber\\
    &\leq \big\|\theta^{t} - \theta^*\big\|^2 + \eta_t^2 M \sum_{i=1}^{M} w_{i,t}^2 \big\|\nabla f_i(\theta^{t})\big\|^2 - 2 \eta_t \sum_{i=1}^{M} w_{i,t} \left\langle \nabla f_i(\theta^{t}), \theta^{t} - \theta^* \right\rangle
\end{align}

Using the convexity of function $f_i$, we have
\begin{align}
\left\langle \nabla f_i(\theta^{t}), \theta^{t} - \theta^{*} \right\rangle \geq f_i(\theta^t) - f_i(\theta^{*}). 
\end{align}
Therefore combining, we obtain
\begin{align}
    \big\| \theta^{t+1} - \theta^*\big\|^2 
    &\leq \big\|\theta^{t} - \theta^*\big\|^2 + \eta_t^2 M \sum_{i=1}^{M} w_{i,t}^2 \big\|\nabla f_i(\theta^{t})\big\|^2 - 2 \eta_t \sum_{i=1}^{M} w_{i,t} \left( f_i(\theta^t) - f_i(\theta^{*}) \right) \label{eq:desc1}
\end{align}

Next, we upper-bound the quantity $\big\|\nabla f_i(\theta^{t})\big\|^2$. 
\begin{align}
    \big\|\nabla f_i(\theta^{t})\big\|^2 &\leq 2\big\|\nabla f_i(\theta^{t}) - \nabla f_i(\theta^{*})\big\|^2 + 2\big\|\nabla f_i(\theta^{*})\big\|^2 \nonumber \\
    &\leq 4L \left(f_i(\theta^t) - f_i(\theta^*)\right) + 2 \sigma_*^2 \label{eq:gradnorm}
\end{align}
where \Cref{eq:gradnorm} follows from \cref{lemma:convlip}. 

Now, combining \Cref{eq:desc1} and \Cref{eq:gradnorm}, we obtain
\begin{align}
    \big\| \theta^{t+1} - \theta^*\big\|^2 
    \leq& ~\big\|\theta^{t} - \theta^*\big\|^2 + 4ML\eta_t^2 \sum_{i=1}^{M} w_{i,t}^2 \left(f_i(\theta^t) - f_i(\theta^*)\right) +
    2M \eta_t^2 \sigma_*^2 \sum_{i=1}^{M} w_{i,t}^2 \\
    &- 2 \eta_t \sum_{i=1}^{M} w_{i,t} \left( f_i(\theta^t) - f_i(\theta^{*}) \right) \nonumber \\
    =& ~\big\|\theta^{t} - \theta^*\big\|^2 + \sum_{i=1}^{M} \left(4ML\eta_t w_{i,t} -2\right)\eta_t w_{i,t}\left(f_i(\theta^t) - f_i(\theta^*)\right) \nonumber \\
    & + 2M \eta_t^2 \sigma_*^2 \sum_{i=1}^{M} w_{i,t}^2, \label{eq:desc2}
\end{align}
Selecting $\eta_t$ such that $4ML\eta_t w_{i,t} - 2 \leq -1 ~~ \forall ~i$, i.e., $\eta_t \leq \frac{1}{4MLw_{\textrm{max},t}}$, yields
\begin{align}
\big\| \theta^{t+1} - \theta^*\big\|^2 
    \leq& ~\big\|\theta^{t} - \theta^*\big\|^2 -\eta_t \sum_{i=1}^{M} w_{i,t}\left(f_i(\theta^t) - f_i(\theta^*)\right) + 2M \eta_t^2 \sigma_*^2, 
\end{align}
where we also used the fact that $\sum_{i=1}^{M} w_{i,t}^2 \leq 1$ due to $\sum_{i=1}^{M} w_{i,t} = 1$ and $w_{i,t}\geq0$. Hence, 
\begin{align}
\big\| \theta^{t+1} & - \theta^*\big\|^2 \nonumber \\
\leq& ~\big\|\theta^{t} - \theta^*\big\|^2 - \frac{\eta_t}{M}\sum_{i=1}^{M} \left(f_i(\theta^t) - f_i(\theta^*)\right) + \eta_t\sum_{i=1}^{M} \left(\frac{1}{M} - w_{i,t}\right)\left(f_i(\theta^t) - f_i(\theta^*)\right) + 2M \eta_t^2 \sigma_*^2 \nonumber\\
=& ~\big\|\theta^{t} - \theta^*\big\|^2 - \eta_t \left(f(\theta^t) - f(\theta^*)\right) + \eta_t\sum_{i=1}^{M} \left(\frac{1}{M} - w_{i,t}\right)\left(f_i(\theta^t) - f_i(\theta^*)\right) + 2M \eta_t^2 \sigma_*^2. \label{eq:desc3}
\end{align}
Telescoping \Cref{eq:desc3} from $t=0$ to $T-1$ yields
\begin{align}
   \big\| \theta^{T} - \theta^*\big\|^2 \leq & ~\big\| \theta^{0} - \theta^*\big\|^2 - \sum_{t=0}^{T-1} \eta_t \left(f(\theta^t) - f(\theta^*)\right) + \sum_{t=0}^{T-1} \eta_t\sum_{i=1}^{M} \left(\frac{1}{M} - w_{i,t}\right)\left(f_i(\theta^t) - f_i(\theta^*)\right) \nonumber \\
   &+ 2M \sigma_*^2 \sum_{t=0}^{T-1}\eta_t^2. 
\end{align}
For a weighting scheme such that $w_{\textrm{max},t} \leq \frac{2}{M}$ so that $\eta_t$ can be fixed to $\eta = \frac{1}{8L}$, we obtain
\begin{align}
   \eta \sum_{t=0}^{T-1} \left(f(\theta^t) - f(\theta^*)\right) \leq & ~\big\| \theta^{0} - \theta^*\big\|^2 + \eta \sum_{t=0}^{T-1} \sum_{i=1}^{M} \left(\frac{1}{M} - w_{i,t}\right)\left(f_i(\theta^t) - f_i(\theta^*)\right) + 2M \sigma_*^2 \eta^2 T. \nonumber
\end{align}
\begin{align}
   f(\bar \theta^T) - f(\theta^*) \leq & ~\frac{8L\big\| \theta^{0} - \theta^*\big\|^2}{T} + \frac{1}{T}\sum_{t=0}^{T-1} \sum_{i=1}^{M} \left(\frac{1}{M} - w_{i,t}\right)\left(f_i(\theta^t) - f_i(\theta^*)\right) + \frac{M \sigma_*^2}{4L}, \nonumber
\end{align}
where $\bar \theta^T = \frac{1}{T}\sum_{t=0}^{T-1}\theta^t$. Define $\delta^t = \sum_{i=1}^{M} \left(\frac{1}{M} - w_{i,t}\right)\left(f_i(\theta^t) - f_i(\theta^*)\right)$. Hence, we obtain 
\begin{align}
   f(\bar \theta^T) - f(\theta^*) \leq & ~\frac{8L\big\| \theta^{0} - \theta^*\big\|^2}{T} + \frac{1}{T}\sum_{t=0}^{T-1} \delta^t + \frac{M \sigma_*^2}{4L}. 
\end{align}

\subsection{Optimal Weight Strategy}
\label{sec:optimal_weights}
\weightbound*
\begin{proof}
    For the sake of clarity, we drop the index $t$ in $\delta^t$ in the following proof. 
    
    Let $\Delta_i = f(\mathbf{x}_i; \theta^{t}) - f(\mathbf{x}_i; \theta^{*})$ be the loss gap. After dropping all terms that do not depends on the weights $w_i$'s, we have the following constrained optimization problem
    \begin{align}
        &\min_{w_1,\ldots,w_M}\hspace{1cm}-\sum_{i=1}^Mw_i\Delta_i + r\sum_{i=1}^Mw_i\log w_i\\
        &\text{ subject to }\hspace{1.5cm} w_i\leq \frac{2}{M}, i\in [M],\\
        &\hspace{3.5cm}\sum_{i=1}^Mw_i=1.
    \end{align}
    As we have inequality constraints in the above problem, we introduce the slack variable $t$ and consider the Lagrangian function to solve the above problem as follows:
    \begin{equation}
        L(w_1,\ldots,w_M,\nu,\mu)=-\sum_{i=1}^Mw_i\Delta_i+r\sum_{i=1}^Mw_i\log w_i-\nu\left(\sum_{i=1}^Mw_i-1\right)-\mu\left(w_i-\frac{2}{M}+t^2\right),
    \end{equation}
    and we have the gradients
    \begin{align}
        \frac{\partial L}{\partial w_i}&=-\Delta_i+r+r\log w_i-\nu-\mu\nonumber \\
        \frac{\partial L}{\partial \lambda}&=-\left(\sum_{i=1}^M w_i-1\right)\nonumber \\
        \frac{\partial L}{\partial \mu}&=-\left(w_i-\frac{2}{M}+t^2\right).
    \end{align}

     We consider the complementary slackness condition. We have one slack variable $t$, and the corresponding Lagrange multiplier is $\nu$. We now consider whether a slack variable is zero (which the corresponding inequality constraint is active) or the Lagrange multiplier is zero (the constraint is inactive), we obtain
     \begin{itemize}
         \item $\mu=0,t^2>0$: Using $\frac{\partial L}{\partial \nu}=0$, we get $\sum_{i=1}^Mw_i=1$. Using $\frac{\partial L}{\partial w_i}=0,\mu=0$, we get $-\Delta_i+r+r\log w_i-\nu=0$. Hence, we have $w_i=\exp\left(\frac{\nu+\Delta_i-r}{r}\right)$. Therefore, we obtain $\nu=-r\ln\left(\sum_{i=1}^M\exp\left(\frac{\Delta_i-r}{r}\right)\right)$ yielding $w_i=\exp\left(\frac{-r\ln\left(\sum_{i=1}^M\exp\left(\frac{\Delta_i-r}{r}\right)\right)+\Delta_i-r}{r}\right)=C\exp\left(\frac{\Delta_i}{r}\right)$ where $C=\exp\left({-\ln\left(\sum_{i=1}^M\exp\left(\frac{\Delta_i-r}{r}\right)\right)-1}\right)$.
         \item $\mu\neq 0,t^2=0$: Using $\frac{\partial L}{\partial \mu}=0,t=0$, we get $w_i=\frac{2}{M}$ and $\mu+\nu=-\Delta_i+r+r\log\frac{2}{M}$.
     \end{itemize}
     Hence, we obtain $w_i=\min\left\{C\exp\left(\frac{\Delta_i}{r}\right),\frac{2}{M}\right\}$.
\end{proof}

\subsection{Convex Minibatch SGD}
\label{sec:convex_minibatch}
\renewcommand{\minibatchSGDconvex}{\begin{theorem}[Re-statement of \Cref{thm:minibatch_SGD_convex}]
\label{thm:minibatch_SGD_convex_}
Let Assumptions 
\ref{ass:conv}, \ref{ass:lip}, and \ref{ass:interp} hold. Consider a minibatch of size $|\mathcal{B}|=b$ with reweighting scheme satisfying $\max_{i\in \mathcal{B}}w_{i,t}\leq 2/b$. Then, 
\begin{itemize}
    \item {\bf minibatch SGD:} 
    for $\eta = \frac{1}{8L\sqrt{T}}$, we have 
    \begin{align}
    \label{eq:minibatch_convex_bound_restated}
   \mathbb{E}[f(\bar \theta^T) - f(\theta^*)] \leq & ~\frac{8L\big\| \theta^{0} - \theta^*\big\|^2}{\sqrt{T}} +\frac{1}{T}\sum_{t=0}^{T-1}\delta_t,
\end{align}
    \item {\bf minibatch SGD with momentum:} 
    for $\eta = \frac{1}{8L\sqrt{T+1}}, \lambda_t>0$, we have 
    \begin{align}
    \label{eq:minibatch_momentum_convex_bound_restated}
  \mathbb{E}[f( \theta^T) - f(\theta^*)] &\leq\frac{8L\big\| \theta^{0} - \theta^*\big\|^2}{\sqrt{T+1}} +\frac{2}{{T+1}}\sum_{t=0}^{T-1}{\delta_t} + \frac{2}{{T+1}}\sum_{t=0}^{T-1}{\lambda_t\mu_{t}},
\end{align}
\end{itemize}    
where $\delta_t=\mathbb{E}\left[ \sum_{i\in \mathcal{B}} \left(\frac{1}{b}-w_{i,t}\right) ( f_i(\theta^{t})-f_i(\theta^*))\right]$, $\mu_t=\mathbb{E}\big[ \sum_{i\in \mathcal{B}} \left(\frac{1}{b}-w_{i,t}\right) ( f_i(\theta^{t})-f_i(\theta^{t-1}))\big]$, and $\bar \theta^T = \frac{1}{T}\sum_{t=0}^{T-1}\theta^t$.
\end{theorem}}

\minibatchSGDconvex
\begin{proof}
(i) {\bf Minibatch SGD:} For the minibatch SGD based on \Cref{eq:ours}, we have 
\begin{align}
    \big\| \theta^{t+1} - \theta^*\big\|^2 &= \big\| \theta^{t} - \theta^*\big\|^2 + \big\| \theta^{t+1} - \theta^{t}\big\|^2 + 2 \left\langle \theta^{t+1} - \theta^{t}, \theta^{t} - \theta^* \right\rangle \nonumber\\
    & = \big\|\theta^{t} - \theta^*\big\|^2 + \big\|\eta_t \sum_{i\in \mathcal{B}} w_{i,t} \nabla f_i(\theta^{t})\big\|^2 - 2 \left\langle \eta_t \sum_{i\in \mathcal{B}} w_{i,t} \nabla f_i(\theta^{t}), \theta^{t} - \theta^* \right\rangle.
\end{align}
Applying expectation on both sides, we get
\begin{align}
     &\mathbb{E}\big\| \theta^{t+1} - \theta^*\big\|^2 \nonumber\\
    &\leq \big\|\theta^{t} - \theta^*\big\|^2 + \eta_t^2 \mathbb{E}  \big\|\sum_{i\in \mathcal{B}} w_{i,t}\nabla f_i(\theta^{t})\big\|^2 - 2 \eta_t \mathbb{E} \left\langle \sum_{i\in \mathcal{B}} w_{i,t} \nabla f_i(\theta^{t}), \theta^{t} - \theta^* \right\rangle\nonumber\\
    &\overset{(a)}{\leq} \big\|\theta^{t} - \theta^*\big\|^2 + \eta_t^2 \mathbb{E}  \big\|\sum_{i\in \mathcal{B}} w_{i,t}\nabla f_i(\theta^{t})\big\|^2 - 2 \eta_t \mathbb{E}\sum_{i\in \mathcal{B}} w_{i,t} \left( f_i(\theta^{t}) - f_i(\theta^*) \right)\nonumber\\
    &\leq \big\|\theta^{t} - \theta^*\big\|^2 + \eta_t^2b \mathbb{E}\left[ \sum_{i\in \mathcal{B}} w_{i,t}^2 \big\|\nabla f_i(\theta^{t})\big\|^2\right] - 2 \eta_t \mathbb{E}\sum_{i\in \mathcal{B}} w_{i,t} \left( f_i(\theta^{t}) - f_i(\theta^*) \right)\nonumber\\
    &\overset{(b)}{\leq} \big\|\theta^{t} - \theta^*\big\|^2 + 4L\eta_t^2b \mathbb{E}\left[ \sum_{i\in \mathcal{B}} w_{i,t}^2 ( f_i(\theta^{t})-f_i(\theta^*))\right] - 2 \eta_t \mathbb{E}\sum_{i\in \mathcal{B}} w_{i,t} \left( f_i(\theta^{t}) - f_i(\theta^*) \right)\nonumber\\
    &\leq \big\|\theta^{t} - \theta^*\big\|^2 +  \mathbb{E}\left[ \sum_{i\in \mathcal{B}} w_{i,t}(4L\eta_t^2bw_{i,t}-2\eta_t) ( f_i(\theta^{t})-f_i(\theta^*))\right], 
    \label{eq:desc1_mb}
\end{align}

where (a) follows from \Cref{ass:conv}, and (b) follows from \cref{lemma:convlip}. Selecting $\eta_t$ such that $4bL\eta_t w_{i,t} - 2 \leq -1 ~~ \forall ~i$, i.e., $\eta_t \leq \frac{1}{4bLw_{\textrm{max},t}}$, yields
\begin{align}
    \mathbb{E}\big\| \theta^{t+1} - \theta^*\big\|^2 
    \leq& ~\big\|\theta^{t} - \theta^*\big\|^2 
    -\eta_t\mathbb{E}\left[ \sum_{i\in \mathcal{B}} w_{i,t} ( f_i(\theta^{t})-f_i(\theta^*))\right] 
    \nonumber \\
    &=\big\|\theta^{t} - \theta^*\big\|^2 
    -\eta_t\mathbb{E}\left[ \sum_{i\in \mathcal{B}} \frac{1}{b} ( f_i(\theta^{t})-f_i(\theta^*))\right]  \nonumber \\
    &+\eta_t\mathbb{E}\left[ \sum_{i\in \mathcal{B}} \left(\frac{1}{b}-w_{i,t}\right) ( f_i(\theta^{t})-f_i(\theta^*))\right] 
    \nonumber \\
    &=\big\|\theta^{t} - \theta^*\big\|^2 
    -\eta_t ( f(\theta^{t})-f(\theta^*)) +\eta_t\delta_t,
    \label{eq:desc2_mb}
\end{align}
where $\delta_t=\mathbb{E}\left[ \sum_{i\in \mathcal{B}} \left(\frac{1}{b}-w_{i,t}\right) ( f_i(\theta^{t})-f_i(\theta^*))\right]$ and $\sum_{i\in S}w_{i,t}^2\leq 1$. 

Hence, telescoping \Cref{eq:desc2_mb} from $t=0$ to $T-1$ yields
\begin{align}
   \mathbb{E}\big\| \theta^{T} - \theta^*\big\|^2 \leq & ~\big\| \theta^{0} - \theta^*\big\|^2 - \sum_{t=0}^{T-1} \eta_t \left(f(\theta^t) - f(\theta^*)\right) + \sum_{t=0}^{T-1} \eta_t\delta_t.
\end{align}
For a weighting scheme such that $w_{\textrm{max},t} \leq \frac{2}{b}$ so that $\eta_t$ can be fixed to $\eta = \frac{1}{8L\sqrt{T}}$, we obtain
\begin{align}
   f(\bar \theta^T) - f(\theta^*) \leq & ~\frac{8L\big\| \theta^{0} - \theta^*\big\|^2}{\sqrt{T}} +\frac{1}{T}\sum_{t=0}^{T-1}\delta_t, 
\end{align}
where $\bar \theta^T = \frac{1}{T}\sum_{t=0}^{T-1}\theta^t$.

The traditional convergence bound of minibatch SGD for convex functions can be recovered when $\delta_t = 0$ which occurs when the samples in the batch have uniform weights, $w_{i,t} = \frac{1}{b} ~ \forall i\in \mathcal{B}$.  We obtain a looser convergence bound when $\delta_t \geq 0$ resulting from the assignment of larger weights to smaller loss gaps. Moreover, down-weighting the importance of samples with low loss gaps leads to a better bound ($\delta_t \leq 0$). Note that the bound in \Cref{eq:minibatch_convex_bound_restated} only holds for reweighting schemes that satisfy $w_{i,t} \leq 2/b$. Thus, enforcing an upper bound on the maximum weight that can be assigned to a single data point. Typically, the minibatch gradient with uniform weights is unbiased, $\mathbb{E}\left[\sum_{i\in \mathcal{B}}\frac{1}{b}\nabla f_i(\theta^t)\right]=\nabla f(\theta^t)$. However, in our setting, the importance weights are functions of the sample functions. Hence, $\mathbb{E}\left[\sum_{i\in \mathcal{B}}w_{i,t}\nabla f_i(\theta^t)\right]\neq\nabla f(\theta^t)$ leading to the analysis of \Cref{thm:minibatch_SGD_convex} being more involved. Based on our analysis, we note that the step size $\eta_t$ is a function of the maximum value of the weights, $w_{\text{max},t}=\frac{2}{b}$ which leads to the choice of $\eta=\frac{1}{8L\sqrt{T}}$ for convergence.

(ii) {\bf Minibatch SGD with Momentum:} Given the loss-based sample reweighting strategies, \Cref{eq:minibatch_momentum_convex_bound_restated} characterizes the effects of loss-based reweighting on the convergence bound of reweighted minibatch SGD with momentum when the sample losses $f_i$ are convex. Given the loss-based sample reweighting strategies, we incorporate heavy ball momentum method to improve the convergence result and to characterize the effects of loss-based reweighting on the convergence bound. Hence, we have the following update equations:
\begin{align}
    z_{t+1}&=z_t-\eta_t\sum_{i\in \mathcal{B}}w_{i,t}\nabla f_i(\theta_t),\\
    \theta_{t+1}&=\frac{\lambda_{t+1}}{1+\lambda_{t+1}}\theta_t+\frac{1}{1+\lambda_{t+1}}z_{t+1},  
    \label{eq:momentum_equations}
\end{align}
where $\eta_t$ is the step size and $\eta_t,\lambda_t>0$. The above iterates $\theta_t$ are equal to the iterates of the following stochastic minibatch heavy ball momentum method \cite{sebbouh2021almost}.
$$\theta_{t+1}=\theta_t-\frac{\eta_t}{1+\lambda_{t+1}}\sum_{i\in \mathcal{B}}w_{i,t}\nabla f_i(\theta_t)+\frac{\lambda_t}{1+\lambda_{t+1}}(\theta_t-\theta_{t-1}).$$
\Cref{eq:minibatch_momentum_convex_bound_restated} characterizes the effects of loss-based reweighting on the convergence bound of reweighted minibatch SGD with momentum when the sample losses $f_i$ are convex.

 For the minibatch SGD with momentum based on \Cref{eq:momentum_equations}, we have

\begin{align}
    \big\| z^{t+1} - \theta^*\big\|^2 &= \big\| z^{t} - \theta^*\big\|^2 + \big\| z^{t+1} - z^{t}\big\|^2 + 2 \left\langle z^{t+1} - z^{t}, z^{t} - \theta^* \right\rangle \nonumber\\
    & = \big\|z^{t} - \theta^*\big\|^2 + \big\|\eta_t \sum_{i\in \mathcal{B}} w_{i,t} \nabla f_i(\theta^{t})\big\|^2 - 2 \left\langle \eta_t \sum_{i\in \mathcal{B}} w_{i,t} \nabla f_i(\theta^{t}), z^{t} - \theta^* \right\rangle.
\end{align}
Applying expectation on both sides, we get
\begin{align}
    & \mathbb{E}\big\| z^{t+1} - \theta^*\big\|^2 \nonumber\\
    &\leq \big\|z^{t} - \theta^*\big\|^2 + \eta_t^2 \mathbb{E}  \big\|\sum_{i\in \mathcal{B}} w_{i,t}\nabla f_i(\theta^{t})\big\|^2 - 2 \eta_t \mathbb{E} \left[\left\langle \sum_{i\in \mathcal{B}} w_{i,t} \nabla f_i(\theta^{t}), \theta^{t} - \theta^* \right\rangle\right]\nonumber\\
    &-2 \eta_t\lambda_t \mathbb{E}\left[ \left\langle \sum_{i\in \mathcal{B}} w_{i,t} \nabla f_i(\theta^{t}), \theta^{t} - \theta^{t-1} \right\rangle\right]\nonumber \\
    &\overset{(a)}{\leq} \big\|z^{t} - \theta^*\big\|^2 + \eta_t^2 \mathbb{E}  \big\|\sum_{i\in \mathcal{B}} w_{i,t}\nabla f_i(\theta^{t})\big\|^2 - 2 \eta_t \mathbb{E}\left[\sum_{i\in \mathcal{B}} w_{i,t} \left( f_i(\theta^{t}) - f_i(\theta^*) \right)\right]\nonumber\\
    &- 2 \eta_t\lambda_t \mathbb{E}\left[\sum_{i\in \mathcal{B}} w_{i,t} \left( f_i(\theta^{t}) - f_i(\theta^{t-1}) \right)\right]\nonumber \\
    &\overset{(b)}{\leq} \big\|z^{t} - \theta^*\big\|^2 +  \mathbb{E}\left[ \sum_{i\in \mathcal{B}} w_{i,t}(4L\eta_t^2bw_{i,t}-2\eta_t) ( f_i(\theta^{t})-f_i(\theta^*))\right]\nonumber \\
    &- 2 \eta_t\lambda_t \mathbb{E}\left[\sum_{i\in \mathcal{B}} w_{i,t} \left( f_i(\theta^{t}) - f_i(\theta^{t-1}) \right)\right]\nonumber \\
    &= \big\|z^{t} - \theta^*\big\|^2 +  \mathbb{E}\left[ \sum_{i\in \mathcal{B}} w_{i,t}(4L\eta_t^2bw_{i,t}-2\eta_t-2\eta_t\lambda_t) ( f_i(\theta^{t})-f_i(\theta^*))\right] \nonumber \\
    &+ 2 \eta_t\lambda_t \mathbb{E}\left[\sum_{i\in \mathcal{B}} w_{i,t} \left( f_i(\theta^{t-1}) - f_i(\theta^*) \right)\right]\nonumber \\
    &\overset{(c)}{\leq} \big\|z^{t} - \theta^*\big\|^2 -2\eta_t\lambda_{t+1}  \mathbb{E}\left[ \sum_{i\in \mathcal{B}} w_{i,t} ( f_i(\theta^{t})-f_i(\theta^*))\right] 
    + 2 \eta_t\lambda_t \mathbb{E}\left[\sum_{i\in \mathcal{B}} w_{i,t} \left( f_i(\theta^{t-1}) - f_i(\theta^*) \right)\right]\nonumber \\
    &= \big\|z^{t} - \theta^*\big\|^2 -2\eta_t\lambda_{t+1}  \mathbb{E}\left[ \sum_{i\in \mathcal{B}} \frac{1}{b} ( f_i(\theta^{t})-f_i(\theta^*))\right] +2\eta_t\lambda_{t+1}  \mathbb{E}\left[ \sum_{i\in \mathcal{B}} \left(\frac{1}{b}-w_{i,t}\right) ( f_i(\theta^{t})-f_i(\theta^*))\right]\nonumber \\ 
    &+ 2 \eta_t\lambda_t \mathbb{E}\left[\sum_{i\in \mathcal{B}} \frac{1}{b} \left( f_i(\theta^{t-1}) - f_i(\theta^*) \right)\right] + 2 \eta_t\lambda_t \mathbb{E}\left[\sum_{i\in \mathcal{B}} \left(w_{i,t}-\frac{1}{b}\right) \left( f_i(\theta^{t-1}) - f_i(\theta^*) \right)\right]\nonumber \\
    &= \big\|z^{t} - \theta^*\big\|^2 -2\eta_t\lambda_{t+1}  \mathbb{E} ( f(\theta^{t})-f(\theta^*)) +2\eta_t\lambda_{t+1}  \mathbb{E}\left[ \sum_{i\in \mathcal{B}} \left(\frac{1}{b}-w_{i,t}\right) ( f_i(\theta^{t})-f_i(\theta^*))\right]\nonumber \\ 
    &+2 \eta_t\lambda_t \mathbb{E}\left( f(\theta^{t-1}) - f(\theta^*) \right) + 2 \eta_t\lambda_t \mathbb{E}\left[\sum_{i\in \mathcal{B}} \left(w_{i,t}-\frac{1}{b}\right) \left( f_i(\theta^{t-1}) - f_i(\theta^*) \right)\right]\nonumber \\
    &= \big\|z^{t} - \theta^*\big\|^2 -2\eta_t\lambda_{t+1}  \mathbb{E} ( f(\theta^{t})-f(\theta^*)) +\eta_t  \mathbb{E}\left[ \sum_{i\in \mathcal{B}} \left(\frac{1}{b}-w_{i,t}\right) ( f_i(\theta^{t})-f_i(\theta^*))\right]\nonumber \\ 
    &+ 2 \eta_t\lambda_t \mathbb{E}\left( f(\theta^{t-1}) - f(\theta^*) \right) + 2 \eta_t\lambda_t \mathbb{E}\left[\sum_{i\in \mathcal{B}} \left(\frac{1}{b}-w_{i,t}\right) \left(f_i(\theta^t)- f_i(\theta^{t-1})  \right)\right]
    \label{eq:desc1_mbmo}
\end{align}
where (a) follows from applying \Cref{ass:conv} to the third and fourth terms, (b) follows from applying \Cref{lemma:convlip} to the second term, and (c) follows from setting $\eta_t$ as $\eta_t \leq \frac{1}{4Lbw_{\text{max},t}}$ and $\lambda_{t+1}=\lambda_t+\frac{1}{2}$. 
Let $\delta_t=\mathbb{E}\left[ \sum_{i\in \mathcal{B}} \left(\frac{1}{b}-w_{i,t}\right) ( f_i(\theta^{t})-f_i(\theta^*))\right]$, $\mu_t=\mathbb{E}\left[ \sum_{i\in \mathcal{B}} \left(\frac{1}{b}-w_{i,t}\right) ( f_i(\theta^{t})-f_i(\theta^{t-1}))\right]$ and $\sum_{i\in \mathcal{B}}w_{i,t}^2\leq 1$. 
Hence, telescoping \Cref{eq:desc1_mbmo} from $t=0$ to $T-1$ and noting that $z_0=x_0$ yields
\begin{align}
   \mathbb{E}\big\| z^{T} - \theta^*\big\|^2 \leq & ~\big\| \theta^{0} - \theta^*\big\|^2 -  2\eta\lambda_{T+1} \mathbb{E}\left(f(\theta^T) - f(\theta^*)\right) + \sum_{t=0}^{T-1} 2\eta\delta_t + \sum_{t=0}^{T-1} 2\eta\lambda_{t}\mu_{t}.
\end{align}
With $\lambda_t=t/2$, for a weighting scheme such that $w_{\textrm{max},t} \leq \frac{2}{b}$ so that $\eta_t$ can be fixed to $\eta = \frac{1}{8L\sqrt{T+1}}$, we obtain
\begin{align}
   \mathbb{E}[f( \theta^T) - f(\theta^*)] &\leq\frac{8L\big\| \theta^{0} - \theta^*\big\|^2}{\sqrt{T+1}} +\frac{2}{{T+1}}\sum_{t=0}^{T-1}{\delta_t} + \frac{2}{{T+1}}\sum_{t=0}^{T-1}{\lambda_t\mu_{t}}.
\end{align}
\end{proof}

\subsection{Non-convex Minibatch SGD}
\label{sec:nonconvex_minibatch}
{\bf Non-convex:} Unlike the convex case, the importance weight $w_{i,t}$ is set based on the gradient norm instead of the loss gap $f_i(\theta^{t}) - f_i( \theta^{*})$ when the sample functions $f_i$ are non-convex. \Cref{thm:minibatch_SGD_nonconvex} underlines the influence of these importance weights on the convergence bound for the non-convex case. Note that computing the gradient norm requires extra computations, whereas the loss values are readily available in practice.

\begin{assum}
\label{ass:bounded_variance}
The norm of the difference between the sample gradient $\nabla f_i(\theta_t)$ and full gradient $\nabla f(\theta_t)$ is bounded by 
    $\left\|\nabla f_i(\theta_t)-\nabla f(\theta_t)\right\|^2\leq \mathcal{V}^2, \forall \theta_t,i\in [M]$.
\end{assum}

\begin{restatable}{theorem}{minibatchSGDnonconvex}[Non-convex Minibatch SGD]
\label{thm:minibatch_SGD_nonconvex}
Given that the objective functions $f_i,i\in [n]$ satisfy \Cref{ass:lip} and \Cref{ass:bounded_variance}. Consider a minibatch of data points $|\mathcal{B}|=b$ with reweighting scheme $w_{i,t}, \text{ if } i\in \mathcal{B}$ and 0 otherwise where $\max_{i\in \mathcal{B}}w_{i,t}\leq 2/b$. Then, for $\eta = \frac{1}{8L\sqrt{T}}$, we have 
    \begin{align}
    \label{eq:minibatch_nonconvex_bound}
   \frac{1}{T}\sum_{t=0}^{T-1}\mathbb{E}\|\nabla f(\theta^t)\|^2\leq \frac{16L(f(\theta^0)-f^*)}{\sqrt{T}}+\frac{8}{T}\sum_{t=0}^{T-1}\mathbb{E}\left[\sum_{i\in \mathcal{B}}\left(\frac{1}{b}-w_{it}\right)\| \nabla f_i(\theta^t)\|^2\right]+\frac{ b\mathcal{V}^2}{4\sqrt{T}}.
\end{align}
\end{restatable}    

\begin{proof}
Applying $L$-smoothness condition
    \begin{align}
        &f(\theta^{t+1})\nonumber\\
        &\leq f(\theta^{t})+\langle \nabla f(\theta^{t}),\theta^{t+1}-\theta^{t}\rangle +\frac{L}{2}\|\theta^{t+1}-\theta^t\|^2\nonumber \\
        &= f(\theta^{t})-\eta\left\langle \nabla f(\theta^{t}),\sum_{i\in \mathcal{B}}w_{it}\nabla f_i(\theta^t)\right\rangle +\frac{L\eta^2}{2}\left\|\sum_{i\in \mathcal{B}}w_{it}\nabla f_i(\theta^t)\right\|^2\nonumber \\
        &= f(\theta^{t}) - \eta\left\langle \nabla f(\theta^{t}),\sum_{i\in \mathcal{B}}\frac{1}{b}\nabla f_i(\theta^t)\right\rangle +\eta\left\langle \nabla f(\theta^{t}),\sum_{i\in \mathcal{B}}\left(\frac{1}{b}-w_{it}\right)\nabla f_i(\theta^t)\right\rangle +\frac{L\eta^2}{2}\left\|\sum_{i\in \mathcal{B}}w_{it}\nabla f_i(\theta^t)\right\|^2\nonumber \\
        &= f(\theta^{t}) - \eta\left\langle \nabla f(\theta^{t}),\sum_{i\in \mathcal{B}}\frac{1}{b}\nabla f_i(\theta^t)\right\rangle +\eta\sum_{i\in \mathcal{B}}\left(\frac{1}{b}-w_{it}\right)\left\langle \nabla f(\theta^{t}),\nabla f_i(\theta^t)\right\rangle +\frac{L\eta^2}{2}\left\|\sum_{i\in \mathcal{B}}w_{it}\nabla f_i(\theta^t)\right\|^2\nonumber \\
        &\leq f(\theta^{t}) - \eta\left\langle \nabla f(\theta^{t}),\sum_{i\in \mathcal{B}}\frac{1}{b}\nabla f_i(\theta^t)\right\rangle +2\eta\sum_{i\in \mathcal{B}}\left(\frac{1}{b}-w_{it}\right)\|\nabla f(\theta^{t})\|^2 + 2\eta\sum_{i\in \mathcal{B}}\left(\frac{1}{b}-w_{it}\right)\left\| \nabla f_i(\theta^t)\right\|^2\nonumber \\ &+\frac{L\eta^2}{2}\left\|\sum_{i\in \mathcal{B}}w_{it}\nabla f_i(\theta^t)\right\|^2\nonumber \\
        &= f(\theta^{t}) - \eta\left\langle \nabla f(\theta^{t}),\sum_{i\in \mathcal{B}}\frac{1}{b}\nabla f_i(\theta^t)\right\rangle + 2\eta\sum_{i\in \mathcal{B}}\left(\frac{1}{b}-w_{it}\right)\left\| \nabla f_i(\theta^t)\right\|^2+\frac{L\eta^2}{2}\left\|\sum_{i\in \mathcal{B}}w_{it}\nabla f_i(\theta^t)\right\|^2\nonumber \\
        &\leq f(\theta^{t}) - \eta\left\langle \nabla f(\theta^{t}),\sum_{i\in \mathcal{B}}\frac{1}{b}\nabla f_i(\theta^t)\right\rangle + 2\eta\sum_{i\in \mathcal{B}}\left(\frac{1}{b}-w_{it}\right)\left\| \nabla f_i(\theta^t)\right\|^2+\frac{Lb\eta^2}{2}\sum_{i\in \mathcal{B}}w_{it}^2\left\|\nabla f_i(\theta^t)\right\|^2\nonumber \\
        &\leq f(\theta^{t}) - \eta\left\langle \nabla f(\theta^{t}),\sum_{i\in \mathcal{B}}\frac{1}{b}\nabla f_i(\theta^t)\right\rangle + 2\eta\sum_{i\in \mathcal{B}}\left(\frac{1}{b}-w_{it}\right)\left\| \nabla f_i(\theta^t)\right\|^2\nonumber \\&+{Lb\eta^2}\sum_{i\in \mathcal{B}}w_{it}^2\left\|\nabla f_i(\theta^t)-\nabla f_i(\theta^t)\right\|^2+{Lb\eta^2}\sum_{i\in \mathcal{B}}w_{it}^2\left\|\nabla f(\theta^t)\right\|^2\nonumber \\
        &\leq f(\theta^{t}) - \eta\left\langle \nabla f(\theta^{t}),\sum_{i\in \mathcal{B}}\frac{1}{b}\nabla f_i(\theta^t)\right\rangle + 2\eta\sum_{i\in \mathcal{B}}\left(\frac{1}{b}-w_{it}\right)\left\| \nabla f_i(\theta^t)\right\|^2\nonumber \\&+{Lb\eta^2}\mathcal{V}^2+{Lb\eta^2}\sum_{i\in \mathcal{B}}w_{i,t}^2\left\|\nabla f(\theta^t)\right\|^2\nonumber \\
    \end{align}
    where we have used the bound $\left\|\nabla f_i(\theta^t)-\nabla f(\theta^t)\right\|^2\leq \mathcal{V}^2$ and $\sum_{i\in \mathcal{B}}w_{i,t}^2\leq 1$. Applying expectation on both sides, we get
    \begin{align}
        \mathbb{E}f(\theta^{t+1})&\leq f(\theta^{t}) - \eta\mathbb{E}\left\|\nabla f(\theta^{t})\right\|^2 + \left(2\eta\right)\mathbb{E}\left[\sum_{i\in \mathcal{B}}\left(\frac{1}{b}-w_{it}\right)\left\| \nabla f_i(\theta^t)\right\|^2\right]\nonumber \\&+{Lb\eta^2\mathcal{V}^2}+{Lb\eta^2}\sum_{i\in \mathcal{B}}w_{it}^2\mathbb{E}\left\|\nabla f(\theta^t)\right\|^2\nonumber \\
        &\leq f(\theta^{t}) - \eta\mathbb{E}\left\|\nabla f(\theta^{t})\right\|^2 + \left(2\eta\right)\mathbb{E}\left[\sum_{i\in \mathcal{B}}\left(\frac{1}{b}-w_{it}\right)\left\| \nabla f_i(\theta^t)\right\|^2\right]\nonumber \\&+{Lb\eta^2\mathcal{V}^2}+{Lb^2\eta^2}w_{\text{max},t}^2\mathbb{E}\left\|\nabla f(\theta^t)\right\|^2\nonumber \\
        &\overset{}{\leq} f(\theta^{t}) - \frac{\eta}{2}\mathbb{E}\left\|\nabla f(\theta^{t})\right\|^2 + 2\eta\mathbb{E}\left[\sum_{i\in \mathcal{B}}\left(\frac{1}{b}-w_{it}\right)\left\| \nabla f_i(\theta^t)\right\|^2\right]+{Lb\eta^2\mathcal{V}^2}
    \end{align}
    where $\eta\leq \frac{1}{2Lb^2w_{\text{max},t}^2}$ in the last inequality. Telescoping and rearranging the terms yields
    \begin{align}
        \frac{1}{T}\sum_{t=0}^{T-1}\mathbb{E}\|\nabla f(\theta^t)\|^2\leq \frac{2(f(\theta^0)-f^*)}{\eta{T}}+\frac{4}{T}\sum_{t=0}^{T-1}\mathbb{E}\left[\sum_{i\in \mathcal{B}}\left(\frac{1}{b}-w_{it}\right)\| \nabla f_i(\theta^t)\|^2\right]+{2Lb\eta\mathcal{V}^2}.
    \end{align}
    For a weighting scheme such that $w_{\text{max},t}\leq {\frac{2}{b}}$ so that $\eta_t$ can be fixed to $\eta = \frac{1}{8L\sqrt{T}}$, we get
    \begin{align}
        \frac{1}{T}\sum_{t=0}^{T-1}\mathbb{E}\|\nabla f(\theta^t)\|^2\leq \frac{16L(f(\theta^0)-f^*)}{\sqrt{T}}+\frac{4}{T}\sum_{t=0}^{T-1}\mathbb{E}\left[\sum_{i\in \mathcal{B}}\left(\frac{1}{b}-w_{it}\right)\| \nabla f_i(\theta^t)\|^2\right]+\frac{b\mathcal{V}^2}{4\sqrt{T}}.
    \end{align} 
    \end{proof}

From \Cref{thm:minibatch_SGD_nonconvex}, we can obtain the convergence bound of traditional minibatch SGD for non-convex functions when $w_{i,t} = \frac{1}{b} ~ \forall i\in \mathcal{B}$. Moreover, a looser convergence bound is obtained when $\mathbb{E}\left[\sum_{i\in \mathcal{B}}\left(\frac{1}{b}-w_{it}\right)\| \nabla f_i(\theta^t)\|^2\right] \geq 0$, for $w_{i,t}\leq \frac{1}{b}$. 

\end{document}